\documentclass[10pt,journal,compsoc]{IEEEtran}

\usepackage[numbers]{natbib}
\usepackage{graphicx}
\usepackage[pagebackref=true,breaklinks,colorlinks,citecolor=blue,linkcolor=blue,urlcolor=blue,hypertexnames=false]{hyperref}
\usepackage{amsmath,amsfonts}
\usepackage{algorithmic}
\usepackage{algorithm}
\usepackage{array}
\usepackage[caption=false,font=normalsize,labelfont=sf,textfont=sf]{subfig}
\usepackage{textcomp}
\usepackage{stfloats}
\usepackage{url}
\usepackage{verbatim}
\usepackage{bbm}
\usepackage{booktabs}
\usepackage{xcolor}
\usepackage{xspace}
\usepackage{multirow}
\usepackage{tablefootnote}
\usepackage{colortbl}
\usepackage{chngcntr}
\usepackage{inconsolata}
\usepackage{tabulary}
\usepackage{pifont}\newcommand{\cmark}{\ding{51}}\newcommand{\xmark}{\ding{55}}

\newcommand{\txt}[1]{{\texttt{#1}}}
\newcommand{\loss}{\mathcal{L}}

\newcommand{\E}{\mathbb{E}}

\newcommand{\s}{\text{sim}}
\newcommand{\methodname}[1]{}

\usepackage{array}
\newcolumntype{x}[1]{>{\arraybackslash\hspace{0pt}}p{#1}}

\begin{document}

\title{Foundational Models Defining a New Era in Vision: A Survey and Outlook}

\author{Muhammad Awais, Muzammal Naseer, Salman Khan, Rao Muhammad Anwer, Hisham Cholakkal, Mubarak Shah, Ming-Hsuan Yang, Fahad Shahbaz Khan
\IEEEcompsocitemizethanks{\IEEEcompsocthanksitem  M. Awais, M. Naseer, S. Khan, R. M. Anwer, H. Cholakkal, and F. S. Khan are with the MBZ University of AI, Abu Dhabi, UAE. \protect \\
E-mail: awais.muhammad@mbzuai.ac.ae
\IEEEcompsocthanksitem S. Khan and M. Naseer are also with the CECS, Australian National
University, Canberra ACT 0200, Australia.
\IEEEcompsocthanksitem  F. S. Khan is also with the Computer Vision Laboratory, Linköping
University, Sweden.
\IEEEcompsocthanksitem M. Shah is with the Center for Research in Computer Vision, University of Central Florida, Orlando, FL 32816, United States.

\IEEEcompsocthanksitem M.-H. Yang is with the University of California, Merced, CA 95344, Yonsei University, and Google Research.
}
}

\markboth{}%
{}

\IEEEpubid{}

\IEEEtitleabstractindextext{%
\begin{abstract}
Vision systems to see and reason about the compositional nature of visual scenes are fundamental to understanding our world. 
The complex relations between objects and their locations, ambiguities, and variations in the real-world environment can be better described in human language, naturally governed by grammatical rules and other modalities such as audio and depth. 
The models learned to bridge the gap between such modalities coupled with large-scale training data facilitate contextual reasoning, generalization, and prompt capabilities at test time. 
These models are referred to as \emph{foundational models}. 
The output of such models can be modified through human-provided prompts without retraining, e.g.,  segmenting a particular object by providing a bounding box, having interactive dialogues by asking questions about an image or video scene or manipulating the robot's behavior through language instructions. 
In this survey, we provide a comprehensive review of such emerging foundational models, including typical architecture designs to combine different modalities (vision, text, audio, etc), training objectives (contrastive, generative), pre-training datasets, fine-tuning mechanisms, and the common prompting patterns; textual, visual, and heterogeneous. 
We discuss the open challenges and research directions for foundational models in computer vision, including difficulties in their evaluations and benchmarking, gaps in their real-world understanding, limitations of their contextual understanding,  biases, vulnerability to adversarial attacks, and interpretability issues. 
We review recent developments in this field, covering a wide range of applications of foundation models systematically and comprehensively. 
A comprehensive list of foundational models studied in this work is available at \url{https://github.com/awaisrauf/Awesome-CV-Foundational-Models}.
\end{abstract}

\begin{IEEEkeywords}
Language and Vision, Large Language Models, Self-supervised Learning, Contrastive Learning, and Masked Modeling.
\end{IEEEkeywords}}

\IEEEdisplaynontitleabstractindextext
\maketitle

\section{Introduction}

\IEEEPARstart{R}{ecent} years have witnessed remarkable success towards developing \textit{foundation models}, that are trained on a large-scale broad data, and once trained, they operate as a basis and can be adapted (e.g., fine-tuned) to a wide range of downstream tasks related to the original trained model~\cite{bommasani2021opportunities}. 
While the basic ingredients of the foundation models, such as deep neural networks and self-supervised learning, have been around for many years, the recent surge, specifically through large language models (LLMs), can be mainly attributed to massively scaling up both data and model size \cite{Zhao2023LLM}. 
For instance, recent models with billion parameters such as GPT-3 \cite{brown2020language} have been effectively utilized for zero/few-shot learning, achieving impressive performance without requiring large-scale task-specific data or model parameter updating.
Similarly, the recent 540-billion parameter Pathways Language Model (PaLM) has demonstrated state-of-the-art capabilities on numerous challenging problems ranging from language understanding and generation to reasoning and code-related tasks \cite{chowdhery2022palm,anil2023palm}.

Concurrent to LLMs in natural language processing, large foundation models for different perception tasks have also been explored in the literature recently. 
For instance, pre-trained vision-language models (VL) such as CLIP \cite{radford2019language} have demonstrated promising zero-shot performance on different downstream vision tasks, including image classification and object detection. 
These VL foundation models are typically trained using millions of image-text pairs collected from the web and provide representations with generalization and transfer capabilities. 
These pre-trained VL foundation models can then be adapted to a downstream task by presenting it with a natural language description of the given task and prompts.
For instance, the seminal CLIP model utilizes carefully designed prompts to operate on different downstream tasks, including zero-shot classification, where the text encoder dynamically constructs the classifiers via class names or other free-form texts. 
Here, the textual prompts are handcrafted templates, e.g., “\txt{A photo of a \{label\}}", that aid in specifying the text as corresponding to the visual image content. 
Recently, numerous works have also explored adding conversational capabilities to the VL models by fine-tuning them on a specific instruction set \cite{liu2023llava,zhu2023minigpt,dai2023instructblip,maaz2023videochatgpt,ye2023mplugowl}.

\begin{figure*}[t!]
    \centering
    \includegraphics[width=.8\textwidth]{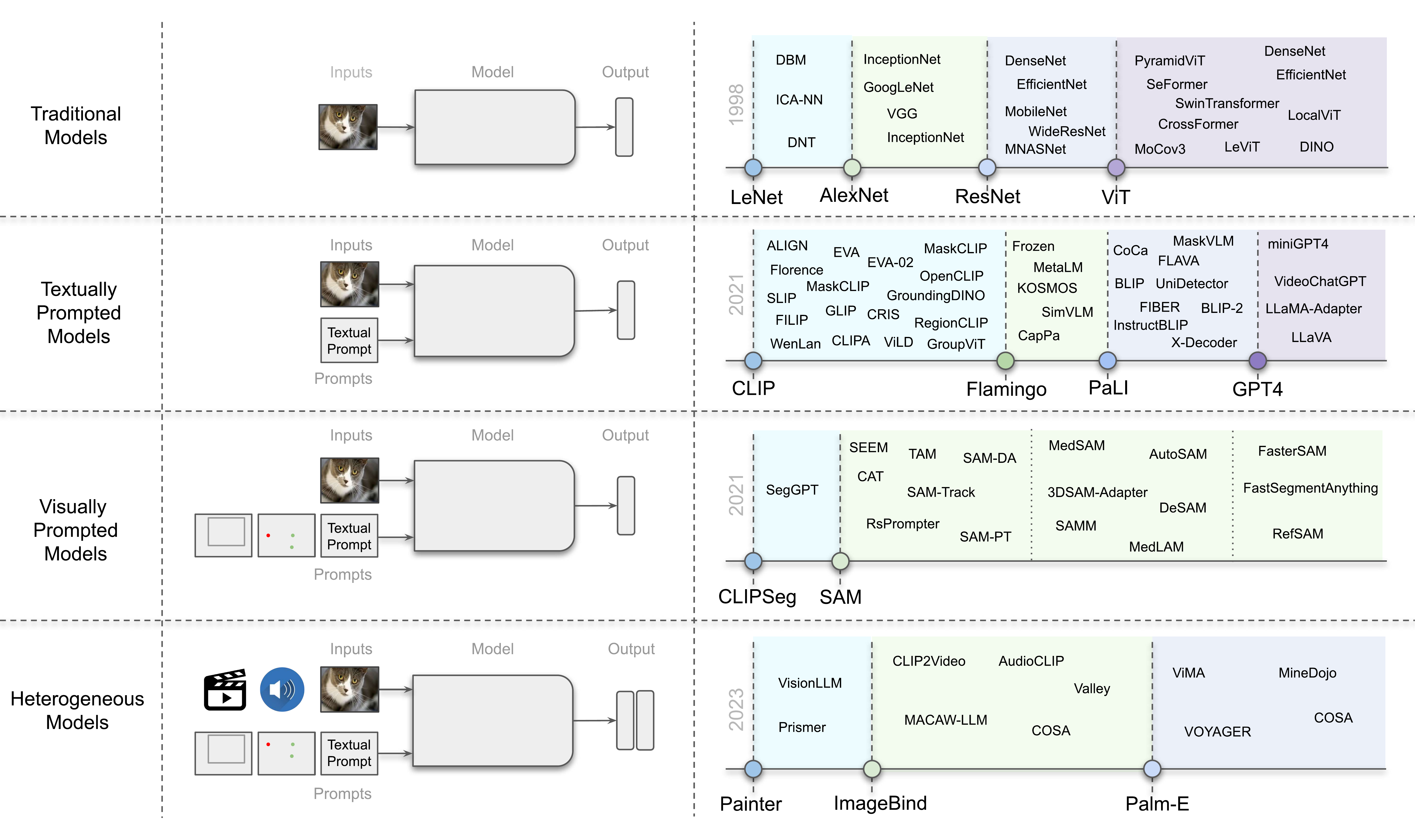}\vspace{-15pt}
    \caption{Overview of the evolution of foundational models in computer vision. (left) We show the progression of models in computer vision. (right) We show the evolution of these models with major milestones reported in the literature shown with dotted lines.}
    \label{fig:teaser}
\end{figure*}

Besides large VL foundation models, several research efforts have been devoted to developing large foundation models that visual inputs can prompt. 
For instance, the recently introduced SAM \cite{kirillov2023segment} performs a class-agnostic segmentation given an image and a visual prompt such as box, point, or mask, which specifies what to segment in an image. 
Such a model is trained on billions of object masks following a model-in-the-loop (semi-automated) dataset annotation setting. 
Further, such a generic visual prompt-based segmentation model can be adapted for specific downstream tasks such as medical image segmentation \cite{ma2023segment,wu2023medical}, video object segmentation \cite{refsam}, robotics \cite{yang2023pave}, and remote sensing \cite{chen2023rsprompter}.
In addition to textual and visual prompt-based foundation models, research works have explored developing models that strive to align multiple paired modalities (e.g., image-text, video-audio, or image-depth) to learn meaningful representations helpful for different downstream tasks \cite{girdhar2023imagebind,guzhov2021audioclip,Macaw-LLM}. 

In this work, we present a systematic review of foundation models in computer vision. 
First, we present the background and preliminaries for foundation models briefly covering common architecture types, self-supervised learning objectives, large-scale training, and prompt engineering (Sec. \ref{sec:background}). 
Then, we distinguish existing works into textually prompted (Sec. \ref{sec:textually-prompted}-\ref{sec:conversationLLMs}), visually prompted (Sec. \ref{sec:visually-prompted}), heterogeneous modality-based (Sec. \ref{sec:Heterogeneous Modalities based Models}) and embodied foundation models (Sec. \ref{sec:Embodied Foundational Agents}). 
Within the textually prompted foundation models, we further distinguish them into contrastive, generative, hybrid (contrastive and generative), and conversational VL models. 
Finally, we discuss open challenges and research directions based on our analysis (Sec. \ref{sec:Open Challenges and Research Directions}). 
Next, we review other surveys related to ours and discuss the differences and uniqueness. 

\noindent \textbf{Related Reviews and 
Differences.}
In the literature,  few recent works have reviewed large language models (LLMs) in natural language processing \cite{Zhao2023LLM,fan2023bibliometric,Huang2023ACL,dong2022survey,zhou2023comprehensive}. 
The work of~\citet{Zhao2023LLM} reviews recent advances in LLMs, distinguishing different aspects of LLMs such as pre-training, adaptation tuning, LLM utilization, and evaluation. 
This survey also summarizes resources available to develop LLMs and discusses potential future directions. 
The work of~\cite{Huang2023ACL} discusses LLMs in terms of their reasoning capabilities in performing a benchmark evaluation. 
A practical guide for practitioners using LLMs is presented in~\cite{Jingfeng2023LLM}, where a detailed discussion and insights are provided regarding the usage of LLMs from the viewpoint of downstream tasks. 
This work also analyzes the impact of pre-training, training, and testing data on LLMs. Furthermore, the work also discusses different limitations of LLMs in real-world scenarios. 
In the context of VLMs, the work of~\cite{Long2022VLM} performs a preliminary review of vision-language pre-trained models regarding task definition and general architecture. 
Similarly, \cite{Yifan2022VLM} discusses different techniques to encode images and texts to embeddings before the pre-training step and reviews different pre-training architectures. 
The work of~\cite{Peng2023MMT} reviews transformers techniques for multimodal data with a survey of vanilla transformers, vision transformers, and multimodal transformers from a geometrically topological perspective. 
In the context of multimodal learning, the recent review~\cite{Zong2023SSML} focuses on self-supervised multimodal learning techniques to effectively utilize supervision from raw multimodal data. 
The survey distinguishes existing approaches based on objective functions, data alignment, and architectures. 
The work of~\cite{Zhang2023VLMs,gan2022vision}  summarizes different vision-language pre-training network architectures, objectives, and downstream tasks and categorizes vision-language pre-training frameworks. 
Recently, the work of~\cite{Chunhui2023SAM} reviews the visually prompted foundation segmentation model, segmenting anything, and discusses its potential downstream tasks. 

The main differences between this survey and the aforementioned works are as follows. 
Unlike previous surveys that primarily focus on textual prompt-based vision-language models, our work focuses on the three different classes of foundation models: textually prompted models (contrastive, generative, hybrid, and conversational), visually prompted models (e.g., SegGPT \cite{wang2023seggpt}, SAM \cite{kirillov2023segment}) and heterogeneous modalities-based models (e.g., ImageBind \cite{girdhar2023imagebind}, Valley \cite{luo2023valley}).
We present the background theory behind foundation models, briefly covering from architectures to prompt engineering (Sec. \ref{sec:background}). Our work provides an extensive and up-to-date overview of the recent vision foundation models (Sec. \ref{sec:textually-prompted}, \ref{sec:visually-prompted}, \ref{sec:Heterogeneous Modalities based Models}, and \ref{sec:Embodied Foundational Agents}).
Finally, we present a detailed discussion on open challenges and potential research directions of foundation models in computer vision (Sec. \ref{sec:Open Challenges and Research Directions}). 

\begin{figure*}[ht]
    \centering
    \includegraphics[width=1\textwidth]{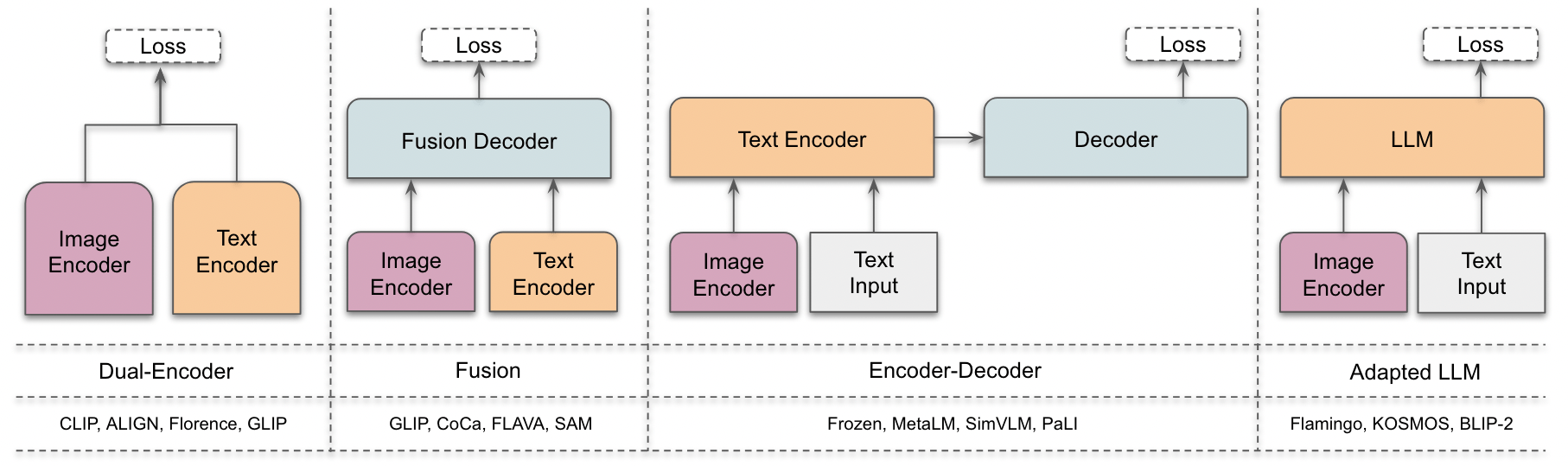}
    \vspace{-2em}
    \caption{Overview of four different architecture styles described in this survey. \emph{Left} to \emph{right}: \textbf{a)} Dual-encoder designs use a parallel visual and language encoder with aligned representations. \textbf{b)} Fusion designs jointly process both image and text representations via a decoder. Here the image encoder can also process visual prompts (e.g., points and boxes in the case of SAM~\cite{kirillov2023segment}). \textbf{c)} Encoder-decoder designs apply joint feature encoding and decoding sequentially. \textbf{d)} Adapter LLM designs input visual and text prompts to the LLMs to leverage their superior generalization ability. Examples for each category are shown in the bottom row. More detail about these architectures is discussed in Sec.~\ref{sec:arch}. }
    \label{fig:arch_types}
\end{figure*}

\section{Preliminaries}
\label{sec:background}
We first define the foundational models and scope of this survey. Then, we present a concise background overview to help readers understand the rest of the material. 
We focus on three main contributing factors for Foundational Models in computer vision: a) Model architecture, b) Training objectives, and c) Large-scale training and prompting. 

\subsection{Foundational Models and Scope of the Survey}

The term “foundational models” was first introduced by \citet{bommasani2021opportunities} at Stanford Institute for Human-Centered AI. Foundational models are defined as \emph{“the base models trained on large-scale data in a self-supervised or semi-supervised manner that can be adapted for several other downstream tasks”}. The paradigm shift towards foundational models is significant because it allows replacing several narrow task-specific models with broader and generic base models that can be once trained and quickly adapted for multiple applications. It not only enables rapid model development and provides better performance for both in-domain and out-domain scenarios, but also leads to the so-called “emergent properties” of intelligence from large-scale foundational models trained on massive datasets \cite{wei2022emergent,bubeck2023sparks}. 

Computer vision has recently witnessed remarkable progress fueled by foundational models \cite{yuan2021florence, kirillov2023segment} with an extensive body of literature encompassing both discriminative and generative models. In this survey, we focus on Multimodal (vision and language) Foundational Models trained on large-scale data that can be adapted for several computer vision tasks involving non-image outputs (e.g., generated text, segmentation masks). Note that we do not cover image generative models aimed to model data distribution such as GANs, VAEs, and Diffusion models owing to dedicated surveys already existing in this area \cite{cao2022survey,zhang2023text,yang2022diffusion,croitoru2023diffusion} and because the former model class can cover a broader range of downstream applications.

\subsection{Architecture Types}
\label{sec:arch}
As depicted in Fig.~\ref{fig:arch_types}, Vision-Language (VL) models primarily use four architectural designs. We start by introducing the Dual-Encoder architecture, wherein separate encoders are utilized for processing visual and textual modalities. The output of these encoders is subsequently optimized through an objective function. The second architecture type, fusion, incorporates an additional fusion encoder, which takes the representations generated by the vision and text encoders and learns fused representations. The third type, Encoder-Decoder, consists of an encoder-decoder-based language model and a visual encoder. Lastly, the fourth architecture type, Adapted LLM, leverages a Large Language Model (LLM) as its core component, with a visual encoder employed to convert images into a format compatible with the LLM. For a more comprehensive understanding of these architectures, we refer the readers to the corresponding sections of the survey where each work is discussed. Next, we discuss the loss functions used to train different architecture types.

\subsection{Training Objectives}
\label{sec:losses}

\subsubsection{Contrastive Objectives}

To learn from unlabeled image-text data, ~\cite{radford2021learning, jia2021scaling} utilized a simple Image-Text Contrastive (\textbf{ITC}) loss which aims to learn representations by learning to predict correct image-text pairs. Given a batch of $N$ examples, ITC loss aims to match correct image-text pairs among $N\times N$ possible configurations. ITC loss maximizes cosine similarity between $N$ correct pairs and minimizes it among $N^2-N$ incorrect pairs. Let $(x_i, t_i)$ be $i$-th image-text example and $(v_i, t_i)$ be its corresponding representations, then image-to-text loss is calculated as follows, 

$$
\loss_{v2t} = -  \log \bigg[ \dfrac{\exp(\s(v_i, t_i)/\tau)}{\sum_{j=1}^N \exp( \s(v_i, t_j)/\tau) } \bigg],
$$
where $\tau$ is temprature. Text-to-image loss is also calculated similarly and the total loss is the sum of these two terms, 
$
\loss_{ITC} = \dfrac{1}{N} \sum_{i=1}^N [\loss_{v2t} + \loss_{t2v}]
$. 

Image-Text Matching (\textbf{ITM}) loss~\cite{li2021align} aims to correctly predict whether a pair of images and text is positively or negatively matched. A few perceptron layers are added which predicts the probability $p^{itm}$ of a pair being matched. The loss is then calculated based on cross-entropy loss. 

Similar to ITC and ITM, several contrastive losses have also been utilized in the subsequent papers. These losses include image-based self-supervision losses (i.e. Simple Contrastive Learning of Representations (\textbf{SimCLR})~\cite{chen2020simple, chen2020big}) and variants of ITC loss (i.e. \textbf{FILIP Loss}~\cite{yao2021filip}, Text-to-Pixel Contrastive (\textbf{TPC}) Loss~\cite{wang2022cris}, Region-Word Alignment (\textbf{RWA})~\cite{li2022grounded}, Multi-label Image-Text Contrastive (\textbf{MITC})~\cite{xu2022groupvit}, Unified Contrastive Learning (\textbf{UniCL})~\cite{yang2022unified}, Region-Word Contrastive (\textbf{RWC}) Loss \cite{zhang2022glipv2}).

\subsubsection{Generative Objectives}

Masked Language Modeling (\textbf{MLM}) loss~\cite{lu2019vilbert} is a bi-directional, non-casual language modeling loss that aims to reconstruct masked-out tokens. Let's assume $\hat{x}^t$ to be masked input tokens where a certain percentage of tokens are randomly masked and replaced with some special token of choice. MLM aims to model $x^t$ given masked tokens, 
$$
\loss_{\text{MLM}} = - \E_{x^t \sim D} \big[ \log p(x^t | \hat{x}^t)  \big].
$$

Language Modeling (\textbf{LM}) loss aims to model language generation in an auto-regressive manner. It models the prediction of the current token ($l$-th) given previous tokens ($<l$),
\begin{align*}
    \loss_{\text{LM}} = - \E_{x^t \sim D} \bigg[ \sum_{l=1}^L \log p(x^t_l | x^t_{<l}) \bigg], 
\end{align*}

where $L$ is the total number of tokens. 

\noindent Standard Captioning (\textbf{Cap}) loss~\cite{hossain2019comprehensive} also aims to predict the next token given previous tokens and image ($x^v$), 
$$
\loss_{\text{Cap}} = - \E_{x \sim D} \sum_{l=0}^L \log p(x^t_l | x^t_{<l}, x^v), 
$$
where $x = [x^v, x^t]$. 
Similarly, \textbf{Flamingo} Loss by \citet{alayrac2022flamingo} is also about the prediction of $l$-th token given previous image and text tokens.
This is different from captioning loss as tokens consist of several interleaved image and text inputs. 
Prefix Language Modeling (\textbf{PrefixML})~\cite{wang2021simvlm} extends language to vision-language modeling loss. It considers images to be a prefix of textual descriptions as they appear before images on the web, and therefore appended image tokens to text tokens $x = [x^v, x^t]$. Then, a prefix sequence of randomly selected length ($L_p$) is truncated from text tokens reconstructed through, 
\begin{equation*}
\begin{aligned}
 \loss_{\text{PrefixLM}} & = - \E_{x \sim D}\big[ \log p_{\theta}(x_{\geq L_p} | x_{(x<L_p)}\big] \\
 & = - \E_{x \sim D} \bigg[ \sum_{l=L_p}^L \log p_{\theta} (x_l | x_{[L_p, l]}, x_{<L_p}) \bigg],
\end{aligned}
\end{equation*}

$x_l$ represents current token, $x_{[L_p, l]}$ is the prefix sequence, and $x_{<L_p}$ is previous sequence. 
  
Similarly, several other generative losses have also been proposed. Some examples include Masked Multimodal Modeling (\textbf{MMM}) loss~\cite{singh2022flava}, Semi Casual Language Modeling (\textbf{SemiCasualLM}) \cite{hao2022language}, Image-conditioned Masked Language Modeling (\textbf{IMLM}) loss \cite{zhang2023toward}, Image-grounded Text Generation (\textbf{ITG}) loss~\cite{li2023blip}, Masked Image Modeling (\textbf{MIM})~\cite{zhang2023toward} and Captioning with Parallel prediction (\textbf{CapPa})~\cite{tschannen2023image}.

\subsection{Large-scale Training}
\label{sec:data}
Large-scale training followed by effective prompting at inference has been a crucial ingredient of vision and language foundational models. We discuss here the role of pre-training, finetuning, and prompting techniques (see Tab.~\ref{tab:datasets}).

\begin{table}[]
    \centering
    \setlength{\tabcolsep}{12pt}
    \resizebox{\columnwidth}{!}{%
    \begin{tabular}{lr}
    \toprule
        Data Type  & Examples \\
    \toprule
        \multicolumn{2}{c}{\textbf{Pre-training} (Sec.~\ref{sec:pre_training})} \\
        \midrule
        Image-Text  & WIT~\cite{radford2021learning}, LAION~\cite{schuhmann2021laion, schuhmann2022laion} \\
        w/ Pseudo Labels  & Cap24M~\cite{li2022grounded}, SA-1B~\cite{kirillov2023segment} \\
        Benchmark Combination & UNITER~\cite{chen2020uniter}, PMD~\cite{singh2022flava} \\
        \midrule
        \multicolumn{2}{c}{\textbf{Fine-tuning} (Sec.~\ref{sec:fine_tuning})} \\
        \midrule
        Task Specific &  ImageNet~\cite{fei2009imagenet}\\
        Capability Specific & OWL-ViT~\cite{minderer2022simple} \\
        Instruction-Following &  InstructBLIP~\cite{dai2023instructblip}\\
        \midrule
        \multicolumn{2}{c}{\textbf{Prompt Engineering} (Sec.~\ref{sec:prompt_engineering})} \\
        \midrule
        Train Time  & GLIP~\cite{li2022grounded} \\
        Evaluation Time  & CLIP~\cite{radford2021learning}\\
        \bottomrule
    \end{tabular}}
    \vspace{0.5em}
    \caption{An overview of different settings under which datasets are utilized for training, fine-tuning, and prompting in foundational models. More details are discussed in Sec.~\ref{sec:data}.}
    \label{tab:datasets}
\end{table}

\subsubsection{Pre-training Data}
\label{sec:pre_training}
Large-scale data is at the heart of modern vision-language foundational models. The datasets that have been utilized to pre-train these models can be divided into three broad categories: image-text datasets (such as WebImageText used in CLIP~\cite{radford2021learning}), partially synthetic datasets (such as SA-1B used in SAM~\cite{kirillov2023segment}), and Combination dataset (such as PMD used in FLAVA~\cite{singh2022flava}). Here, we discuss these categories briefly. 

\noindent \textbf{Image-Text Data: } CLIP~\cite{radford2021learning} showed remarkable effectiveness of web-scale image-text data to pre-train foundational models. This type of data is often combed through from web crawls (e.g., CommonCrawl~\footnote{https://commoncrawl.org}). The final dataset is the result of a filtering process that is applied to remove noisy, useless, or harmful data points. Numerous subsequent works collected similar datasets such as ALIGN1.8B~\cite{jia2021scaling}, RUC-CAS-WenLan~\cite{huo2021wenlan}, FLD900M~\cite{yuan2021florence}, FILIP300M~\cite{yao2021filip}, WebLi~\cite{chen2022pali}, etc). However, these datasets are not public. To make large-scale training more accessible, several open-source curation efforts have contributed significantly to the community such as LAION~\cite{schuhmann2021laion, schuhmann2022laion} and COYO-700M~\cite{kakaobrain2022coyo-700m}. 

\noindent \textbf{Partially Pseudo Labels-based Data: } 
Similar to image-text models, visual grounding can also benefit from the large-scale training data but such datasets are not available on the web. Collecting grounding datasets is also costly as they require significant human annotation effort. One cost-effective method is to leverage a good teacher to convert image-text datasets into mask-description datasets. This strategy was first adopted by GLIP~\cite{li2022grounded}, however, \citet{kirillov2023segment} took it to a billion scale with SA-1B. The process of curation often involves training a good teacher for the generation of masks and then utilizing it on an Image-Text dataset along with NLP parsers. These datasets include SAM, GLIP, and KOSMOS-2. GLIP~\cite{li2022grounded} trained a teacher GLIP on human-annotated LVIS~\cite{gupta2019lvis} and Visual Genome~\cite{krishna2017visual} datasets and then utilize it to predict boxes for image-text data with noun phrases detected by an NLP model. GRIT used in KOSMOS-2~\cite{peng2023kosmos2} is also prepared in a similar fashion. SAM~\cite{kirillov2023segment} introduced a data engine that consists of three stages (assisted manual, semi-automatic, and fully automatic stage). They generated one billion high-quality masks with this process. 

\begin{figure*}[ht!]
    \centering
    \includegraphics[width=1\textwidth]{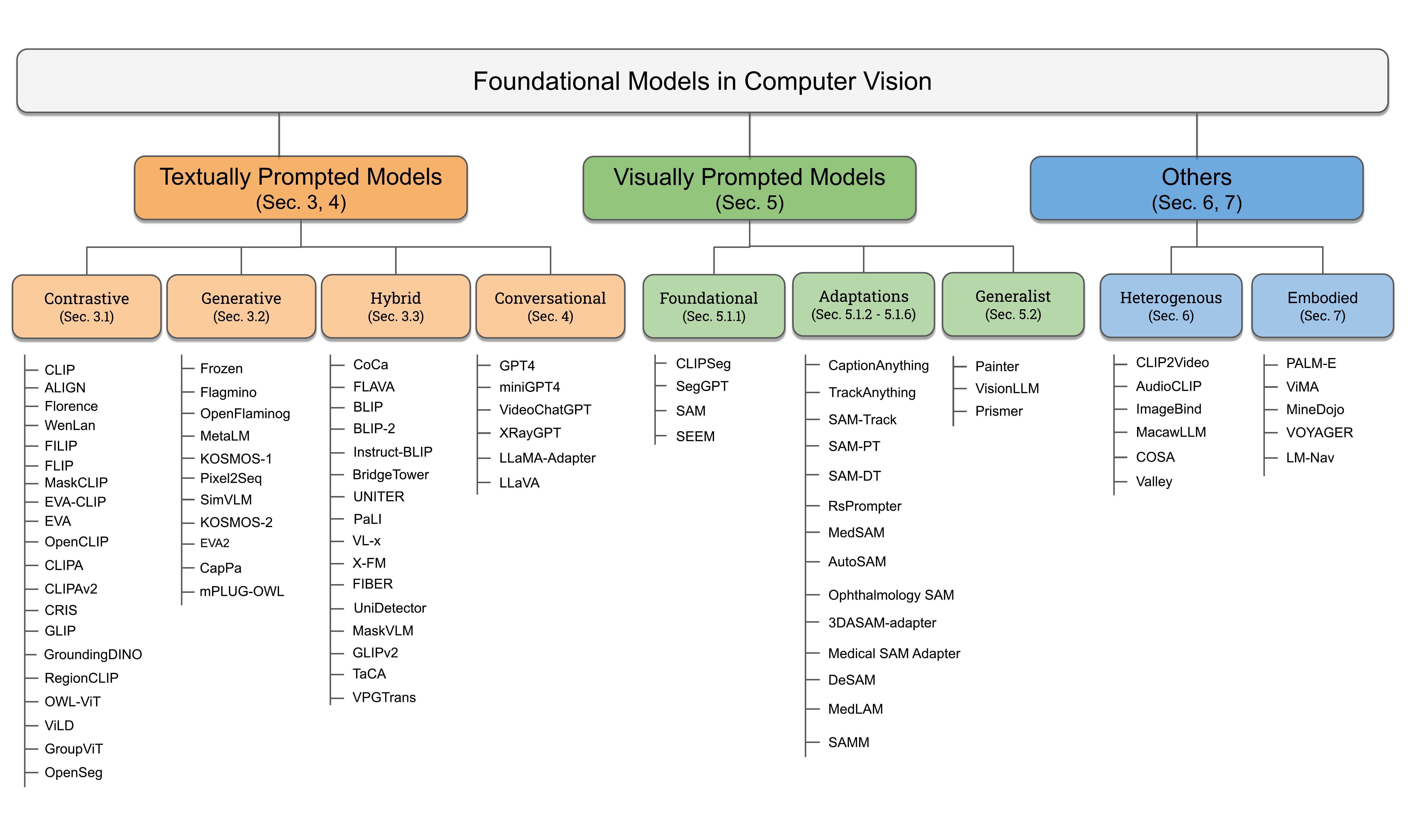}
     \vspace{-3em}
    \caption{An overview of our taxonomy for vision-language foundational models. We categorize these foundational models into six main groups based on their inputs,  outputs, and utilization. Textually prompted models are discussed in Sec.~\ref{sec:textually-prompted} and Sec.~\ref{sec:conversationLLMs}, visually prompted and generalist models are discussed in Sec.~\ref{sec:visually-prompted}, and heterogeneous and embodied models are discussed in Sec.~\ref{sec:Heterogeneous Modalities based Models} and Sec.~\ref{sec:Embodied Foundational Agents}, respectively.}
    \label{fig:categorization}
   
\end{figure*}

\noindent \textbf{Combination of Datasets: } It is not always possible to curate and train on web-scale datasets. To circumvent this problem, several works~\cite{chen2020uniter, tsimpoukelli2021multimodal, xu2022unifying} have utilized a combination of benchmark vision datasets. These works combine datasets that have image-text pairs such as captioning and visual questioning answering, etc. Some works have also used non-image-text datasets and used template-based prompt engineering to convert labels into descriptions. Moreover, visual grounding-related works have also utilized grounding datasets such as COCO~\cite{lin2014microsoft}, OpenImages~\cite{krasin2017openimages}, Objects365~\cite{shao2019objects365}.

\subsubsection{Fine-tuning}
\label{sec:fine_tuning}
Fine-tuning is employed under three primary settings: to improve a model's performance on a specific task (e.g., open-world object detection), to improve a model for a certain capability (e.g., visual grounding), and to instruction-tune a model to make it solve different downstream vision tasks (e.g., InstructBLIP~\cite{dai2023instructblip}). Firstly, a model's performance for a specific task can be improved even if only a linear layer is fine-tuned. Hence, task-specific datasets (e.g., ImageNet) can be used to improve pre-trained models for specific tasks. Second, some works have utilized a pre-trained vision-language model for grounding tasks by fine-tuning the model on grounding datasets. For instance, \citet{minderer2022simple} fine-tuned vision transformer on detection datasets to create an open vocabulary object detector. Finally, some works (such as InstructBLIP~\cite{dai2023instructblip}) transformed vision datasets into instruction-tuning datasets to enable VL models to work for downstream tasks.

\subsubsection{Prompt Engineering}
\label{sec:prompt_engineering}
Prompt engineering has primarily been used with Large Language Models (LLMs) to make them do certain tasks~\cite{brown2020language, gao2021making}. In the context of vision-language models or visually-prompted models, prompt engineering is predominately used for two purposes: to convert vision datasets to image-text training data (e.g., CLIP for image classification) to provide human intractability to the foundational models, and use vision-language models for vision tasks. Most vision datasets consist of images and corresponding one-word labels. To leverage vision-language models on vision datasets, several works have utilized template-based prompt engineering. In this prompt engineering, a set template is used to generate a description from the label. For instance, \txt{`image of a \{label\}', `a type of \{type\}'}. As noted by \cite{radford2021learning,yuan2021florence}, additional context helps the model, hence, these text prompts can be utilized by vision-language models during training or evaluation.

With this context into architecture types, objectives, and data used for training foundational models in vision, next we explain their main classes i.e., textually prompted (Sec. \ref{sec:textually-prompted} and \ref{sec:conversationLLMs}) and visually prompted (Sec. \ref{sec:visually-prompted}) models as well as heterogeneous (Sec. \ref{sec:Heterogeneous Modalities based Models}), generalist (Sec. \ref{subsec:Generalist Models}) and embodied (Sec. \ref{sec:Embodied Foundational Agents}) foundational models (see Fig. \ref{fig:categorization} for a taxonomy of vision-language foundation models).

\section{Textually Prompted Models}
\label{sec:textually-prompted}

\begin{table*}
\centering
\resizebox{1\textwidth}{!}{%
\begin{tabular}{l|cx{4cm}r|lx{3cm}l|lx{4cm}|cr}
\toprule
\multirow{2}[3]{*}{Method} &  \multicolumn{3}{|c}{\underline{Multimodal Pretraining data}} & \multicolumn{3}{|c}{\underline{Pretraining Objectives}} &  \multicolumn{2}{|c}{\underline{Architecture}} & \multicolumn{2}{|c}{\underline{Information}} \\
 & Public & \multicolumn{1}{c}{Dataset(s)} & Size & Contrastive & Generative & Others  & Type & \multicolumn{1}{c}{Base} &  \url{Link} & Venue  \\
 \midrule
CLIP~\cite{radford2021learning}  & \xmark     & WebImageText~\cite{radford2021learning}   & 400M    & ITC     & -  & - & Dual-Enc & ResNet~\cite{he2016deep}, ViT~\cite{dosovitskiy2020image}, GPT2~\cite{radford2019language} & \href{https://github.com/openai/CLIP}{Link} & arXiv'21\\

ALIGN~\cite{jia2021scaling}      & \xmark     & ALIGN1.8B~\cite{jia2021scaling}      & 1800M   & ITC      & -  & - & Dual-Enc & EffNet-L2~\cite{tan2019efficientnet}, BERT-Large~\cite{devlin2018bert}&  - & ICML'21 \\

WenLan~\cite{huo2021wenlan}      & \xmark     & RUC-CAS-WenLan~\cite{huo2021wenlan} & 30M    & InfoCE   & -  & - &  Dual-Enc & RoBERTAa-Large\footnote{\url{https://github.com/brightmart/roberta_zh}}, FasterRCNN~\cite{ren2015faster}, EffNet-B7~\cite{tan2019efficientnet} &  \href{https://github.com/BAAI-WuDao/BriVl}{Link} & arXiv'23 \\ 
 Florence~\cite{yuan2021florence} & \xmark     & FLD-900M~\cite{yuan2021florence}      & 900M    & UniCL    & - & -  & Dual-Enc& CoSwnT~\cite{dong2022cswin}, GPT2~\cite{radford2019language} & - & ECCV'22\\

FILIP~\cite{yao2021filip}      & \xmark     & FILIP300M~\cite{yao2021filip}, CC3M~\cite{sharma2018conceptual}, C12M~\cite{changpinyo2021conceptual}, YFCC100M~\cite{thomee2016yfcc100m} & 340M         & FILIP &  & -      &  Dual-Enc & ViT~\cite{dosovitskiy2020image}, GPT2~\cite{radford2019language}   & - & ICLR'22  \\

SLIP~\cite{mu2022slip}           & \cmark & YFCC15M~\cite{thomee2016yfcc100m, radford2019language} & 15M         & ITC, SimCLR & - & - & Dual-Enc & ViT~\cite{dosovitskiy2020image}, ViT-S~\cite{touvron2021training}, GPT2~\cite{radford2019language}  & \href{https://github.com/facebookresearch/SLIP}{Link} & ECCV'22 \\

FLIP~\cite{li2023scaling}        & \cmark & LAION400M~\cite{schuhmann2021laion} & 400M &   ITC & - & - & Dual-Enc &  ViT~\cite{dosovitskiy2020image}, Transformer~\cite{vaswani2017attention} & \href{https://github.com/facebookresearch/flip}{Link} & arXiv'23 \\

MaskCLIP~\cite{dong2023maskclip} & \cmark &  YFCC15M~\cite{thomee2016yfcc100m, radford2019language} & 15M & ITC & MLM & Distil & Dual-Enc & ViT~\cite{vaswani2017attention}, GPT2~\cite{radford2019language}&  \href{https://github.com/LightDXY/MaskCLIP}{Link} & CVPR'23 \\

CLIPA~\cite{li2023inverse} & \cmark & LAION-400M~\cite{schuhmann2021laion} & 400M & ITC & - & - & Dual-Enc & ViT~\cite{dosovitskiy2020image}, Transformer~\cite{vaswani2017attention} & \href{https://github.com/UCSC-VLAA/CLIPA}{Link} & arXiv'23 \\

CLIPAv2~\cite{li2023clipav2} & \cmark &  LAION-2B~\cite{schuhmann2021laion}, DataComp-1B~\cite{gadre2023datacomp}  & 3000M & ITC & - & - & Dual-Enc  & ViT~\cite{dosovitskiy2020image}, Transformer~\cite{vaswani2017attention}  & \href{https://github.com/UCSC-VLAA/CLIPA}{Link} & arXiv'23 \\

EVA~\cite{fang2023eva}  & \cmark & IN21K~\cite{fei2009imagenet}, CC12M~\cite{changpinyo2021conceptual}, CC3M~\cite{sharma2018conceptual}, O365~\cite{shao2019objects365}, COCO~\cite{lin2014microsoft}, ADE~\cite{zhou2019semantic} & 29.6M& ITC & - & - & Dual-Enc & ViT-G~\cite{zhai2022scaling}, BEiT-3~\cite{wang2022image} & \href{ https://github.com/baaivision/EVA}{Link} & CVPR'23\\

EVA-CLIP~\cite{sun2023eva} & \cmark & Merged-2B~\cite{sun2023eva} & 2000M    & ITC & - & - & Dual-Enc & ViT-G~\cite{zhai2022scaling}, BEiT-3~\cite{wang2022image} & \href{https://github.com/baaivision/EVA/tree/master/EVA-CLIP}{Link} & arXiv'23\\

EVA-02~\cite{fang2023eva2} & \cmark & Merged-2B~\cite{sun2023eva} & 2000M & ITC & - & - & Enc-Dec & TrV~\cite{fang2023eva2}, - &  \href{https://github.com/baaivision/EVA/tree/master/EVA-02}{Link} & arXiv'23   \\

OpenCLIP~\cite{cherti2023reproducible} & \cmark & LAION-400M~\cite{schuhmann2021laion}LAION-5B~\cite{schuhmann2022laion} & 5400M & ITC & - & - & Dual-Enc & ViT~\cite{dosovitskiy2020image}, GPT2~\cite{radford2019language} & \href{https://github.com/mlfoundations/open_clip}{Link} & CVPR'23\\

\cmidrule{2-11}
CRIS~\cite{wang2022cris}  & \cmark & RefCOCO, RefCOCO+~\cite{kazemzadeh2014referitgame}, G-Ref~\cite{nagaraja2016modeling} & 0.4M & TPC & - & - & Fusion & ResNet~\cite{he2016deep}, GPT2~\cite{radford2019language}  & \href{https://github.com/DerrickWang005/CRIS.pytorch}{Link} & CVPR'22  \\

MaskCLIP~\cite{zhou2022maskclip}  & \cmark & PASCAL VOC12~\cite{everingham2015pascal}, COCO Stuff~\cite{caesar2018coco}, PASCAL Context~\cite{mottaghi2014role} & 0.17M & - 
& - & Task & Dual-Enc & ResNet~\cite{he2016deep}, ViT~\cite{dosovitskiy2020image}, GPT2~\cite{radford2019language} & \href{https://github.com/chongzhou96/MaskCLIP}{Link} & ECCV'22 \\ 

GLIP~\cite{li2022grounded}  & \cmark & GoldG~\cite{kamath2021mdetr} (Flicker30K~\cite{young2014image}, VG~\cite{krishna2017visual}, GQA~\cite{hudson2019gqa}), OI~\cite{krasin2017openimages},  O365~\cite{shao2019objects365}, IN-Boxes~\cite{krizhevsky2012imagenet}, CC12M~\cite{changpinyo2021conceptual}, SBU~\cite{ordonez2011im2text}, Cap24M~\cite{li2022grounded}  & - & RWA & & & Fusion & ViT~\cite{dosovitskiy2020image}, GPT2~\cite{radford2019language} & \href{https://github.com/DerrickWang005/CRIS.pytorch}{Link}&  CVPR'22 \\

G-DINO~\cite{liu2023grounding} & \cmark & O365~\cite{shao2019objects365},OI~\cite{krasin2017openimages}, GoldG~\cite{kamath2021mdetr} (Flicker30K~\cite{young2014image},  VG~\cite{krishna2017visual}, GQA~\cite{hudson2019gqa}), Cap4M~\cite{li2022grounded}, COCO~\cite{lin2014microsoft}, RefCOCO~\cite{kazemzadeh2014referitgame} & - & GLIP & - & Task & Fusion & Swin-\{T, L\}~\cite{liu2021swin}, BERT-base~\cite{lu2019vilbert} &\href{https://github.com/IDEA-Research/GroundingDINO}{Link} & arXiv'23\\

OWL-ViT~\cite{minderer2022simple}  & \cmark & OI~\cite{krasin2017openimages}, O365~\cite{shao2019objects365}, VG~\cite{krishna2017visual} & 2M & - & - & DETR & Dual-Enc & ViT~\cite{dosovitskiy2020image}, Transformer~\cite{vaswani2017attention} & \href{https://github.com/google-research/scenic/tree/main/scenic/projects/owl_vit}{Link} & ECCV'22\\

ViLD~\cite{gu2021open}   & \cmark & LVIS~\cite{gupta2019lvis}, COCO~\cite{lin2014microsoft} &- & & & Task & Dual-Enc & Mask-RCNN~\cite{he2017mask}, FPN~\cite{lin2017feature}, CLIP-GPT2~\cite{radford2021learning} &\href{https://github.com/tensorflow/tpu/tree/master/models/official/detection/projects/vild}{Link} & ICLR'22 \\

GroupViT~\cite{xu2022groupvit}  & \cmark & YFCC~\cite{thomee2016yfcc100m}, CC12M~\cite{changpinyo2021conceptual} & 26M & ITC, MITC & - & - & Dual-Enc & ViT~\cite{vaswani2017attention, touvron2021training}, GPT2~\cite{radford2019language} &  \href{https://github.com/NVlabs/GroupViT}{Link} & CVPR'22 \\

OpenSeg~\cite{ghiasi2021scaling} & \cmark & COCO~\cite{lin2014microsoft}, LocalNarr~\cite{pont2020connecting} & 0.77M & & & Task & Dual-Enc& EffNet-B7~\cite{tan2019efficientnet}, ResNet~\cite{he2016deep}, BERT-Large~\cite{devlin2018bert} & \href{https://github.com/tensorflow/tpu/tree/641c1ac6e26ed788327b973582cbfa297d7d31e7/models/official/detection/projects/openseg}{Link} & ECCV'22\\

\midrule

Frozen~\cite{tsimpoukelli2021multimodal} & \cmark & CC3M~\cite{sharma2018conceptual} & 3M & - & Cap & -  & Enc-Dec & NF-ResNet~\cite{brock2021high}, GPT2~\cite{radford2019language} &  - & NeurIPS'21\\

Flamingo~\cite{alayrac2022flamingo}  & \xmark & M3W~\cite{alayrac2022flamingo} & 43M &  - & Flamingo & -  & AdaptedLLm & NF-ResNet~\cite{brock2021high}, Chinchilla~\cite{hoffmann2022training} & - & NeurIPS'22 \\

OpenFlamingo~\cite{anas_awadalla_2023_7733589} & \cmark & LAION2B~\cite{schuhmann2022laion}, MMC4~\cite{zhu2023multimodal} & 2571M & - & Flamingo & - & AdaptedLLm & CLIPL-ViT~\cite{radford2019language}, LLMs  & \href{https://github.com/mlfoundations/open_flamingo}{Link} & github'23 \\

{MetaLM}~\cite{hao2022language}  & \cmark & CC~\cite{sharma2018conceptual}, VG~\cite{wu2023visual}, COCO Captions~\cite{chen2015microsoft}, SBU~\cite{ordonez2011im2text}& 10M & - & SemiCasualLM & - & Enc-Dec &  ViT~\cite{dosovitskiy2020image}, GPT2~\cite{radford2019language} & - &arXiv'23\\

{KOSMOS-1}~\cite{huang2023language} & \cmark & LAION-\{400M, 2B\}~\cite{schuhmann2021laion, schuhmann2022laion}, COYO-700M~\cite{kakaobrain2022coyo-700m}, CC3M~\cite{sharma2018conceptual}, CC12M~\cite{changpinyo2021conceptual}& 3115M & - & SemiCasualLM & - & AdaptedLLm &  CLIP-ViT~\cite{radford2021learning}, MAGNETO~\cite{wang2022foundation} & - & arXiv'23\\

{KOSMOS-v2}\cite{peng2023kosmos2}   & - & GRIT~\cite{peng2023kosmos2}, LLaVA-Instruct~\cite{liu2023llava} & 115M & - & SemiCasualLM & - & AdaptedLLm  & CLIP-ViT~\cite{radford2021learning}, MAGNETO~\cite{wang2022foundation} & \href{https://github.com/microsoft/unilm/tree/master/kosmos-2}{Link} & arXiv'23\\

SimVLM~\cite{wang2021simvlm} & \cmark & ALIGN1.8B~\cite{jia2021scaling} & 1800M &  & PrefixLM & -   & Enc-Dec & ViT~\cite{dosovitskiy2020image}, BERT~\cite{devlin2018bert}&  - & ICLR'22 \\
MaskVLM~\cite{kuwon2022masked} & \cmark & CC, SBU~\cite{ordonez2011im2text}, VG~\cite{krishna2017visual}, COCO-C & 4M & ITC, ITM & MVLM & - & Fusion & ViT~\cite{dosovitskiy2020image}, RoBERTa~\cite{liu2019roberta} &  - & ICLR23 \\
mPLUG-OWL~\cite{ye2023mplugowl} & \cmark & LAION-400M~\cite{schuhmann2021laion}, COYO-700M~\cite{kakaobrain2022coyo-700m}, CC~\cite{sharma2018conceptual}, COCO~\cite{chen2015microsoft} & 1100+M & - & LM & - & AdaptedLLm & CLIP-ViT~\cite{radford2021learning}, LLaMA-7B~\cite{touvron2023llama} & \href{https://github.com/X-PLUG/mPLUG-Owl}{Link} & arXiv'23  \\

CapPa~\cite{tschannen2023image} & \xmark & WebLI~\cite{chen2022pali} & 12B & - & Cap, CapPa & - & Dual-Enc & &  - & arXiv'23  \\
\midrule
UNITER~\cite{chen2020uniter}  & \cmark & COCO~\cite{lin2014microsoft}, VG~\cite{krishna2017visual}, CC3M~\cite{sharma2018conceptual}, SBU~\cite{ordonez2011im2text} & 9.5M & ITM & MLM & - & Fusion  & Faster-RCNN~\cite{ren2015faster}, BERT~\cite{devlin2018bert} & \href{https://github.com/ChenRocks/UNITER}{Link} & ECCV'20 \\

Pixel2Seqv2~\cite{chen2022unified} & \cmark & COCO~\cite{lin2014microsoft} & 0.12M & - & VLM & - &   Enc-Dec & ViT-B~\cite{dosovitskiy2020image}, Transformer~\cite{vaswani2017attention} &  - & NeurIPS'22\\

VL-x ~\cite{cho2021unifying} & \cmark & COCO~\cite{lin2014microsoft, chen2015microsoft}, VG~\cite{krishna2017visual}, VQAv2~\cite{goyal2016making}, GQA~\cite{hudson2019gqa}, Visual7W~\cite{zhu2016visual7w, tan2019lxmert}& 9.18M & ITM & MLM & Task & Enc-Dec &  BEiTv2~\cite{peng2022beit}, RoBERTa~\cite{liu2019roberta} &  &ICML'21\\

CoCa~\cite{yu2022coca}        & \xmark & JFT3B~\cite{zhai2022scaling}, ALIGN~\cite{jia2021scaling} & 4800M   & NSL  & Cap & - & Fusion & ViT~\cite{dosovitskiy2020image}, Transformer~\cite{vaswani2017attention} & - & TMLR'23 \\

FLAVA~\cite{singh2022flava}   & \cmark & PMD (COCO~\cite{lin2014microsoft}, SBU~\cite{ordonez2011im2text}, LocaNarr~\cite{pont2020connecting}, VG~\cite{krishna2017visual}, WikiIT~\cite{srinivasan2021wit}, CC12M~\cite{changpinyo2021conceptual}, RedCaps~\cite{DBLP:conf/nips/DesaiKA021}, YFCC100M~\cite{thomee2016yfcc100m}) & 70M & ITC,ITM & MMM,MIM,MLM & - & Fusion & ViT~\cite{dosovitskiy2020image}, Transformer~\cite{vaswani2017attention} & \href{https://github.com/facebookresearch/multimodal/tree/main/examples/flava}{Link}&  CVPR'22  \\

PaLI~\cite{chen2022pali} & \xmark & WebLI~\cite{chen2022pali}, CC3M~\cite{sharma2018conceptual}+   & 1600M & - & - & - & Enc-Dc & ViT-e~\cite{chen2022pali}, mT5~\cite{xue2020mt5} & - & arXiv'23 \\

BLIP~\cite{li2022blip} & \cmark & COCO~\cite{lin2014microsoft}, VG~\cite{wu2023visual}, CC3M~\cite{sharma2018conceptual},C12M~\cite{changpinyo2021conceptual}, SBU~\cite{ordonez2011im2text}, LAION~\cite{schuhmann2021laion}, WebCapFit~\cite{li2022blip} & 129M & ITC,ITM &  LM & - & Fusion & ViT~\cite{dosovitskiy2020image}, BERT~\cite{devlin2018bert} & \href{https://github.com/salesforce/BLIP}{Link}& arXiv'22 \\

BridgeTower~\cite{xu2022bridge} & \cmark & CC3M~\cite{sharma2018conceptual}, SBU~\cite{ordonez2011im2text}, COCO~\cite{chen2015microsoft}, VG~\cite{krishna2017visual} & 4M & ITM & MLM & - & Fusion & CLIP-ViT~\cite{radford2021learning}, RoBERTa~\cite{liu2019roberta} &  
\href{https://github.com/microsoft/BridgeTower}{Link} & AAAI'23\\

X-FM~\cite{zhang2023toward} & \cmark & COCO~\cite{lin2014microsoft}, SBU~\cite{ordonez2011im2text}, RefCOCO~\cite{yu2016modeling},  VG~\cite{wu2023visual}, CC3M~\cite{sharma2018conceptual}, O365~\cite{shao2019objects365}, OI~\cite{krasin2017openimages}, IN21K~\cite{fei2009imagenet}, C4~\cite{raffel2020exploring}, RoBERTa Corpus~\cite{liu2019roberta} & 20M & ITC,ITM & MLM,MIM,IMLM & BBP & Fusion & BEiTv2~\cite{wang2022image}, RoBERTa~\cite{liu2019roberta} &  \href{https://github.com/zhangxinsong-nlp/XFM}{Link} &  arXiv'23 \\

BLIP-2~\cite{li2023blip} & \cmark & COCO~\cite{lin2014microsoft}, VG~\cite{wu2023visual}, CC3M~\cite{sharma2018conceptual}, C12M~\cite{changpinyo2021conceptual},SBU~\cite{ordonez2011im2text}, LAION~\cite{schuhmann2021laion} & 129M & ITC,ITM & ITG & - & AdaptedLLm & CLIP-ViT~\cite{radford2021learning}, EVA-CLIP~\cite{fang2023eva}, OPT~\cite{zhang2022opt}, FlanT5~\cite{chung2022scaling}  & \href{
https://github.com/salesforce/LAVIS/tree/main/projects/blip2}{Link}& arXiv'23\\

Instruct-BLIP~\cite{dai2023instructblip} & \cmark &COCO~\cite{lin2014microsoft}, WebCapFit~\cite{li2022blip}, NoCaps~\cite{agrawal2019nocaps}, Flicker30k~\cite{young2014image}, TextCaps~\cite{sidorov2020textcaps}, VQAv2~\cite{goyal2016making}, VizWiz~\cite{gurari2018vizwiz}, GQA~\cite{hudson2019gqa},  VSR, IconQA~\cite{lu2021iconqa}, OKVQA~\cite{marino2019ok}, A-OKVQA~\cite{schwenk2022okvqa}, SciQA~\cite{lu2022learn}, VD~\cite{das2017visual}, OCRVQA~\cite{mishra2019ocr}, TVQA~\cite{singh2019towards}, HM~\cite{kiela2020hateful}, LI~\cite{liu2023visual}, MQA, MSQA~\cite{xu2017video}, iVQA~\cite{yang2021just} & - & & LM & - &  AdaptedLLm & EVA-ViT~\cite{sun2023eva}, LLM~\cite{vicuna2023, chung2022scaling}\footnote{FlanT5, Vicuna} & \href{https://github.com/salesforce/LAVIS/tree/main/projects/instructblip}{Link} & arXiv'23\\

TaCA~\cite{zhang2023taca} & \cmark & LAION-400M~\cite{schuhmann2021laion} & 400M & ITC &  - & Distil & - & CLIP-ViT~\cite{radford2019language}, GPT2~\cite{radford2019language} &  \href{https://github.com/TencentARC/TaCA}{Link} & arXiv'23\\

VPGTrans~\cite{zhang2023transfer} & \cmark & COCO-C, SBU & 1.4M & - & - & Sim & - & BLIP-2 &    \href{https://github.com/VPGTrans/VPGTrans}{Link} & arXiv'23\\

FIBER~\cite{dou2022coarse} & \cmark & COCO~\cite{lin2014microsoft}, CC, SBU~\cite{ordonez2011im2text}, VG~\cite{wu2023visual}, O365~\cite{shao2019objects365} & 4.8M & ITC,ITM & MLM & Task & Fusion & Swin~\cite{liu2021swin}, RoBERTa~\cite{liu2019roberta} &   \href{https://github.com/microsoft/FIBER}{Link} & NeurIPS'22\\

UniDetector~\cite{wang2023detecting} & \cmark & COCO~\cite{lin2014microsoft}, O365~\cite{shao2019objects365}, OI~\cite{krasin2017openimages} & - & - & - & Task & Dual-Enc & RegionCLIP~\cite{zhong2022regionclip} & \href{https://github.com/zhenyuw16/UniDetector}{Link} &  CVPR'23\\

X-Decoder~\cite{zou2022xdecoder}  & \cmark & COCO17, CC, SBU~\cite{ordonez2011im2text}, VG~\cite{wu2023visual} COCO-C & 4.07M & ITC & Cap & Task & Fusion & Focal-T~\cite{yang2022focal}, DaViT~\cite{ding2022davit}, GPT2~\cite{radford2019language}  & \href{https://github.com/microsoft/X-Decoder/tree/main}{Link} & CVPR'23 \\

GLIPv2~\cite{zhang2022glipv2} &  \cmark & FiveODs, GoldG, CC15M, SBU~\cite{ordonez2011im2text}, COCO, LVIS & - & RWC & - & Task & Fusion & Swin-T~\cite{liu2021swin}, Transformer~\cite{vaswani2017attention}  & \href{https://github.com/microsoft/GLIP}{Link}& NeurIPS'22\\
\bottomrule
\end{tabular}}
    \caption{An overview of Textually Prompted Models. We exhibit different facets of these models that contrast them, including pre-training datasets and their sizes, pre-training objectives, architecture, publication venues, and online information. More details are discussed in Section~\ref{sec:background}.}
    \label{tab:textually_prompted_models}
\end{table*}

Traditionally, vision-language models have primarily been employed for tasks that necessitate the joint understanding of both visual and textual modalities. However, with the remarkable performance exhibited by CLIP, language supervision-based models have gained significant prominence and have become the mainstream approach. In this section, we focus on exploring methods that rely on language as their primary source of supervision. These textually prompted models are broadly categorized into three main types based on their training objectives: contrastive, generative, and hybrid approaches. We discuss contrastive-based methods in Sec.~\ref{sec:contrastive}, generative-based methods in Sec.~\ref{sec:generative}, and combination methods in Sec.~\ref{sec:hybrid}. We provide an overview of these methods in Tab.~\ref{tab:textually_prompted_models}. We also exhibit a comparison of these models for a representative set of tasks in Tab.~\ref{tab:results_textually_prompted}.

\subsection{Contrastive Learning (CL)}
\label{sec:contrastive}
\begin{figure*}
    \centering
    \includegraphics[width=1\textwidth]{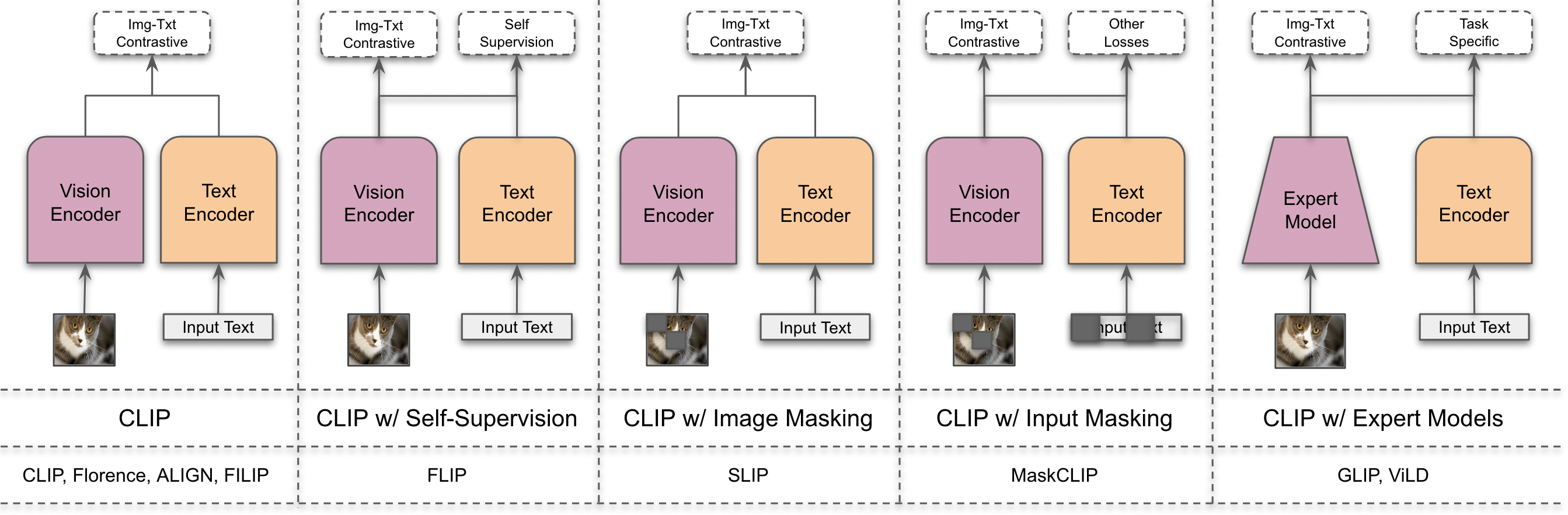}
    \vspace{-20pt}
    \caption{An overview of CLIP and its variants. Several works after CLIP have investigated the efficacy of different variants including new losses, image-based self-supervision, masking of inputs, and use of expert models. Here, expert model means a model that is trained for a specific task (e.g., ViLD~\cite{gu2021open} used a Feature Pyramid Network~\cite{lin2017feature} as an object detection expert model.}
    \label{fig:clip_variants}
    
\end{figure*}

SOTA computer vision models are trained to predict a set of pre-determined categories which restricts their generalization and usability. Conventionally, most deep learning methods have used supervised pre-training such as training on ImageNet~\cite{fei2009imagenet}, and weak supervision such as training on hash-tags of images~\cite{mahajan2018exploring}. \citet{radford2021learning} argued for learning perception from the visual concepts present in the natural language and proposed Contrastive Language Image Pre-training (CLIP). In this section, we discuss CLIP and subsequent contrastive-based approaches. We have divided them into two parts: contrastive approaches for general purpose models (Sec.~\ref{sec:cl_general}) and approaches for visual grounding foundational models (Sec.~\ref{sec:cl_grounding}). We illustrate CLIP architecture and its main variants in Fig. \ref{fig:clip_variants}.

\subsubsection{CL for General Purpose Foundational Models}
\label{sec:cl_general}
In this section, we explain contrastive methods that aim to train general-purpose vision-language foundational models. The mainstream traction of these approaches started with CLIP~\cite{radford2021learning}, however, many subsequent efforts have provided better ways to utilize datasets, proposed modified architectures and training methods, expanded its utility, reproduced it, and studied its properties and scaling laws. We describe these methods here.

\noindent\textbullet \textbf{Methods Solely based on CL.}
\methodname{CLIP~\cite{radford2021learning}: }\citet{radford2021learning} proposed jointly training an image and text encoder on the contrastive pre-training task of the correct pairing of images and their captions in a batch. 
The \textbf{CLIP} model consists of an image encoder (a ViT or a scaled CNN) and a text encoder (a GPT-like transformer~\cite{brown2020language}). These encoders produce a multi-modal embedding space for $N$ image-text pairs. 
The CLIP is trained to minimize the cosine similarity of embeddings of $N$ correct image-text pairs and maximize the cosine similarity of embeddings of $N^2-N$ incorrect pairs via a symmetric cross-entropy loss.
One main motivation of CLIP-framework is the scale of natural language supervision data. To train models at scale, authors also curated a 400 million image-text pairs dataset from the internet. The CLIP framework shows excellent performance when trained on such large-scale datasets. CLIP shows good zero-shot generalization, has significantly higher robustness to natural and synthetic distribution shifts, and works well with linear probe-based fine-tuning. 

\methodname{ALIGN~\cite{li2021align}:} Visual-language dataset used by \citet{radford2021learning} require non-trivial and computationally expensive pre-processing and cleaning, thereby limiting the scale of the dataset. Instead of applying these pre-processing steps, Jia et al. in \textbf{ALIGN}~\cite{jia2021scaling} collected one billion noisy image-caption datasets curated from Conceptual Captions Dataset~\cite{sharma2018conceptual}. They trained dual encoder-based architecture with CLIP-like normalized contrastive objectives on this dataset. To align visual and language embeddings, cosine similarity of image-text embeddings are optimized via. normalized softmax loss~\cite{zhai2018classification}. The authors showed that the scale of the dataset can make up for the noisy nature of it. The resulting aligned image-text representations show excellent performance for cross-modal matching/retrieval tasks and zero-shot classification. 

\methodname{Florence~\cite{yuan2021florence}: } \citet{yuan2021florence} argued that a truly foundational model should work for Space-Time-Modality space. Specifically, a foundational model should be able to handle representation from coarse to fine (Space), static to dynamic (Time), and from RGB to multi-modalities (Modality). To achieve this level of generalizability, they introduced \textbf{Florence} model that starts with CLIP-like pre-training on the large curated dataset and uses improved contrastive objective, and efficient training. A pre-trained model is then extended to have three different adapter heads for each space. The Dynamic DETR-based adapter learns representation for fine-grained dense tasks with large-scale object detection datasets. Similarly, a METER~\cite{dou2022empirical} head is used for vision language representation, and CSwin~\cite{dong2022cswin} is used for video-based understanding. This framework results in a foundational model that generalizes across domains.

Most vision-language methods are focused on language supervision and have overlooked the role of the visual part. \methodname{SLIP~\cite{mu2022slip}: } \citet{mu2022slip} investigated whether image-based self-supervised learning can help language supervision frameworks. To this end, authors proposed \textbf{SLIP} that adds an adaptation of SimCLR~\cite{chen2020simple, chen2020big} loss for self-supervision based on different views or augmentation of the input image. The authors trained the CLIP-like model on the YFCC15M dataset and showed that SLIP performs better compared with either language supervision or self-supervision alone on a battery of tasks including zero-shot and linear probe-based methods. 

\methodname{WenLan~\cite{huo2021wenlan}: }Most text-image-based methods assume a strong semantic correlation between the pair. However, web-scale data is littered with pairs that have weak correlations (e.g., captions that do not reflect images accurately). \citet{huo2021wenlan} propose \textbf{WenLan} to solve this by using two-tower architecture and cross-modal contrastive learning based on MoCo~\cite{he2020momentum}, which can leverage more negative samples in limited GPU resources. This strategy leverages both negative and positive examples and text-to-image and image-to-text-based contrastive losses. This results in a better model that is also efficient to train and shows improved performance on many tasks. They also curated the first large-scale Chinese image-text dataset with 500 million data points. They trained their proposed model to solve Chinese-language tasks and demonstrated its superior zero-shot ability. 

\methodname{FILIP~\cite{yao2021filip}:} CLIP-like methods use separate encoders for each modality which makes them inference efficient as each encoder can be decoupled and pre-computed representations can be utilized. However, such models rely solely on global features for cross-modal interaction which makes it hard for them to capture finer-level information between modalities. \citet{yao2021filip} proposed a cross-modal late interaction approach to model the token-wise cross-modal interaction which helps capture fine-grained semantic alignment. The proposed \textbf{FILIP} (Fine-grained Interactive Language Image Pre-training) loss maximizes the similarity between token-wise visual and text embeddings. Specifically, similarity for each input visual token with all text-tokens is calculated, and maximum similarity is used. Similarly, the maximum similarity of each text token is also calculated. Then, a simple average is used to calculate the overall loss. This means each image token's closest text token is used. Similarly, each text's closest image patch is also used. This helps in modeling fine-grained interaction between the two modalities without sacrificing the inference-efficient nature of CLIP. Moreover, the authors also collected 340M large image-text pairs to train their model. Their method outperforms CLIP and other methods on zero-shot classification as well as for image-text retrieval tasks.

\noindent\textbullet \textbf{Masked Contrastive Learning. }
\methodname{FLIP~\cite{li2023scaling}: }Inspired by the Masked Auto Encoders~\cite{he2022masked}, \citet{li2023scaling} presented an efficient alternative of CLIP called \textbf{FLIP} which masks 50-75\% of input pixels in CLIP training. This masking scheme reduces computation by 2-4$\times$, allows 2-4$\times$ larger batches, and improves accuracy. FLIP is $> 3\times$ faster to reach the same accuracy as CLIP. Compared with the CLIP baseline, their method can save ~1800 TPU days. Based on this faster method, they also studied the scaling in CLIP across models, datasets size, and training length. 

\methodname{MaskCLIP~\cite{dong2023maskclip}: } \citet{dong2023maskclip} argued that the language description of an image can not express complete information as an image is a continuous and fine-grained signal. To fully leverage images in contrastive vision-language training, they proposed \textbf{MaskCLIP} that randomly masks the input image along with mean teacher-based self-distillation~\cite{tarvainen2017mean} to learn local semantic features. Specifically, the representation of the whole image and masked image are obtained from the mean teacher and student respectively, and cross-entropy loss between the two representations is minimized. Similarly, BERT~\cite{lu2019vilbert} pre-training is used in the language encoder. These two modifications in the contrastive learning framework help the model learn local and fine-grained semantics. MaskCLIP significantly improves CLIP in zero-shot, linear probe, and fine-tuning settings on several vision datasets.

\methodname{EVA-CLIP~\cite{sun2023eva}: } While the previous approaches in this section mainly focus on the efficiency aspect of CLIP via masking, \textbf{EVA-CLIP} addresses instability and optimization efficiency together with masking visual inputs. 
Specifically, \citet{sun2023eva} offered solutions to improve the stability of training and reduce computational costs, including improved initialization, better optimizer, and random masking of images~\cite{li2023scaling}. Their efficient solution-based model, EVA-CLIP, performs better and is trained on the version of datasets that are curated from open-source resources.
\methodname{EVA~\cite{fang2023eva}: }\citet{fang2023eva} supplemented this effort to scale models. They masked out image-text inputs along with CLIP loss to scale the model to 1 billion parameters. Their scaled model, named EVA, performs strongly on several downstream tasks, including COCO, LVIS, ImageNet1k, etc.

\noindent\textbullet \textbf{Scaling and Reproducing CLIP. }
OpenAI released pre-trained weights and code for CLIP. However, they did not release the training mechanism and dataset which limits the ability to study it. To increase the accessibility, several subsequent works have open-sourced large-scale image-text datasets, reproduced CLIP, and studied its properties. 
\methodname{LAION-400M dataset~\cite{schuhmann2021laion}: } The excellent performance of CLIP hinges on the large-scale image-text dataset, which is not publicly available. To address this problem, \citet{schuhmann2021laion} released LAION-400M, an image-text dataset consisting of 400 million data points curated after filtering common crawl. 
\methodname{LAION-5B dataset~\cite{schuhmann2022laion}: } \citet{schuhmann2022laion} further scaled it up and released a multilingual, multi-modal dataset called LAION-5B which contains 5.8 billion data points curated from Common Crawl after filtering through existing CLIP model. 
\methodname{OpenCLIP~\cite{ilharco_gabriel_2021_5143773, cherti2023reproducible}: } Utilizing large-scale LAION datasets~\cite{schuhmann2021laion, schuhmann2022laion}, \textbf{Open-CLIP} \cite{ilharco_gabriel_2021_5143773} trained and reproduced CLIP training experiments and studied its properties. 
\citet{cherti2023reproducible} supplemented this open-source effort by studying the scaling laws of CLIP. The trained OpenCLIP on LAION-5B~\cite{schuhmann2022laion} dataset demonstrated consistent improvements in performance as data, model, and compute are scaled. They also observed some divergence of scaling from the OpenAI's CLIP~\cite{radford2021learning} and postulated that the difference arises because of the difference in training distributions. 

\methodname{CLIPA~\cite{li2023inverse}: } It is well known that the performance of CLIP scales with model and dataset sizes~\cite{radford2021learning, cherti2023reproducible}. \citet{li2023inverse} revealed a surprising finding: larger image-text models allow the use of smaller token sizes during the training without significant sacrifice in accuracy. Based on this finding, called \emph{inverse scaling law}, they introduced a new efficient training recipe and  CLIP-like model trained on academic-scale resources and dubbed it \textbf{CLIPA}. CLIPA can achieve 63.2\%, 67.8\%, and 6.9.3\% zero-shot ImageNet accuracy in 2, 3, and 4 days of training on 8 A100 GPUs, respectively. 
\methodname{CLIPv2~\cite{li2023clipav2}: } Building on the \emph{inverse scaling law} observed by CLIPA~\cite{li2023inverse}, \citet{li2023clipav2} trained CLIP-like model on a large scale with significantly less computational budget and training costs. With their large-scale training, they demonstrated two interesting results. First, they showed that the inverse scaling law is also applicable to fine-tuning: models can be fine-tuned on fewer input tokens. Second, larger models exhibit a smaller performance drop compared with smaller models when fine-tuned with the same number of input tokens. Their trained CLIPA model achieves 69.3\% zero-shot ImageNet classification accuracy in 4 days of training on 8 A100 GPUs. Several works have explored CLIP and contrastive methods from different prespectives~\cite{liu2023remoteclip, liu2023stone, kim2023cream}

\subsubsection{CL for Visual Grounding Foundational Models}
\label{sec:cl_grounding}
\begin{figure}
    \centering
    \includegraphics[width=\columnwidth]{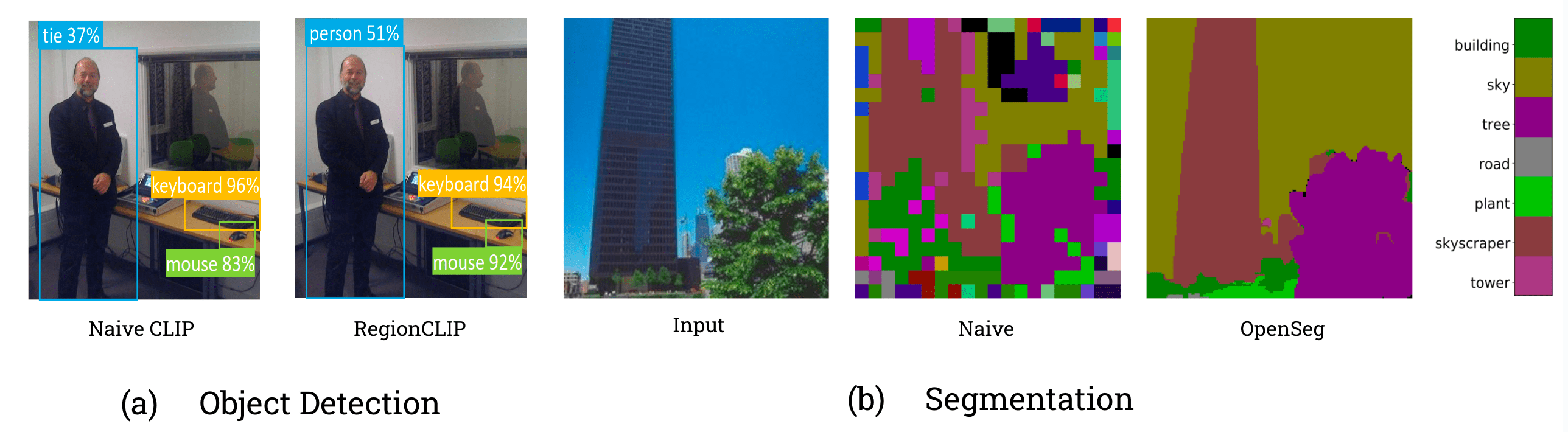}
    \vspace{-2em}
    \caption{Naive application of CLIP does not work well for localization tasks. Failure case for object detection is taken from \citet{zhong2022regionclip} and for segmentation from \citet{ghiasi2021scaling}. }
    \label{fig:clip_error_localization}
\end{figure}

CLIP and its variants have shown impressive performance for tasks that require global information (e.g., classification and image-text retrieval). However, they perform poorly on localization tasks that need fine-grained, pixel, and region-level information. 
We illustrate two failure cases obtained from \citet{zhong2022regionclip,ghiasi2021scaling} in Fig.~\ref{fig:clip_error_localization}. In this section, we discuss foundational models that are designed to leverage contrastive learning for visual grounding tasks. 

\noindent\textbullet \textbf{CLIP-adaptation for Grounding: }
\methodname{MaskCLIP~\cite{zhou2022maskclip}: } 
MaskCLIP by \citet{dong2023maskclip} model earlier investigated masked self-distillation for contrastive learning. Different f that, \textbf{MaskCLIP} by \citet{zhou2022maskclip} proposes to use the CLIP model for dense prediction with minimal changes. To this end, they proposed to extract dense features from the vision encoder and use text embeddings for classification. To further enhance dense predictions, they also proposed to train a backbone for classification. Their method shows a reasonable performance of CLIP for localization tasks. 

\methodname{RegionCLIP \cite{zhong2022regionclip}: } 
\citet{zhong2022regionclip} proposed \textbf{RegionCLIP} that extends CLIP to explicitly align image regions and their textual descriptions for object detection. Its training consists of three phases: a CLIP-based image-text pre-training, a CLIP-like region-text contrastive training, and object-detection-specific fine-tuning. Since large-scale region-description datasets are not widely available, authors utilized region-class names, prompt templates, and a pre-trained CLIP to bootstrap a dataset. Specifically, they used a pre-trained teacher encoder to extract regions. All class labels are converted into phrases following simple prompt templates and a teacher language encoder is utilized to get corresponding embeddings. The matching score between image features and class embeddings is calculated and the highest score pair is used as a pseudo-region-text pair. The authors pre-trained their dual-encoder-based model on these pseudo-region-description pairs. Finally, they also proposed a simple fine-tuning method to mitigate the noisy nature of region-text pairs. For task-specific finetuning, the visual encoder is used as a base network initiated from a pre-trained ViT. An off-the-shelf region proposal network (RPN) localizes objects and the language encoder's embeddings are used to obtain object categories. RegionCLIP has zero-shot capabilities and when transferred, established a new SOTA for open-vocabulary object detection. 

\methodname{CRIS \cite{wang2022cris}: } \citet{wang2022cris} extended CLIP for referring image segmentation task~\cite{hu2016segmentation, ye2019cross} and proposed CLIP-Driven Referring Image Segmentation (\textbf{CRIS}). Referring image segmentation task aims to segment a region of the image based on an input text prompt~\cite{hu2016segmentation}, and hence a natural fit for a CLIP-like framework. 
However, CLIP is not designed for learning pixel-level information as it is focused on global features. 
\citet{wang2022cris} proposed two modifications in the CLIP framework to make it learn pixel-level information. First, a visual-language decoder is introduced to capture long-range dependencies. Second, a text-to-pixel contrastive loss is introduced to align textual features with the corresponding pixel-level features. CRIS outperforms previous SOTA for three referring image segmentation tasks.  

\noindent\textbullet \textbf{Direct localized Visual-Semantic Alignment: } Instead of adapting CLIP for grounding tasks, some works leveraged strong pre-trained specialized models and modified them to add language-vision modeling through contrastive learning.
\methodname{GLIP \cite{li2021align}: } Phrase grounding is the task of identifying phrases in the input text for corresponding regions of an image. \citet{li2022grounded} argues that phrase grounding is a scalable and effective pretraining task for object detection, and hence reformulated the object detection task to phrase grounding. This provides benefits for both tasks: phrase grounding provides better visual concepts and object detection provides more annotations for bounding boxes. They proposed Grounded Language Image Pretraining (\textbf{GLIP}) which trains a dual vision-language encoder-based architecture with fusion layers on phrase-region dataset. To train models at scale, a pre-trained grounding model is applied to the image-text datasets to get phrase-region pseudo labels. To use GLIP models for object detection datasets, all the class names are combined in one sentence, and the model is prompted to output the correct class associated with a region. This simple scaling approach brings significant improvements for 14 downstream tasks and its fine-tuned version sets a new SOTA on the COCO dataset.

\methodname{Grunding DINO~\cite{liu2023grounding}: } Instead of extending CLIP framework, \citet{liu2023grounding} proposed to ground state-of-the-art transformer-based object detector DINO~\cite{caron2021emerging} with language pre-training for open-set generalization, hence named \textbf{Grounding-DINO}. To this end, they partitioned the closed-set object detector into three parts consisting of a backbone, a neck, and a head and fused language features at each level. A text and image backbone is utilized to extract multi-scale features that are fed to the neck. The text and image features produced by the neck are then used to create language-guided query selection. These cross-modal queries along with image and text features are fed to a cross-modality decoder which has image and text cross-attention and an FFN layer. The model is trained end-to-end with a contrastive loss between predicted objects and language tokens as well as task-specific losses such as L1 loss, Grounded Intersection over Union (GIOU) loss~\cite{rezatofighi2019generalized}, and focal loss~\citet{lin2017focal}. Grounding DINO outperforms GLIP and other competitors for closed-set, open-set, and referring object detection by significant margins. 

\methodname{OWL-ViT \cite{minderer2022simple}: } \citet{minderer2022simple} introduced a CLIP-based training recipe for open vocabulary object detection called \textbf{OWL-ViT}. Their proposed training method consists of two phases: a CLIP-like pre-training for learning image-level features and a fine-tuning stage for object-level features for open-vocabulary object detection. Their dual encoder architecture is similar to CLIP except for task-specific modifications. Specifically, the output of a ViT-based image encoder consists of a projection layer for classification embeddings and an MLP head for box predictions and respective probabilities. To enable open vocabulary detection, the language encoder produces text embeddings (queries) based on input prompts which can be different for each image. The visual encoder's role, then, is to predict the bounding box and probabilities with which the query is applied to it. This dual-encoder architecture is first trained with CLIP-like contrastive learning and then fine-tuned on object detection datasets with bipartite matching loss~\cite{carion2020end} adapted for long-tailed/open vocabulary object detection.

\methodname{OpenSeg \cite{ghiasi2021scaling}: } \citet{ghiasi2021scaling} argued that CLIP-like methods perform poorly for localization tasks because they do not group first and hence lose local information. To solve this issue, they propose \textbf{OpenSeg} that performs visual-semantic alignment after the grouping. Their method involves learning segmentation masks, visual-semantic alignment of these masks, and generating pseudo masks for large-scale pre-training. Their model represents an image with segmentation masks which enables segmentation-based weakly supervised learning along with region-word grounding. This type of training requires segmentation labels and therefore difficult to scale. To solve the scaling problem, the authors followed MuST~\cite{ghiasi2021multi} and first trained the model on segmentation data with only segmentation loss. This model is used to generate pseudo labels for image-text pairs. The openSeg-based model can generalize well for new datasets and outperforms previous SOTA on several benchmarks. 

\begin{figure*}
    \centering
    \includegraphics[width=0.9\textwidth]{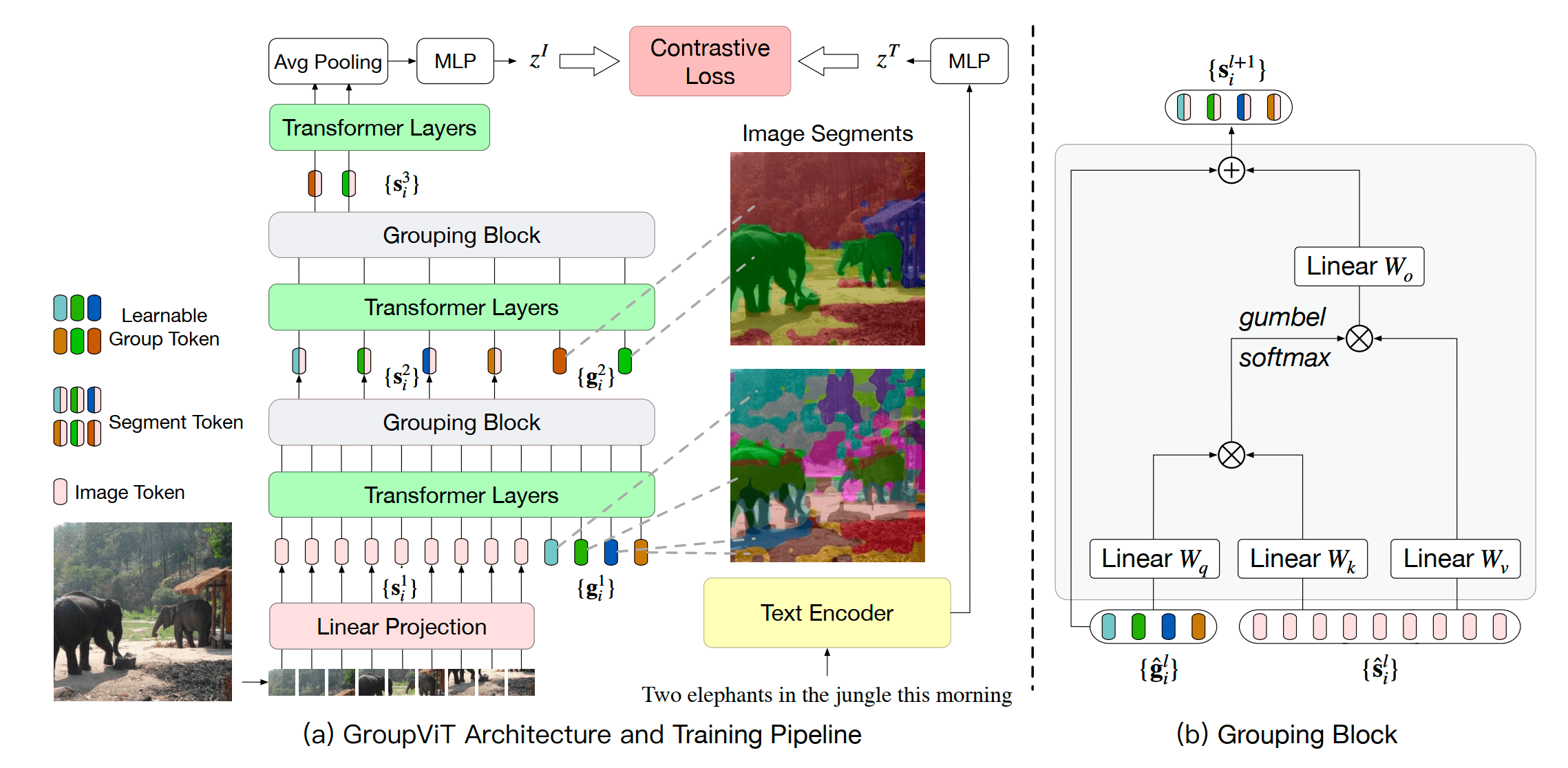}
    \caption{GroupViT~\cite{xu2022groupvit} consists of grouping blocks transformer layers. Image tokens and learnable group tokens are input to the first transformer layer and the output from the last transformer layer is average pooled and used in image-text-based contrastive learning.
The figure is taken from \citet{xu2022groupvit}. }
    \label{fig:groupvit}
\end{figure*}

\methodname{GroupViT \cite{xu2022groupvit}: } \citet{xu2022groupvit} proposed to leverage visual grouping mechanism~\cite{tu2005image, zhu2007stochastic} to get semantic segmentation with only language supervision. To this end, they proposed a hierarchical Grouping Vision Transformer (\textbf{GroupViT}) as an image encoder along with the standard CLIP-like language encoder. The proposed GroupViT has multiple grouping layers that learn to group regions of an image into progressively larger segments of similar visual concepts by learning segment tokens. Each stage also consists of transformer layers that aggregate segment tokens from smaller groups from previous stages to progressively larger segments. An overview of their architecture is shown in Fig.~\ref{fig:groupvit}. The GroupViT is trained with Image-Text contrastive (ITC) loss and multi-label contrastive loss with prompt engineering, which uses prompts to create multiple descriptions of a single image. For zero-shot segmentation, the segment token in the last layer corresponds to an arbitrarily shaped segment whose class can be determined by finding a class label that has maximum similarity with the segment token. GroupViT performs competitively compared to specialized SOTA methods without requiring any supervision. Similarly, ODISE \cite{xu2023open} is also an open vocabulary segmentation model that utilizes pre-trained diffusion features.

\subsection{Generative Learning}
\label{sec:generative}
Large Language Models (LLMs) have shown impressive zero and few-shot performance for NLP tasks. However, these LLMs lack vision modality and only recently multimodal models have been trained with vision and language modalities. Contrastive vision-language models have also shown good generalization capabilities but they can only address limited problems since they provide a similarity score between text and image. Here, we describe works that aim to equip LLMs with eyes to see the world by training them on vision-conditioned language generation tasks. 

\noindent\textbullet \textbf{In-context Learning with Multimodal Inputs: }
Large language models are excellent few-shot learners~\cite{brown2020language} but in their conventional form, they are blind to the visual modality. Here, we explain methods that aim to endow LLMs with visual modality using interleaved image-text data. 

\methodname{Frozen~\cite{tsimpoukelli2021multimodal}:}  \citet{tsimpoukelli2021multimodal} proposed \textbf{Frozen}, an efficient approach to add visual modality in the LLMs without updating their weights. Frozen consists of an image encoder that encodes input images to the word embedding space of LLMs such that these LLMs can generate image captions. To learn joint embeddings, LLM is kept frozen and the vision encoder is trained on captioning datasets with the task of the conditional generation of caption given an image.
Although Frozen is trained on single image-text pairs, it can work with an ordered set of multiple image-text pairs enabling it to do few-shot tasks. At inference, the LLM encoder and vision encoder are prompted with the ordered textual and visual prompts. The textual and visual embeddings are concatenated and fed to the decoder of the LLM, which generates a textual output autoregressively. Frozen has demonstrated few-shot visual-language capabilities across vision-language tasks.

\methodname{Flamingo~\cite{alayrac2022flamingo}:} Similar to Frozen, \citet{alayrac2022flamingo} also aimed to build models that can adapt to new tasks using only a few examples. For this purpose, they proposed a new family of \textbf{Flamingo} models that leverage fixed pre-trained vision and language models along with a Perceiver Resampler-based bridge. The Perceiver Resampler connects the visual encoder to LLM by producing a fixed number of visual tokens. These visual tokens conditions LLM's output using gated cross-attention dense blocks that are interleaved between LM's layers. These new layers provide an efficient way for the LLM to incorporate visual information and are trained with next-token prediction tasks conditioned on preceding text and a set of images or videos.

Flamingo models can handle large image and video inputs since Perceiver Resample converts varying size visual inputs to a few visual tokens. During inference, interleaved support examples,  (image, text) or (video, text), are followed by a query visual input fed to the model for computation. Flamingo shows excellent few-shot performance for several vision-language tasks, even surpassing state-of-the-art for fine-tuned models despite requiring significantly fewer annotated examples. 
\methodname{OpenFlamingo~\cite{anas_awadalla_2023_7733589}:} \citet{anas_awadalla_2023_7733589} aimed to build an open-source version of the Flamingo model called \textbf{OpenFlamingo}. They mostly followed Flamingo's original architecture but trained it on a new Multimodal C4 dataset and 10M samples from LAION-2B and released open-source checkpoints. Their models utilized LLaMA-7B and CLIP's visual encoder and achieve 80\% performance of the respective Flamingo model.

\noindent\textbullet \textbf{LLMs as a General Interface for other Modalities: }
\citet{hao2022language} proposed to utilize language models as a universal task layer and to dock other modalities to it through the pre-trained encoders. Here, we describe these methods.
\methodname{MetaLM~\cite{hao2022language}: }\citet{hao2022language} proposed \textbf{MetaLM}, a semi-casual model that consists of a unidirectional transformer decoder and multiple bi-directional encoders that are connected with the decoder through the connector layers. In this way, MetaLM enjoys excellent zero and few-shot capabilities of the casual language models~\cite{brown2020language} and better transferability of non-casual encoders~\cite{devlin2018bert,raffel2020exploring}. The authors propose to jointly train encoders and decoders on a new semi-casual language modeling objective which learns to generate the next word given the previous tokens and encoded representations. This joint framework inherits in-context learning, instruction following, and fine-tuning abilities. To understand the capabilities of MetaLM a multitude of experiments are performed. On NLP tasks, it outperforms GPT on a multitude of multi-task fine-tuning, single-task fine-tuning, instruction-tuned zero-shot, and in-context learning. Similarly, zero-shot generalization, in-context learning, and fine-tuning capabilities on two vision-language tasks (captioning and VQA) show MetaLM's superior performance compared with previous strong baselines.

\methodname{KOSMOS-1~\cite{huang2023language}: }Following MetaLM~\cite{hao2022language}, \citet{huang2023language} aimed to align perception with LLMs to create models that can work with multiple modalities. The proposed model, \textbf{KOSMOS-1}, consists of a Magneto-based LLM~\cite{wang2022foundation} as a general interface and xPos~\cite{sun2022length} encoders to encode different modalities. This model is trained on web-scale data consisting of text corpus, image-caption pairs, and interleaved image-captions pairs to generate the next token given context. To further align it with the human interface, training data also consists of several language-only instruction tuning datasets that are also dealt with as language modeling tasks. To demonstrate the capabilities of KOSMOS-1, a large set of experiments are performed across NLP (e.g., language generation, OCR-free text classification), cross-modal transfer (e.g., common-sense reasoning ), non-verbal reasoning (Raven's Progressive Matirces-based IQ tast~\cite{carpenter1990one, raven2003raven}), vision-language (e.g, captioning), and vision (e.g, zero-shot classification). These experimental results show the general-purpose nature of LLMs.

\methodname{KOSMOS-2~\cite{peng2023kosmos2}:} \citet{peng2023kosmos2} extended KOSMOS-1~\cite{huang2023language} for grounding capabilities and named it \textbf{KOSMOS-2}. To this end, they kept the KOSMOS-1's architecture and training objective and proposed a pipeline to extract text spans (i.e. noun phrases and referring expressions) and link them to corresponding regions in the images. This pipeline consists of two steps. First, non-chunks are extracted from the text and linked with regions in the image based on a pre-trained detector. Second, noun chunks are expanded to referring expressions by traversing nouns' dependency tree. Based on this pipeline, they curated GRIT (GRounded Image-Text pairs) from COYO-700M~\cite{kakaobrain2022coyo-700m} and LIAON-2B~\cite{schuhmann2022laion} consisting of 91M images, 115 text spans, and 137M bounding boxes. The input text is represented as a hyperlink similar to markdown where bounding box coordinates are converted into discrete location tokens and added in the appropriate segments. The model is trained on a combination of multi-modal corps from KOSMOS-1~\cite{huang2023language} and GRIT for the next token prediction. After this training, instruction tuning is carried out on language-only and grounded-instruction data. KOSMOS-2 achieves excellent performance on language and vision tasks, grounding tasks, and referring tasks which expands it to a more diverse set of downstream tasks.

\noindent\textbullet \textbf{Training with a General Generative Objective: }
LLMs demonstrate excellent capabilities although they are trained on simple language modeling tasks. Inspired by this success, many works aimed to transfer this to vision-language modeling. Here, we describe methods that propose or train models on simple modeling tasks for pre-training of vision-language models. 
\methodname{SimVLM~\cite{wang2021simvlm}: } \citet{wang2021simvlm} proposed a minimalist pre-training framework for vision-language models. The proposed Simple Vision Language Modeling (\textbf{SimVLM}) framework trains encoder-decoder style models on Prefix Language Modeling (PrefixLM) objective. The prefixLM considers the image to be a prefix for textual description, and thereby, forces the model to complete a description ($x_{\geq T_p}$) given an image and its partial description ($x_{< T_p}$) of randomly selected length ($T_p$). A simple transformer-based encoder-decoder architecture is utilized, and textual and visual embeddings (extracted by the first three blocks of a ResNet) are fed to the encoder, and the decoder outputs a text string. This model is trained on prefixLM on noisy image-text pairs dataset~\cite{jia2021scaling}. SimVLM does not require task-specific architecture or training and beats previous pre-training and state-of-the-art methods on several vision-language tasks. 

\methodname{MaskVLM~\cite{kwon2022masked}: } Motivated by the fact that both text and image can represent the same reality in different formats, \citet{kwon2022masked} proposed joint masked reconstruction language modeling where masked input of one is reconstructed conditioned on other unmasked input. Their model called \textbf{MaskVLM}, consists of an image and language encoder to encode corresponding modalities and a cross-modal decoder with cross-attention to align both modalities. Image and text are masked randomly following \citet{devlin2018bert} and \citet{he2022masked} and the model is trained on joint conditional reconstruction task as well as Image-text Contrastive (ITC) ~\cite{radford2021learning,jia2021scaling} and Image Text Matching (TIM)~\cite{chen2020uniter} task. This results in an efficient model that outperforms similar models for vision-language tasks in a low data regime.

\methodname{mPLUG-OWL~\cite{ye2023mplugowl}: }\citet{ye2023mplugowl} proposed \textbf{mPLUG-OWL}, a modular vision-languages model trained on language modeling objective. The model consists of an image encoder, an image abstractor, and a frozen LLM. This model is trained in two phases. In the first phase, the image encoder and visual abstractor are trained on image-text pairs with language modeling tasks. In the second phase, a visual abstractor along with a Low-Rank Adaptation module is fine-tuned with language-only and multi-modal datasets. mPLUG-OWL performs well on multi-turn dialogue as well as on instruction understanding, visual understanding, and knowledge transfer. 

\begin{figure*}
    \centering
    \includegraphics[width=\textwidth]{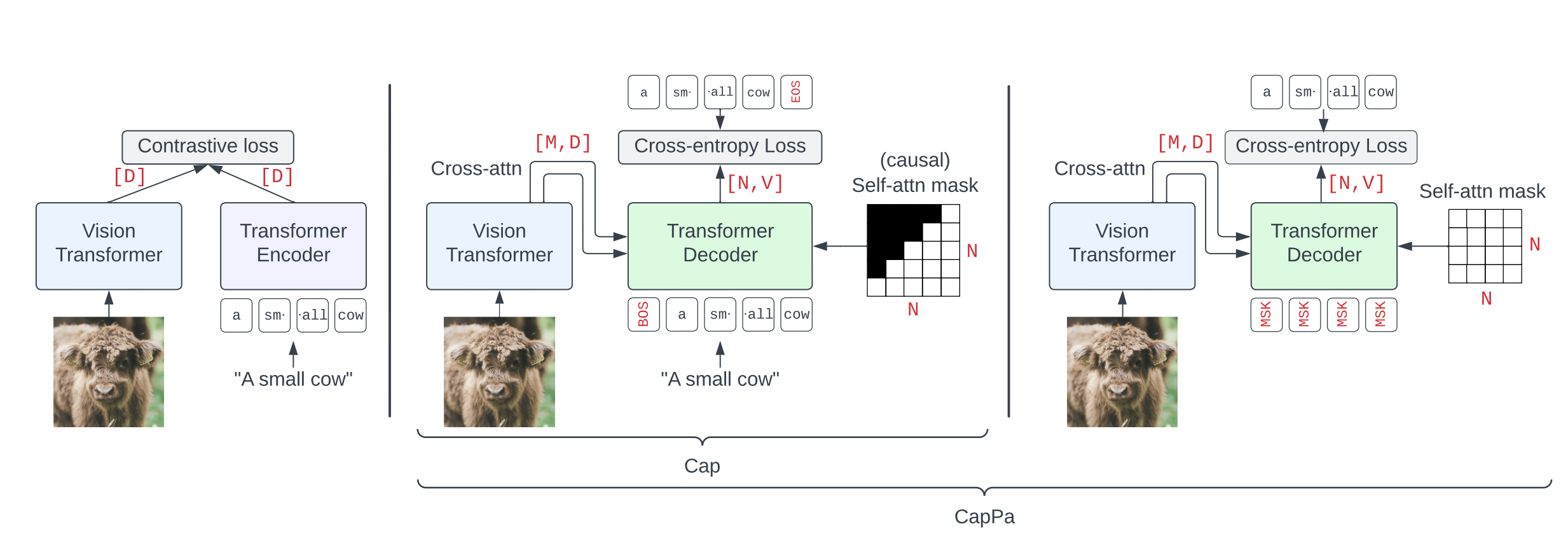}\vspace{-2em}
    \caption{A comparison of contrastive, captioning and CapPa-based~\cite{tschannen2023image} models. In contrastive models (shown on the left), visual and textual encoders are individually optimized. In captioning-based models (shown in the middle), a vision encoder's output combined with input text is fed to a decoder. CapPa (shown on the right) architecture also follows captioning models but with modifications in the training recipe. The figure is taken from \citet{tschannen2023image}. }
    \label{fig:CapPa}
\end{figure*}

\methodname{CapPa~\cite{tschannen2023image}: }Since the CLIP's~\cite{radford2021learning} demonstration of remarkable scaling properties of contrastive learning, contrastive methods have become a norm in vision-language pre-training. \citet{tschannen2023image} revisited the effectiveness of captioning for vision-language pre-training on webscale image-text pair datasets and systematically compared them with contrastive approaches. First, they compared the performance of Captioning-based models (Cap) with CLIP-style models on similar scales and compute budgets. They trained a simple encoder-decoder architecture (ViT as a vision encoder and transformer as a decoder) on a standard next-word prediction task. Their experimental results show that captioning models: a) generally lags behind CLIP-style models in zero-shot classification but the gap reduces with scale, b) matches or outperforms them in few-shot classification, c) have competitive performance for classification task when fine-tuned with large labeled data, and d) with ViT backbone outperforms CLIP-style models for multi-modal tasks. Second, they proposed a new generative pre-training method called \textbf{CapPa}, which, as shown in Fig.~\ref{fig:CapPa}, alternates training between standard auto-regressive prediction (CaP) and parallel prediction (Pa) where the entire caption is predicted in a single pass. The CapPa pre-training improves the performance of ViT. Third, they revealed the scaling properties of captioners by studying various architectures and training procedures and showed performance improvements when training data and architecture are scaled. 

\subsection{Hybrid Contrastive and Generative Learning}
\label{sec:hybrid}

\subsubsection{Foundational Models for Generic Vision-Language Learning}

\noindent\textbullet \textbf{Unification of tasks: }

\methodname{UNITER~\cite{chen2020uniter}: } Inspired by the generalizability of BERT~\cite{lu2019vilbert} in Natural Langauge Processing (NLP) tasks~\cite{devlin2018bert}, \citet{chen2020uniter} proposed Universal Image-TExt Representation (\textbf{UNITER}), a method to leverage conventional image-text datasets (COCO~\cite{lin2014microsoft}, Visual Genome~\cite{krishna2017visual}, Conceptual Captions~\cite{sharma2018conceptual}, SBU Captions~\cite{ordonez2011im2text}) to train foundational models that can be used for heterogeneous vision-language tasks. The authors designed four pre-training tasks spanning generative (i.e., Masked Language Modeling (MLM), Masked Region Modeling (MRM)) and contrastive (Image-Text Matching (ITM), and Word Region Alignment (WRA)) objectives. The UNITER architecture consists of an image and text embedder and a cross-modality contextualized embedding transformer. The text and images of these datasets are fed to respective embedders to extract embeddings. Individual embeddings are fed to the cross-modality transformer for a cross-modal representation. This model is trained on a combination of four different vision-language datasets to optimize the pre-training tasks mentioned earlier. UNITER exhibits excellent generalizability across nine different vision-language tasks, setting state-of-the-art for most tasks.

\methodname{Pixel2Seqv2~\cite{chen2022unified}: }

\citet{chen2022unified} proposed to reformulate and unify four core vision tasks (object detection, instance segmentation, key points prediction, and captioning) into a single pixel-to-sequence interface where both task description and outputs are converted into tokens. Their proposed method, called \textbf{Pixel2Seqv2}, utilized an encoder-decoder architecture with a vision encoder that encodes image inputs and a sequence decoder that generates a single token conditioned on previous tokens, and encoded image. This model is trained on a simple language modeling task conditioned on previous tokens and encoded images. At inference time, output tokens are sampled given a task prompt and input image, and task-specific de-tokenization is performed. This Pixel2Seq can solve four vision tasks efficiently performs without requiring any specialized architecture or losses.

\methodname{VL-x~\cite{cho2021unifying}: }\citet{cho2021unifying} proposed a unified framework to learn different computer vision tasks in a single architecture. The unification is achieved by reformulating these tasks into multi-modal conditional text generation tasks. The proposed Vision-Language (\textbf{VL-x}) employs a pre-trained encoder-decoder language model, such as BART or T5~\cite{lewis2019bart, raffel2020exploring}. Text and visual embeddings are fed to the encoder of this language model. Visual embeddings are extracted from a pre-trained object detector model and consist of Region of Interest (RoI) object features, RoI bounding box coordinates, and image and region IDs. The output of the visual task is converted into a sequence of characters and a task-specific prefix is added (e.g., classification: bird). This augmented text is encoded as learned embeddings and fed to the encoder of the language model along with the visual embeddings. This model is then trained with the task of multi-modal language modeling as well as related vision-language pre-training tasks, such as visual question-answering, image-text matching, visual grounding, and grounded captioning. This framework results in a multi-tasking model that can handle a diverse set of outputs. The models achieve comparable performance to specialized models.

\begin{table*}[t]
    \centering
    \resizebox{1\textwidth}{!}{%
    \begin{tabular}{l|cccccc|cc|cc|cc|cc|cccc|}
    \toprule
        & \multicolumn{2}{c}{\underline{ImageNet}} & \multicolumn{4}{c}{\underline{Zero-shot ImageNet}} & \multicolumn{2}{|c}{\underline{TR}} & \multicolumn{2}{|c}{\underline{IR}} &  \multicolumn{2}{|c}{\underline{VQA}} &  \multicolumn{2}{|c}{\underline{Detection}} &  \multicolumn{2}{|c}{\underline{Segmentation}}\\
        &LP &FT&Std&R&A&V2&Flicker&COCO &Flicker&COCO~&VQA&VQAv2&COCO &RefCOCOg&Pascal&ADE20K&\\
        & \cite{fei2009imagenet} & \cite{fei2009imagenet} & \cite{fei2009imagenet}& \cite{hendrycks2021many} & ~\cite{djolonga2021robustness} & ~\cite{recht2019imagenet} & ~\cite{young2014image} & ~\cite{chen2015microsoft} & \cite{young2014image} & \cite{chen2015microsoft} & \cite{antol2015vqa} & ~\cite{goyal2016making} & ~\cite{lin2014microsoft} &  ~\cite{mao2016generation} & ~\cite{mottaghi2014role} & ~\cite{zhou2017scene}  \\
        \toprule
         CLIP~\cite{radford2021learning}&85.4&-&76.2&88.9&77.1&70.1&88&58.4&76.7&37.8&76.7&-&&-&-&-&\\
        ALIGN~\cite{jia2021scaling}&85.5&88.64&76.4&92.2&75.8&70.1&88.6&58.6&75.7&45.6&-&-&-&-&-&-&\\
        WenLan~\cite{huo2021wenlan}&-&-&-&-&-&-&-&-&-&-&-&-&-&-&-&-&\\
        Florence~\cite{yuan2021florence}&-&90.05&83.74&-&-&-&90.9&64.7&76.7&47.2&80.36&&62.4&-&-&-&\\
        FILIP~\cite{yao2021filip}&-&-&77.1&&&&95.4&61.3&84.7&45.9&-&-&-&-&-&-&\\
        SLIP~\cite{mu2022slip}&75.1&-&47.9&-&-&-&-&-&-&-&-&&&-&-&-&\\
        FLIP~\cite{li2023scaling}&83.6&-&74.3&86.5&51.2&66.8&89.8&61.3&75&45.9&&77.3&-&-&-&-&\\
        MaskCLIP~\cite{dong2023maskclip}&73.7&-&44.5&-&-&-&-&-&70.1&-&-&-&45.4&-&-&50.5&\\
        CLIPA~\cite{li2023inverse}&-&-&69.3&84&43.6&61.7&81.9&56.8&67.5&39.8&-&-&-&-&-&-&\\
        CLIPAv2~\cite{li2023clipav2}&-&-&81.8&94.4&82.7&75.6&-&-&-&-&-&-&-&-&-&-&\\
        EVA~\cite{fang2023eva}&86.5&-&89.6&88.3&86.2&81.6&-&-&-&-&-&-&58.9&-&-&62.3&\\
        EVA-CLIP~\cite{sun2023eva}&-&-&82&94.5&82.1&75.7&-&-&-&-&-&-&-&-&-&-&\\
        EVA-02~\cite{fang2023eva2}&&90&80.4&93.2&82.9&73.8&89.2&64.1&77.9&47.9&-&-&64.1&-&-&62&\\
        OpenCLIP~\cite{cherti2023reproducible}&-&-&77.97&89.32&59.32&70.82&-&-&-&-&-&-&-&&&&\\
        CRIS~\cite{wang2022cris}&-&-&-&-&-&-&-&-&-&-&-&-&-&60.36&&&\\
        MaskCLIP~\cite{zhou2022maskclip}&73.7&83.6&44.5&&&&70.1&&45.6&-&-&-&-&-&31.1&18&\\
        GLIP~\cite{li2022grounded}&-&-&-&-&-&-&-&-&-&-&-&-&61.5&-&17.2&-&\\
        Ground.DINO~\cite{liu2023grounding}&-&-&-&-&-&-&-&-&-&-&-&-&63&87.02&-&-&\\
        OWL-ViT~\cite{minderer2022simple}&-&-&-&-&-&-&-&-&-&-&-&-&-&-&-&-&\\
        ViLD~\cite{gu2021open}&-&-&-&-&-&-&-&-&-&-&-&-&59.5&-&-&-&\\
        GroupViT~\cite{xu2022groupvit}&-&-&-&-&-&-&-&-&-&-&-&&-&-&22.4&-&\\
        OpenSeg~\cite{ghiasi2021scaling}&-&-&-&-&-&-&-&-&-&-&-&&-&-&29.1&-&\\
        \midrule
        Frozen~\cite{tsimpoukelli2021multimodal}&-&-&-&-&-&-&-&-&-&-&-&48.4*&-&-&-&-&\\
        Flamingo~\cite{alayrac2022flamingo}&-&-&-&-&-&-&89.3&65.9&79.5&48&-&82.1&-&-&-&-&\\
        OpenFlamingo~\cite{anas_awadalla_2023_7733589}&&&&&&&&&&&&&&&&&\\
        {MetaLM}~\cite{hao2022language}&-&-&-&-&-&-&-&-&-&-&-&74.5&-&-&-&-&\\
        {KOSMOS-1}~\cite{huang2023language}&-&-&-&-&-&-&-&-&-&-&-&46.7*&-&-&-&-&\\
        {KOSMOS-v2}\cite{peng2023kosmos2}&&-&-&-&-&-&-&-&-&-&-&45.6*&&61.65*&&&\\
        SimVLM~\cite{wang2021simvlm}&83.6&-&-&-&-&-&-&-&-&-&80.3&&-&-&-&-&\\
        CapPa~\cite{tschannen2023image}&-&-&77.5&-&-&-&&62.6&&45.4&&68.6&-&-&-&-&\\
        \midrule
        UNITER~\cite{chen2020uniter}&-&-&-&-&-&-&93.03&65.68&84.34&52.93&74&-&-&87.73&-&-&\\
        Pixel2Seqv2~\cite{chen2022unified}&-&-&-&-&-&-&-&-&-&-&-&-&46.5&-&-&-&\\
        VL-x ~\cite{cho2021unifying}&-&-&-&-&-&-&-&-&-&-&&&-&71.3&-&-&\\
        CoCa~\cite{yu2022coca}&90.6&91&86.3&96.5&90.2&80.7&92.5&66.3&80.4&51.2&82.3&-&-&-&-&-&\\
        FLAVA~\cite{singh2022flava}&75.54&-&-&-&-&-&69.3&43.48&65.22&38.46&-&73.75&-&-&-&-&\\
        PaLI~\cite{chen2022pali}&-&-&72.11&81.97&44.7&64.46&-&-&-&-&-&83.4&-&-&-&-&\\
        BLIP~\cite{li2022blip}&-&-&-&-&-&-&97.4&82.4&97.7&65.1&78.3&-&-&-&-&-&\\
        BridgeTower~\cite{xu2022bridge}&-&-&-&-&-&-&94.7&75&85.8&62.4&-&74.52&-&-&-&-&\\
        X-FM~\cite{zhang2023toward}&81&86.3&-&-&-&-&97.9&81.8&89.4&64.7&79.1&79.5&-&-&-&-&\\
        BLIP-2~\cite{li2023blip}&-&-&-&-&-&-&97.6&85.4&89.7&68.3&&84.03&-&-&-&-&\\
        FIBER~\cite{dou2022coarse}&&&&&&&96&80.06&91.08&69.63&-&78.46&58.4&87.32&-&52&\\
        UniDetector~\cite{wang2023detecting}&-&-&-&-&-&-&-&-&-&-&-&-&49.3&-&-&-&\\
        X-Decoder~\cite{zou2022xdecoder}&-&-&-&-&-&-&94.4&76.1&84.4&58.6&-&77&46.7&64.6&-&58.1&\\
        MaskVLM~\cite{kuwon2022masked}&-&-&-&-&-&-&95.6&76.3&84.5&60.1&75.4&&-&-&-&-&\\
        GLIPv2~\cite{zhang2022glipv2}&-&-&-&-&-&-&-&-&-&-&-&-&62.4&-&-&-&\\
        
    \bottomrule
    \end{tabular} 
    }
\vspace{0.5em}
\caption{Comparison of textually prompted models for a representative set of tasks. For ImageNet, we curate results with Linear Probe (LP), Fine Tuning (FT), Standard (Std) Zero-shot, Renditions, Adversarial, and V2. Furthermore, we present @1 results for text Retrieval (RT) and Image Retrieval (IR) for Flicker and COCO datasets. We also show results for visual question answering, detection, and segmentation tasks for two datasets each. Results with * show different evaluation mechanisms.} 
\label{tab:results_textually_prompted} 
\end{table*}

\noindent\textbullet \textbf{Universal Architectures: }
Purely contrastive vision-language models with separate encoders (e.g., CLIP, ALIG) have shown impressive performance but are not well suited for multi-modal problems that require dealing with both modalities at the same time. On the other hand, multi-modal models, that have fusion and shared attention across modality encoders, are not suitable for uni-modal vision or language tasks. Here, we describe methods that aim to perform well across uni, cross, and multi-modal tasks by proposing new novel architectures and training them on multiple objectives including contrastive, generative, and task-specific losses. 

\methodname{CoCa~\cite{yu2022coca}: }\citet{yu2022coca} proposed a single unified encoder-decoder-based model called Contrastive Captioner (\textbf{CoCa}) that has the capabilities of the single encoder, dual encoder, and encoder-decoder models. The CoCa model consists of a unimodal image and text encoder and a decoupled multi-modal decoder with a cross-attention layer. The unimodal encoders are trained on a contrastive loss like CLIP. This helps the model learn robust and aligned global representations. The decoupled decoder is trained with a generative approach to captioning loss, which helps it learn detailed granularity and region-level information. A combination of these two approaches endows the model with both contrastive and generative capabilities. This strategy results in a foundational model that performs well across a set of diverse vision datasets. A single-trained CoCa model outperforms many specialized models under zero-shot and few-shot and light fine-tuning settings. For instance, it achieves 86.3\%, 88.0\%, and 91.0\% accuracy on ImageNet under zero and few-shot settings and light fine-tuning, respectively. 

\begin{figure*}[t]
    \centering
    \includegraphics[width=1\textwidth]{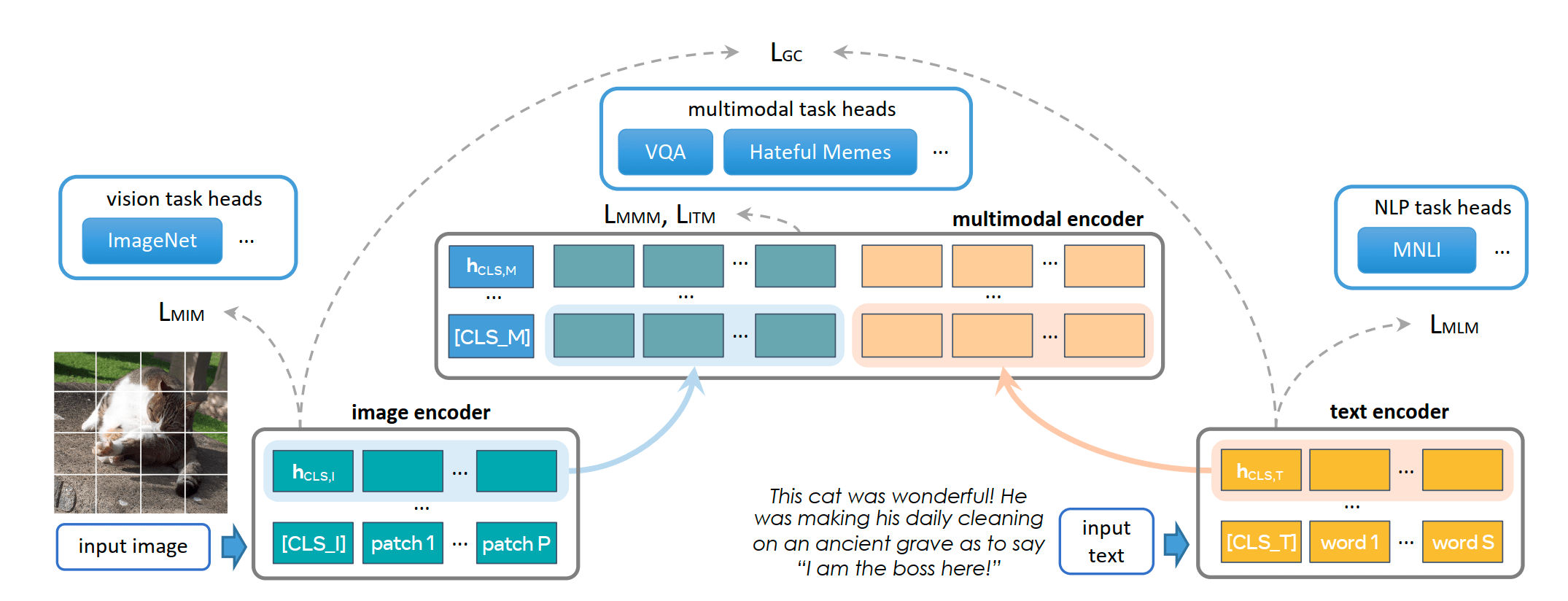}
    \caption{An overview of FLAVA~\cite{singh2022flava} architecture. The model consists of an image, text, and multimodal encoder as well as vision, NLP, and multimodal heads. Figure from \citet{singh2022flava}. }
    \label{fig:falava}
\end{figure*}

\methodname{FLAVA~\cite{singh2022flava}: }\citet{singh2022flava} argued that a truly foundational model must perform well across vision, language, and vision-language tasks. To this end, they proposed an architecture named \textbf{FLAVA} that consists of image and text encoders as well as multi-modal encoders, vision-task heads, language task heads, and multi-modal task heads. This makes FLAVA suitable for both uni-modal and multi-modal tasks. An overview of the proposed architecture is shown in Fig.~\ref{fig:falava}. The image and text encoders convert the input to a representation that is fed to a multi-modal encoder. The multi-modal encoder transformers apply cross-attention and fuse both modalities. This multi-modal representation is fed to modality-specific heads (vision, language, and vision language). To get strong generalization capabilities, this model is trained on multiple uni and multi-modal losses, including a global contrastive loss similar to CLIP~\cite{radford2021learning} for cross-modal alignment, masked multi-modal masking and image-text matching, masked-image modeling, masked-language modeling, etc. The training consists of a uni-modal pre-training of image and text encoders on supervised datasets followed by joint uni-modal and multi-modal training on image-text datasets. To demonstrate the generalizability of FLAVA, it is evaluated on 35 tasks across vision, language, and vision-language tasks and shows impressive performance. 

\methodname{BridgeTower~\cite{xu2022bridge}: }\citet{xu2022bridge} explored how to combine information from different layers of uni-modal encoders. They proposed \textbf{BridgeTower}, an architecture to combine information from different layers of uni-modal decoders without affecting their ability to perform uni-modal tasks. Their architecture consists of a standard vision and language encoder and a cross-modal encoder with multiple bridge layers that connects the top layers of both encoders by co-attention~\cite{lu2019vilbert}. These multi-modal bridges enable bottom-up, cross-modal alignment and fusion of different levels of semantic visual and textual features. Their results demonstrate superior performance on downstream VL tasks despite training on a smaller dataset. 

\methodname{PaLI~\cite{chen2022pali}: }\citet{chen2022pali} investigated the effect of scaling for large image-text models by proposing a new jointly scaled architecture and a new large multilingual image-text dataset. First, they proposed \textbf{PaLI}, a jointly scaled, multilingual, modularized language-vision model which can perform unimodal (language, vision) and multimodal tasks. PaLI architecture consists of a text encoder-decoder transformer (mT5)~\cite{xue2020mt5} and a ViT~\cite{zhai2022scaling} for visual tokens. Both components are pre-trained and only the language component is updated and trained on a myriad of vision and language tasks. The language model is also trained on pure language understanding tasks to avoid catastrophic forgetting.

Second, they introduced a WebLI, 10 billion images, and 12 billion alt-text (textual alternative for images~\footnote{\url{https://webaim.org/techniques/alttext/}}) datasets. For their training, authors used a 1 billion clean subset of this dataset. They also used a combination of vision and language task-specific datasets, such as span corruption and object detection. For task-specific vision dataset training (such as object detection), the output is reformulated with the help of a template-based prompt. Third, they studied scaling laws in vision-language models and showed the importance of scaling vision components and the benefits of a mixture of language models. The PaLI model is pre-trained on over 100 languages and achieves SOTA results on various vision, language, and vision-language tasks. 

\methodname{X-FM~\cite{zhang2023toward}: }Existing models can work across modalities but their performance is not comparable to individual-type foundational models. To solve this problem, ~\citet{zhang2023toward} proposed a new foundational model called \textbf{X-FM} and a new training mechanism. The X-FM architecture consists of three modular encoders, including language, vision, and fusion encoders. The language and vision encoders are a stack of BERT~\cite{devlin2018bert} and ViT~\cite{dosovitskiy2020image} like transformer layers along with post-layer and pre-norm, respectively. In the fusion encoder's self-attention sub-layers, queries are from language, and keys and values are from vision. 

The proposed learning method for X-FM trains encoder with a combination of uni and multi-modal objectives and two new techniques. The language is trained with an encoder with masked language modeling (MLM) and image-text contrastive learning (ITC), a vision encoder with masked image modeling (MIM) and ITC, and a fusion encoder is trained with ITM and image-conditioned masked language modeling (IMLM), and bounding box prediction (BBP). The first new training technique is stopping gradient from vision-language when learning the language encoder. This way, the language encoder is isolated from fusion and trained on MLM and ITC for language modeling and language-vision alignment. The second new technique is the training vision encoder with masked images where the difference between its masked and unmasked outputs is minimized by using MSE loss. This way, the vision encoder is trained on cross-modal and unimodal objectives. This MIM training is resource-efficient, and convenient, enhancing vision and fusion encoder mutually. The X-FM outperforms other general foundational models on twenty-two tasks across language, vision, and language-vision tasks. 

\methodname{BLIP~\cite{li2022blip}: }Previous works rely on the large-scale nature of noisy image-text datasets which, \citet{li2022blip} argues, is sub-optimal. They introduced the \textbf{BLIP} framework which has both understanding and generation capabilities as it effectively utilizes image-text datasets and uses a new architecture. First, they proposed a captioner that produces synthetic captions and a filter that filters noisy captions. This makes it possible to synthesize and filter noisy captions cost-effectively. Second, BLIP has a Multimodal mixture of Encoder-Decocer (MED) architecture which consists of unimodal encoders for image and text and image-grounded text encoder and image-grounded text decoder. This model is trained on two understanding-based (i.e. image-text contrastive, image-text matching) objectives and one generative objective (i.e. language modeling). This framework achieves significant improvements and state-of-the-art performance across a wide variety of tasks. 

\noindent\textbullet \textbf{Efficient Utilization of Pre-Trained Models: }
BLIP and other similar models are prohibitively expensive to train as they require large-scale, end-to-end image-text training, often from scratch. Here, we explain methods that aim to leverage pre-trained vision and language models efficiently for vision-language modeling. 

\methodname{BLILP-2~\cite{li2023blip}: } \citet{li2023blip} proposed \textbf{BLIP-2}, a method to align pre-trained and frozen uni-modal text and image encoders on an image-caption dataset in a computationally efficient way. BLIP-2 bridges the modality gap of frozen uni-modal encoders by using a querying transformer. The parameters of the Q-former are trained to align two modalities by using image-text contrastive learning, image-grounded text generation, and image-text matching losses. This framework is computationally efficient and can leverage large pre-trained uni-modal models. 
\methodname{InstructBLIP\cite{dai2023instructblip}: } \citet{dai2023instructblip} argued that aligning pre-training models on just image-caption can not allow broader generalization. They proposed \textbf{InstructBLIP}, a vision-language instruction tuning framework that enables general foundational models to solve multi-modal tasks through a unified language interface. Similar to BLIP-2~\cite{li2023blip}, their proposed architecture consists of a visual encoder, a Q-Former, and an LLM. Different from BLIP-2, they propose instruction-aware visual feature extraction. Specifically, the Q-former takes encoded images as well as instruction embeddings. This enables Q-Former to extract instruction-related visual features. Like BLIP-2, their model is trained in two phases: first Q-former is trained on image-text pairs, and then instruction-tuning is performed where both LLM and visual encoder are kept frozen. To train this model to perform multi-modal tasks, authors converted a suite of 26 datasets into instruction-tuning format following set templates. InstructBLIP achieves state-of-the-art zero-shot performance across a wide range of vision-language tasks.

\methodname{VPGTrans~\cite{zhang2023transfer}: } Most multi-modal models add a visual encoder (Visual Prompt Generator or VPG) and projection layer to the input of LLMs to enable perception. However, training these visual components is computationally expensive. \citet{zhang2023transfer} proposed \textbf{VPGTrans}, an efficient method to transfer visual encoders across LLMs. To this end, they performed an extensive experimental study to understand how to transfer VPGs across different sizes and types of LLMs. Based on their exploratory analysis, they proposed a two-stage strategy for VPG transfer. In the first stage, trained VPG from source LLM is kept frozen, and the projection module is fine-tuned on target LLM. In the second phase, both VPG and projection layer are trained with target LLM. Their empirical results across different sizes and types of LLMs show performance transfer with significantly less training data and compute resources. 

\methodname{TaCA~\cite{zhang2023taca}: } \citet{zhang2023taca} proposed an efficient framework to upgrade an old foundational model to a new task. To this end, they proposed an adapter called \textbf{TaCA} (Task Agnostic Compatible Adapter), a small module that aligns representations of old and new encoders with distillation loss between the features of old and new encoders and cross-modal contrastive loss. This results in a framework that makes it possible to upgrade modules of these models without the need to re-train. 

\begin{figure*}
    \centering
    \includegraphics[width=1\textwidth]{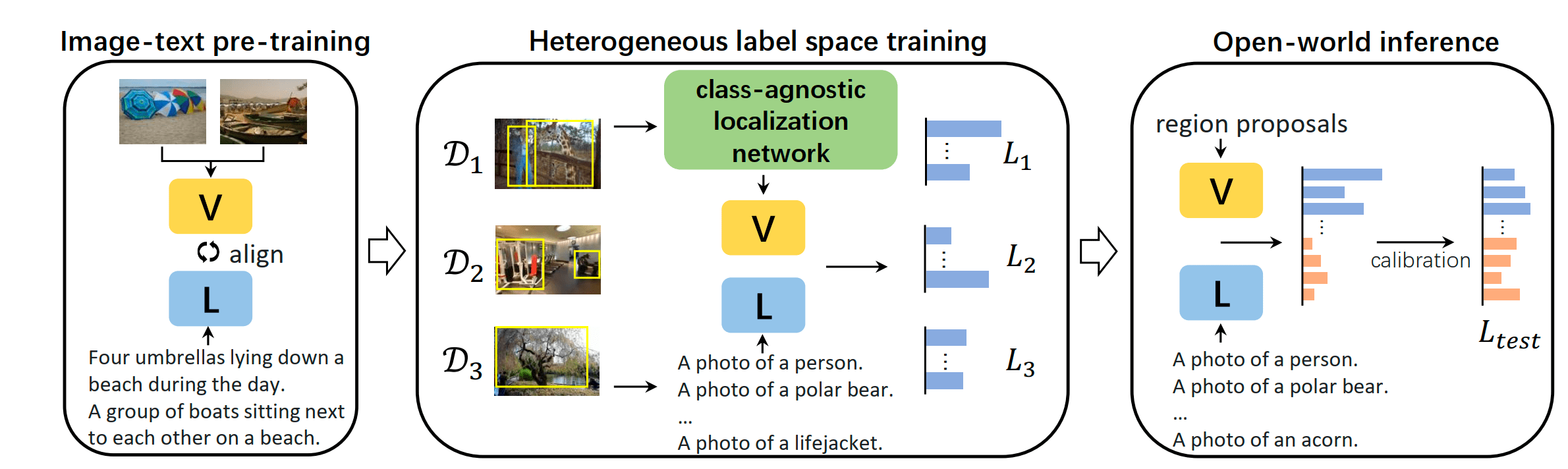}
    \vspace{-1em}
    \caption{An overview of three phases proposed by UniDetector~\cite{wang2023detecting} to prepare a CLIP-like model for open-world object detection: general vision-language training (similar to CLIP~\cite{radford2019language}), object detection-based heterogenous label space training on data combined data from various sources, and inference-time output calibration. Here, V and L refer to vision and language, respectively. The figure is taken from \citet{wang2023detecting}.}
    \label{fig:UniDetector}
\end{figure*}

\subsubsection{Foundational Models for Visual Grounding Tasks}
In this section, we describe methods that aim to solve visual grounding tasks by utilizing contrastive, generative, and other objectives.

\methodname{ViLD~\cite{gu2021open}: } Open-vocabulary object detectors are difficult to train as data requires scales with the number of categories. On the other hand, vision-language models trained on web-scale image-text pairs have shown impressive open-vocabulary performance for classification. \citet{gu2021open} proposed an efficient, two-stage open-vocabulary object detection method that distills knowledge from a pre-trained one-vocabulary classification model. The proposed ViLD (Vision Language Distillation) method consists of a Region Proposal Network (RPN) and a CLIP-like pre-trained vision-language model. First, a Mask-RCNN~\cite{he2017mask} is modified to output class-agnostic object proposals and respective embeddings. Second, the vision head of the pre-trained vision-language model is utilized to extract embeddings, these embeddings are then used to distill knowledge into an object detector. $\ell_1$-loss. Third, the classifier is replaced with a pre-trained text encoder of the vision-language model which generates text embeddings for all categories. Similar to CLIP, cross-entropy loss followed is employed for this purpose. During inference, text embeddings of novel categories are also computed and each region is classified based on the highest similarity with the corresponding text embeddings. ViLD has shown significant improvements over supervised state-of-the-art open-vocabulary methods. 

\methodname{FIBER~\cite{dou2022coarse}: }Most vision-language foundational models either work with image-level understanding tasks (e.g., classification) or region-level understanding tasks(e.g., object detection). \citet{dou2022coarse} proposed two new ideas in FIBER to make vision-language models work for both types of tasks. Specifically, they propose to insert cross-attention layers inside the vision and language encoder for alignment. Moreover, they propose a two-stage, coarse-to-fine training pipeline. In this pipeline, the model is first trained with coarse-grained datasets such as image-text matching, masked language modeling, and image-text contrastive loss. During fine-grained training, the coarse pre-trained model is used at initialization and trained with high-resolution images with bounding box loss, word-region alignment, etc. FIBER can handle a diverse set of image understanding and groundings tasks and consistently improves upon strong baselines for these tasks.

\methodname{UniDetector~\cite{wang2023detecting}: } \citet{wang2023detecting} proposed a method for universal object detection that aims to detect novel categories in the open world. The UniDetector method aims to solve two main problems of universal object detection: training on heterogeneous object detection datasets and novel category discrimination. To this end, they proposed a three-phase training method for universal object detection. First, RegionCLIP-like pre-training methodology is adapted to align vision and text encoders. Second, training on heterogeneous datasets is performed where a Class-agnostic Localization Network (CLN) extracts regions that are fed to the vision encoder. Similarly, template-based category text is fed to the language encoder, and vision-language training is performed. Finally, during inference, a probability calibration is applied to improve novel category detection. An overview of UniDetector is shown in Fig.~\ref{fig:UniDetector}. UniDetector beats supervised state-of-the-art models for large-vocabulary object detection without being trained on training data and set new SOTA for closed-vocabulary object detection.

\methodname{X-Decoder~\cite{zou2022xdecoder}: }Different vision methods operate at different levels of granularity such as image level, object level, and pixel level. \citet{zou2022xdecoder} argued that a method operating at all three levels can exploit the synergy of these tasks and proposed X-Decoder which formulates all levels of a task into a general decoding procedure. X-Decoder is built on top of a vision backbone based on Mask2Former~\cite{cheng2021mask2former}. The decoder takes multi-scale image features extracted by a vision encoder and two sets of queries: textual queries encoded by a text encoder and generic non-semantic queries that aim to decode segmentation masks. The decoder has two different types of outputs: pixel-level masks and token-level semantics. Different combinations of queries and output types can facilitate different tasks. This architecture is trained end-to-end on panoptic segmentation, referring segmentation, and image-text pairs datasets with task-specific semantic losses and mask-based loss. X-decoder exhibits strong zero-shot and task-specific transferability across a wide range of segmentation tasks as well as vision-language tasks.

\methodname{GLIPv2~\cite{zhang2022glipv2}: } \citet{zhang2022glipv2} propose to use vision-language grounding as a meta capability for both localization and understanding tasks. To this end, they proposed a new inter-image region-word contrastive loss that utilizes negative phrases from different examples as potential negatives. A generic model consisting of a vision and a text encoder and a fusion decoder is trained with their proposed loss and grounding loss and masked language modeling loss~\cite{lu2019vilbert} on both image-text and detection datasets. The model can be used in zero-shot settings as well as in fine-tuning settings. Experimental results show both task benefits from each other. 

\section{Conversational Vision-Language Models}
\label{sec:conversationLLMs}

After the impressive performance of Large Language Models (LLMs) for comprehending, reasoning, and holding human-like conversations, there have been several works to incorporate visual modality in them. Conversational VLMs are a subcategory of textually prompted models, however, are equipped to hold human-like conversations based on multi-modal inputs. In this section, we review efforts to create conversational VLMs.

\begin{figure}
    \centering
    \includegraphics[width=0.5\textwidth]{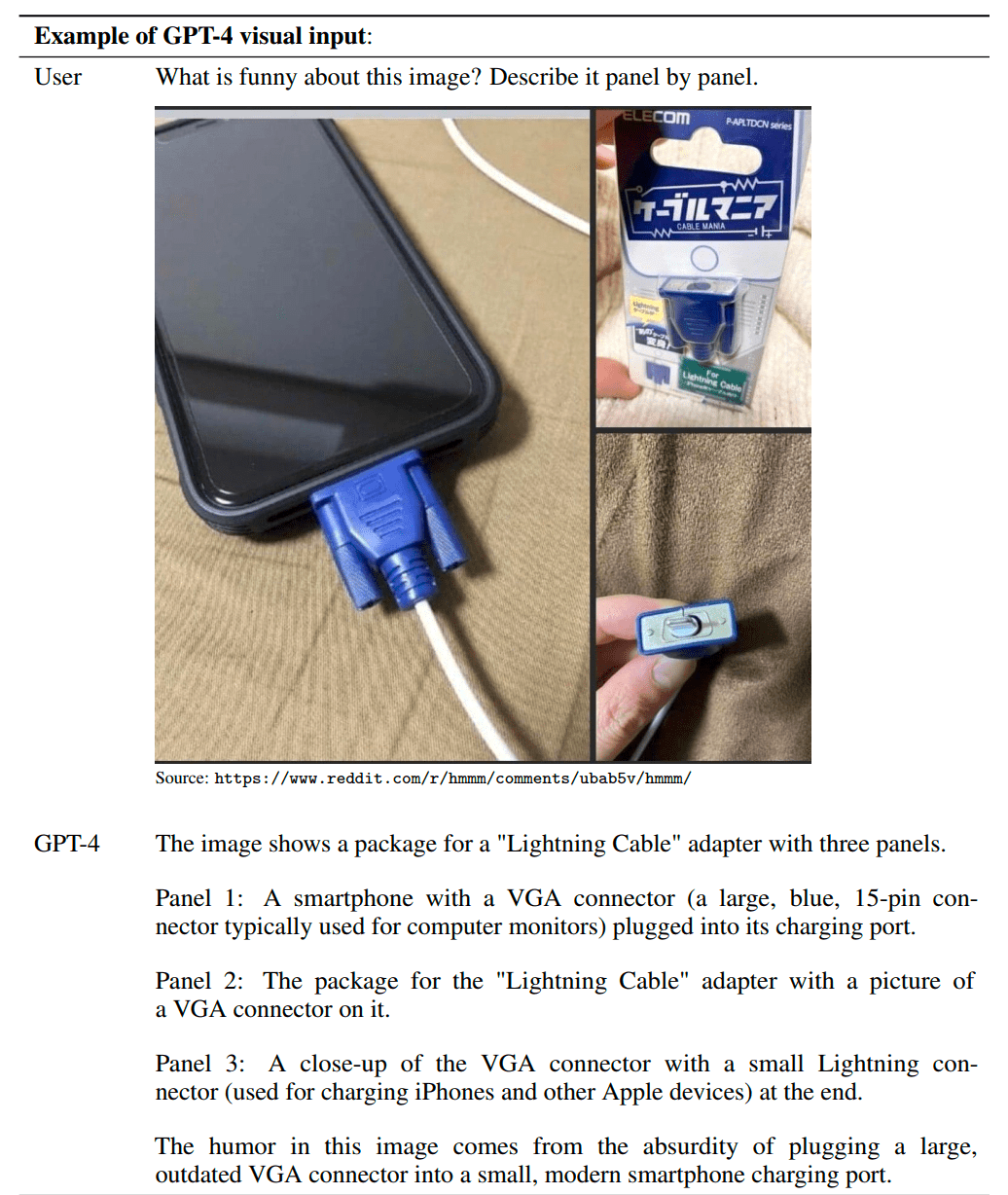}
    \vspace{-1em}
    \caption{An example of GPT4's ability to describe and explain complex situations. Figure from \cite{openai2023gpt4}. }
    \label{fig:gpt4_example}
\end{figure}

\methodname{GPT4~\cite{openai2023gpt4}:} OpenAI developed the first vision-language \textbf{GPT4} model~\cite{openai2023gpt4} which c hold multi-modal conversations and can describe intricate images and solve complex real-world problems. Due to `competitive landscape and ethical considerations', they decided not to open-source this model but provided an API-access through a paywall\footnote{https://help.openai.com/en/articles/7127956-how-much-does-gpt-4-cost}. This model is based on transformer-based architecture~\cite{vaswani2017attention}, pre-trained to predict the next word token using public and private datasets. GPT4 is then fine-tuned with Reinforcement Learning from Human Feedback (RLHF)~\cite{christiano2017deep}. GPT4 shows excellent performance across a span of conventional and real-world NLP, vision, and vision-language tasks. GPT4 performs exceptionally well on HumanEval~\cite{chen2021evaluating}, has a human-level performance on professional and academic exams designed for humans, outperforms previous SOTA language models on conventional NLP tasks, works well on even MMLU~\cite{hendrycks2020measuring} dataset translated across languages, and substantially improves model's ability to follow human intent. GPT4 also shows remarkable performance on vision tasks and describing intricate and complex scenes, of which an example is shown in Fig.~\ref{fig:gpt4_example}. 

\begin{figure}
    \centering
    \includegraphics[width=0.45\textwidth]{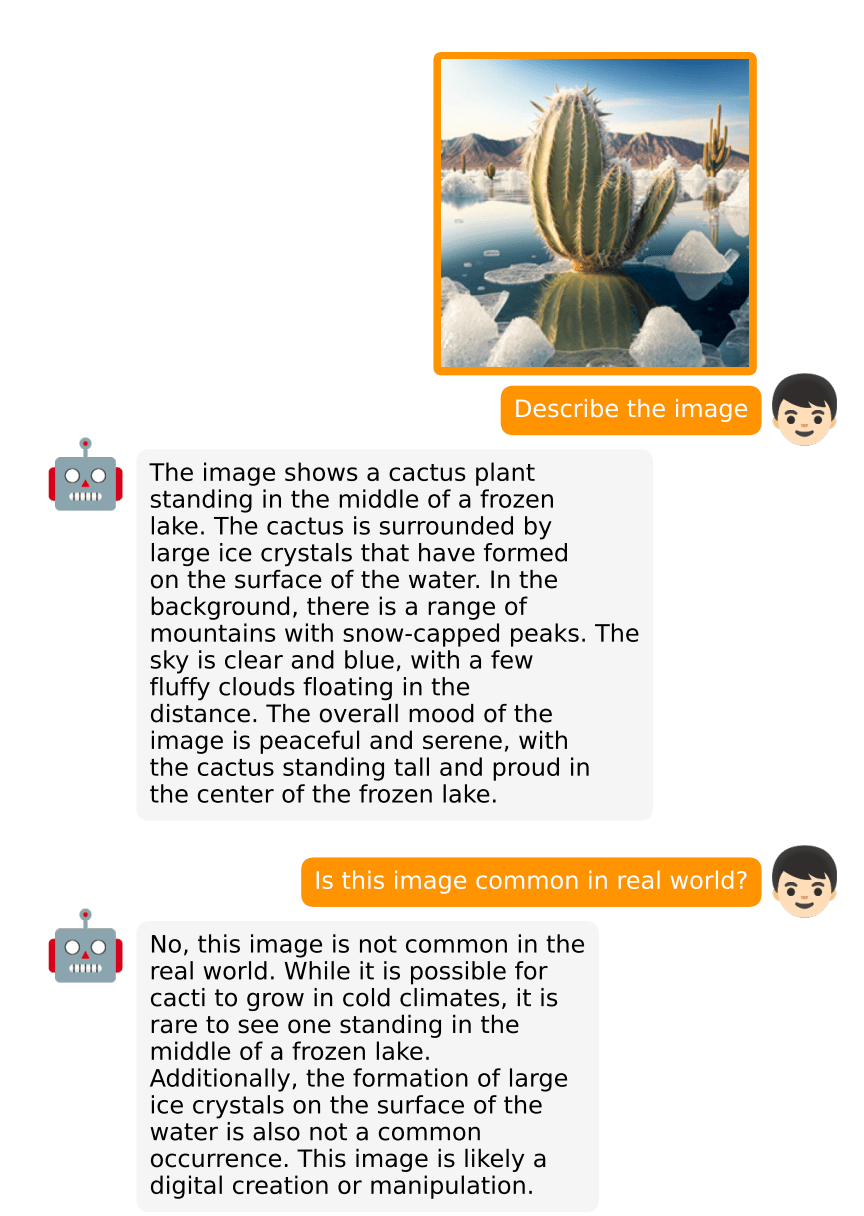}
    \caption{An example demonstrating multi-modal capabilities of MiniGPT4~\cite{zhu2023minigpt}. Figure from \citet{zhu2023minigpt}.}
    \label{fig:minigpt4_example}
\end{figure}

\methodname{MiniGPT-4\cite{zhu2023minigpt}: }GPT4~\cite{openai2023gpt4} has remarkable emerging properties but the model behind it is closed-sourced and even its architectural details are unknown. \citet{zhu2023minigpt} aimed to unravel this and hypothesized that these models utilize large language models. To this end, they presented an open-source version of GPT4 called \textbf{miniGPT4} that consists of a pre-trained large language model (LLM) called Vicuna~\cite{vicuna2023} and a visual component that consists of ViT-G~\cite{sun2023eva} and a Q-Former. MiniGPT-4 adds a single linear projection layer on top of the vision encoder and freezes all other parameters. To align visual features with the LLM, the authors proposed a two-stage training-finetuning scheme. First, MiniGPT-4 is trained on a large set of multi-modal examples consisting of Conceptual Captions~\cite{sharma2018conceptual}, SBU~\cite{ordonez2011im2text}, and LAION~\cite{schuhmann2021laion}. Second, to improve the naturalness and usability, MiniGPT-4 is fine-tuned on a high-quality curated dataset of instructions and respective image and text pairs. MiniGPT-4 exhibits several intriguing properties of GPT4 such as generating intricate image descriptions, creating a website from its sketch, and explaining visual scenarios (see Fig.~\ref{fig:minigpt4_example}).

\methodname{Video-ChatGPT~\cite{maaz2023videochatgpt}: } \citet{maaz2023videochatgpt} proposed \textbf{Video-ChatGPT}, a model that aligns video representation with a Vicuna LLM~\cite{vicuna2023} to enable interaction with videos. Their model consists of a pre-trained Vicuna LLM for language encoding, and a CLIP visual encoder pre-trained with visual instruction following in LLaVA. To make it compatible with videos, frame-level embeddings are averaged pooled along the temporal and spatial dimensions, and concatenated. These concatenated features are fed to a linear layer which transforms them for LLM. This model is instruction-tuned on video-text pairs using auto-regressive training objectives. To enable large-scale instruction fine-tuning, 100,000 video-text data is also developed using human annotations and semi-automatic annotations.

\methodname{XrayGPT~\cite{thawkar2023xraygpt}: } \citet{thawkar2023xraygpt} proposed a model that can analyze and answer open-ended questions about x-ray radiographs. Their proposed model, \textbf{XrayGPT}, consists of a Vicuna LLM as a text encoder and a MedClip for an image encoder. Multi-modal alignment is performed by updating a single linear projection layer on large radiology-specific data curated by authors. The resulting open-source model shows an impressive ability to hold conversations about a radiograph. 

Instruction tuning has played an important role in the alignment of LLMs for following instructions and solving various tasks. Here, we explain methods that aim to extend this behavior to vision-language models. 

\methodname{LLaVA~\cite{liu2023llava}: } Inspired by the amazing instruction-following capabilities of LLMs and closed-sourced vision-language models, \citet{liu2023llava} proposed an open-source visual instruction tuning framework and model dubbed \textbf{LLaVA}. To this end, they have two main contributions. First, they proposed a cost-effective method to curate multi-modal instruction following data. This method leverages ChatGPT~\cite{chatgpt} and GPT4~\cite{openai2023gpt4} for data curation. This data consists of conversations, detailed descriptions of visual inputs, and complex reasoning. Second, they developed a large multi-modal model which utilizes a large pre-trained language model (LLaMA~\cite{touvron2023llama}) and CLIP's vision encoder (ViT). The vision encoder converts input images into features athat are fed to a linear projection layer. This layer converts these features into a space compatible with the LLM. This model is then trained with a two-phase strategy where the model is first trained for vision-language alignment and only the projection layer's parameters are updated. In the second phase, the LLMs and projection layer parameters are fine-tuned end-to-end on the curated dataset. LLaVA is the first visual-instruction following method and demonstrates excellent performance and can even explain complex visual scenes. Similarly, \citet{zhang2023llavar} also presented a visual tuning dataset. 

\methodname{LLaMA-Adapter~\cite{zhang2023llama}: }\citet{zhang2023llama} proposed \textbf{LLaMA-Adapter}, an efficient method that makes LLaMA~\cite{touvron2023llama} model into instruction following method. LLaMA-Adapter is primarily for text-based instruction fine-tuning but it also incorporates visual knowledge. The primary idea behind LLaMA-Adapter is to append a set of learnable adaption prompts as prefixes in the input tokens of the early transformer layers. LLaMA-Adapter can also handle input images by converting them into visual tokens using CLIP~\cite{radford2021learning} based image encoders. These adaptable visual prompts are also incorporated into LLaMA for vision-language tasks. LLaMA is primarily designed for the language but authors also show its superior performance on a large-scale multi-modal dataset called ScienceQA~\cite{lu2022learn}.

\begin{figure}
    \centering
    \includegraphics[width=0.45\textwidth]{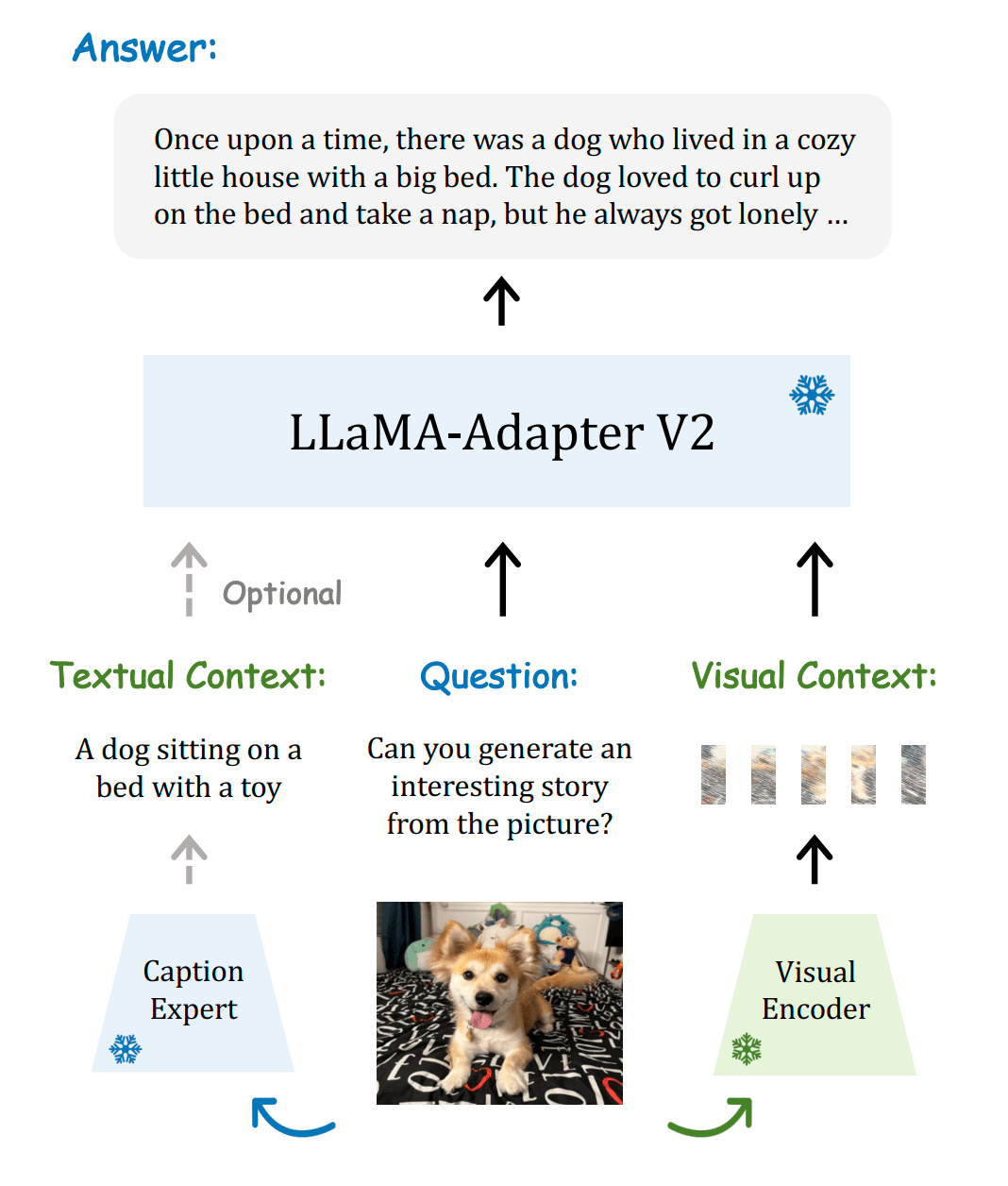}
    \vspace{-1em}
    \caption{Generation pipeline of LLaMA-Adapter v2~\cite{gao2023llamaadapterv2}. Figure from Reference \cite{gao2023llamaadapterv2}.
    }
    \label{fig:llama_adapterv2}
\end{figure}

\methodname{LLaMA-Adapter V2~\cite{gao2023llamaadapterv2}:} Due to lack of instruction following dataset, LLaMA-Adapter~\cite{zhang2023llama} can only work with traditional vision-langauge tasks. \citet{gao2023llamaadapterv2} designed a parameter-efficient visual instruction that shows better performance on language-vision tasks and can conduct multi-run dialogs. LLaMA-Adapter V2 introduces the following improvements for this purpose. First, the authors introduced an early fusion of visual knowledge and added adaptable tokens to different transformer layers which avoids fusion. Second, they introduced disjoint visual-language and language-only training on disjoint parameters with image-captioning and instruction-following data. Third, more learnable parameters are introduced in LLM which includes retraining normalization layers and introduction of a new bias and scale factor for each linear layer of transformer. Finally, to improve image understanding ability, a visual expert is introduced which adds its expert knowledge as a context as shown in Fig.~\ref{fig:llama_adapterv2}. LLaMA-Adapter V2 enhances the instruction following ability of LLaMA, performs better on conventional vision-language tasks, and it is better at visual instruction following compared with V1.

We end our discussion about large visual language models with a brief discussion of their extensions and applications. The above conversational VLMs are generally not adept at visual grounding tasks and can reason about holistic images. Similar to the visual grounding works based on contrastive learning framework, recent efforts try to have visually grounded conversations, e.g., answering questions about a particular object \cite{zhao2023bubogpt,zhang2023gpt4roi,koh2023grounding,berrios2023towards}.
\citet{wu2023visual, zhu2023chatgpt} proposed ways to incorporate visual inputs to ChatGPT~\cite{chatgpt}. Several works also introduced methods to incorporate programming~\cite{surís2023vipergpt, skreta2023errors, mai2023llm, ren2023robots, zhang2023building, wu2023embodied, yoneda2023statler, park2023clara}. Several works have also expanded ChatGPT and other LLMs for diverse applications such as robotics~\cite{stella2023can, yang2023mmreact, jiang2022vima, wu2023tidybot, lin2023match, chakraborty2023re, you2023robot, wanna2023multimodal},  and a multitude of dimensions~\cite{zhao2023chat, ding2023task, chen2023video, ge2023openagi, wake2023chatgpt, hu2023advancing, zhao2023erra, shen2023hugginggpt, guo2023viewrefer, lu2023chameleon, zheng2023gpt4, zhang2023graphtoolformer, skreta2023errors, shukor2023ep, tong2023mass, liu2023softgpt, zhang2023sprint, li2023mimic, doveh2023dense, mu2023embodiedgpt, yuan2023artgpt, li2023evaluating, chen2023video, liu2023visual, zheng2023can, zhang2023graph, zhong2023chatabl, wang2023chatvideo, zhang2023recognize}.

\section{Visually Prompted Models}
\label{sec:visually-prompted}

In this section, we discuss foundational models that can be prompted by non-textual prompts and have been designed for various visual tasks. In Sec. \ref{subsec:Foundational Models for Segmentation}, we discuss foundational models for image segmentation; CLIPSeg \cite{luddecke2022image}, SegGPT \cite{wang2023seggpt}, SAM \cite{kirillov2023segment}, and  SEEM \cite{zou2023segment}. These models can be prompted using diverse prompt types such as text, points, bounding boxes, or even a mask of the desired region to get the target segmentation. A visual foundational model like SAM \cite{kirillov2023segment} is trained on large-scale datasets that contain more than one billion masks and 11 million images. In other domains, such as medical image understanding, large-scale datasets on this scale may not be available. Our discussion then moves on to how SAM can be effectively adapted for other domains, such as medical \cite{ma2023segment, Lei2023medlam, Gong20233DSAMadapterHA, shaharabany2023autosam, gao2023desam, qiu2023learnable, wu2023medical, fedorov20123d}, tracking \cite{yang2023track, sam-pt}, remote sensing \cite{chen2023rsprompter}, and captioning \cite{wang2023caption}. Further, the models like SAM are based on high-complexity Vision Transformer-based architecture \cite{khan2022transformers} and trained on high-resolution inputs, making them less friendly for edge devices. We then discuss how these models can be efficiently adapted \cite{zhang2023faster, zhao2023fast, refsam} to mobile devices. In Sec. \ref{subsec:Generalist Models}, we describe generalist models \cite{wang2023images, wang2023visionllm}  that can perform different tasks simultaneously and can even adapt to the new task given a prompt and very few tasks-specific examples (a.k.a in-context learning). 


\begin{figure*}[!t]
    \centering
    \includegraphics[width=1\textwidth]{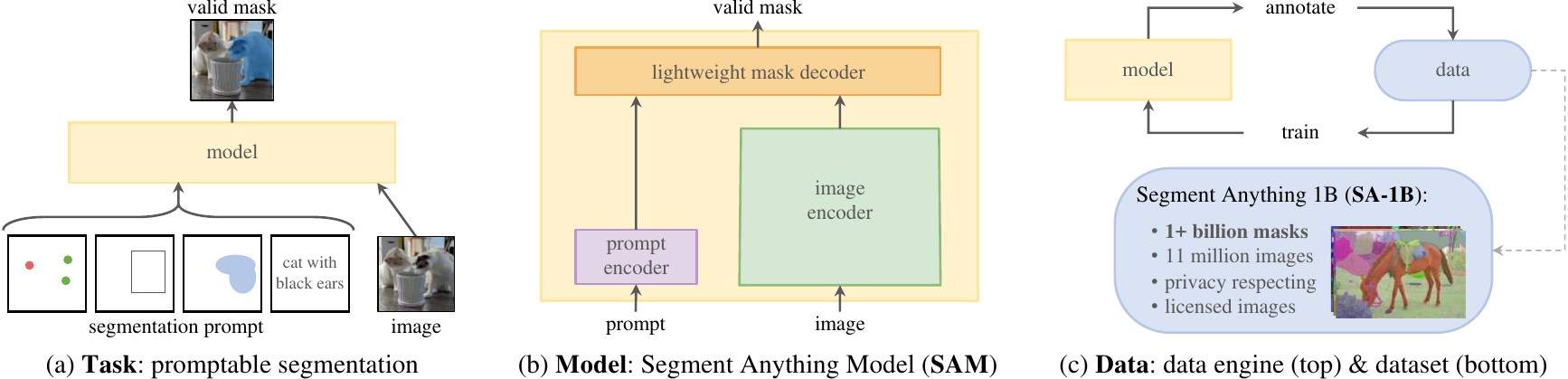}
    \caption{SAM~\cite{kirillov2023segment} consists of a general, prompt-based task, a novel architecture with the capability to take text and mask inputs, and a model-in-the-loop data engine that produced a large-scale segmentation dataset. Figure from Reference \cite{kirillov2023segment}}. 
    \label{fig:sam}
\end{figure*}

\subsection{Foundational Models for Segmentation}
\label{subsec:Foundational Models for Segmentation}
Segmentation involves grouping pixels in meaningful concepts within an image and pixel-wise object identification. 
There are different types of segmentation based on how pixels are grouped, including panoptic, instance, and semantic segmentation. The existing segmentation models are specialized based on the type of segmentation or dataset. The foundational segmentation models aim to develop models that can be generalized universally for segmentation tasks. 

A classic segmentation model cannot generalize to new categories or incorporate new queries without retraining on a relevant dataset.  \textbf{CLIPSeg \cite{luddecke2022image}} exploits CLIP \cite{radford2021learning} generalization capabilities for zero-shot and one-shot segmentation tasks. Their proposed model CLIPSeg achieves this feat by conditioning transformer-based decoder on joint text-visual CLIP embeddings. The CLIPSeg consists of CLIP-based image and text encoders and a transformer-based decoder with U-net \cite{ronneberger2015u} inspired skip connections. The visual and text queries are passed through relevant CLIP encoders to get embeddings which are then fed to the CLIPSeg decoder. In this manner, target segmentation can be prompted by text or an image using CLIP. Therefore, CLIPSeg can generate image segmentations based on arbitrary prompts at test time. %

\textbf{Diversifying Segmentation Tasks:} The segmentation task can be applied to a variety of datasets, including part, semantic, instance, panoptic, person, medical, and aerial images. Here, we cover a few selected methods. \textbf{SegGPT \cite{wang2023seggpt}} provides an in-context learning paradigm and aims to train a single foundational model with a generalizable training scheme for these diverse segmentation tasks. The challenge is to accommodate different segmentation tasks and datasets in a single training framework. SegGPT accomplishes this by mapping different kinds of segmentation data into the same format of images  (random color mapping for each data sample) using an in-context learning framework \cite{wang2023images}. The goal is to color the appropriate areas, such as classes, object instances, parts, etc., in accordance with the context.  After training, the SegGPT model can perform few-shot semantic segmentation, video object segmentation, semantic segmentation, and panoptic segmentation without fine-tuning for the downstream tasks. 

\textbf{SAM \cite{kirillov2023segment}} is a zero-shot segmentation model that does not depend on CLIP and is instead trained from scratch on 1.1 billion masks and only 11 million images. Given an image and visual prompt (box, points, text, or mask) that specifies what to segment in an image, SAM encodes image and prompt embeddings using an image and prompt encoder, respectively which are then combined in a lightweight mask decoder that predicts segmentation masks (Fig. \ref{fig:sam}). SAM is trained to output a valid segmentation mask even when the given prompt is ambiguous e.g., given a point prompt on a person wearing a shirt, the model has to segment the shirt or the person wearing it. SAM is trained on over 1 billion masks with privacy-respecting images and model-in-the-loop dataset annotation settings. The data annotation has three stages: assisted-manual, semi-automatic, and fully automated. In the first stage, SAM assists annotators in annotating masks. By prompting SAM with likely object locations, SAM can generate masks for a subset of objects, while annotators focus on annotating the remaining objects. The final step involves prompting SAM with a regular grid of foreground points, yielding on average 100 high-quality masks per image.

\textbf{Diversifying SAM's Prompting Mechanism:} Inspired by the success of interactive LLMs like ChatGPT, \citet{zou2023segment} argued that the AI-human interaction is important but not well-explored in vision. SAM~\cite{kirillov2023segment} has a limited number of options for this purpose but still lacks a more comprehensive interactive system based on human conversations and it does not support high-level semantic tasks. Hence, the authors aimed to purpose a `universal interface for segmentation everything, everywhere with multi-modal prompts' a.k.a. SEEM~\cite{zou2023segment}. It can take multiple types of prompts including points, masks, text, boxes, and refereed regions of another image, and thus has strong composability. SEEM consists of a text and image encoder as well as a visual sampler for such prompts. These encoded inputs are projected to a joint image-text representations space, which is then fed to a decoder that outputs classes and mask embeddings. SEEM exhibits strong generalization capabilities and is efficient to run. It is also more interactive as it can take five different types of prompts. 

\begin{figure}[!t]
    \centering
    \includegraphics[width=\linewidth]{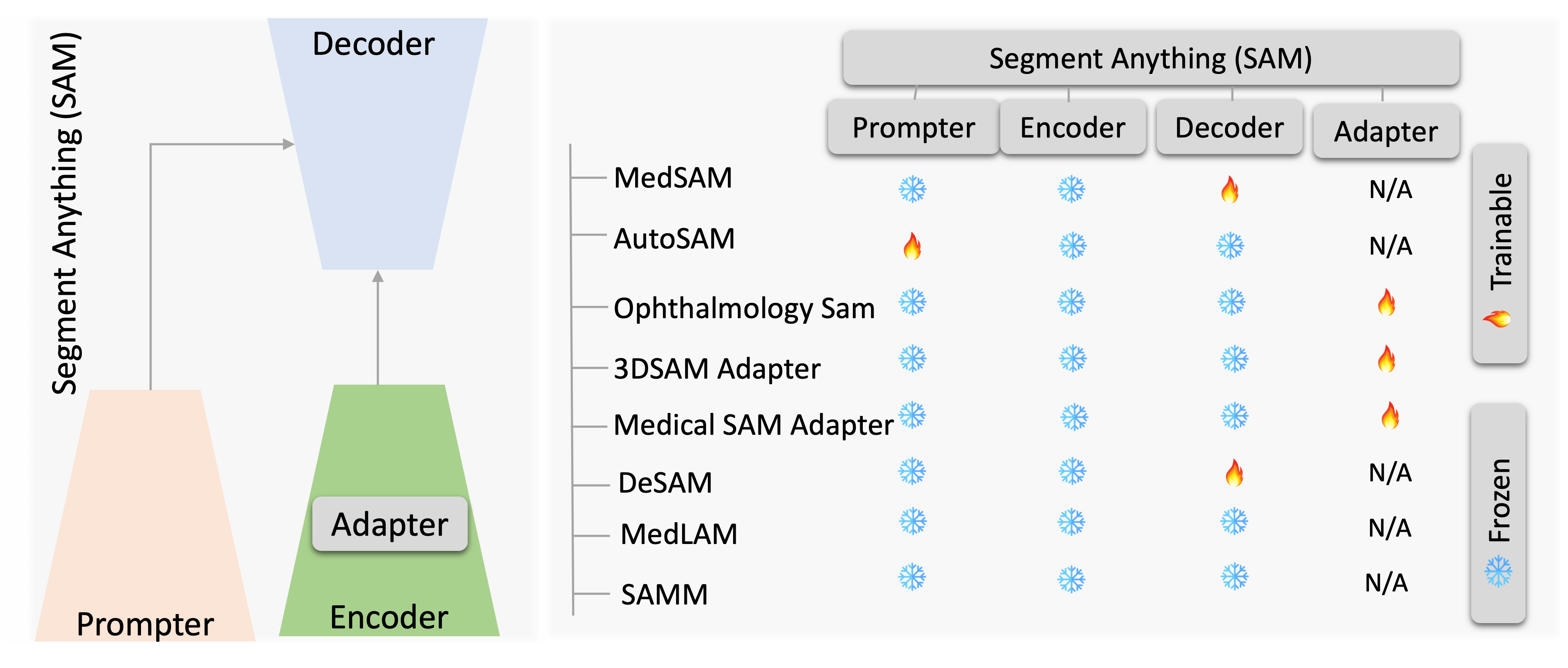}
    \vspace{-2em}
    \caption{An overview of different adaption strategies for SAM for medical datasets. In most methods, adapter layers are introduced within the SAM's encoder to minimize the domain gap from the natural to the medical domain.}
    \label{fig:sam_med}
\end{figure}

Many vision tasks may not have large-scale datasets to train task-specific large-scale models. Next, we will discuss how a foundational model SAM can be adapted for a variety of vision tasks in Sec. \ref{subsubsec:SAM for Captioning} to \ref{subsubsec:SAM for Mobile Applications}. An overview of various adaptations of SAM is exhibited in Fig.~\ref{fig:sam_med}.

\subsubsection{SAM for Medical Segmentation}
\label{subsubsec:SAM for Medical Segmentation}
 Segmenting medical images is fundamental to medical image analysis, which identifies and labels regions of interest (ROI) in different medical images, including organs, lesions, and tissues \cite{moor2023foundation}. SAM \cite{kirillov2023segment}, which is trained on natural images, has difficulty generalizing to medical image segmentation. In this section, we discuss strategies to effectively adapt SAM for medical image datasets.

\noindent \textbf{ Adapting by Fine-Tuning:}  Authors in \cite{ma2023segment} developed \textbf{MedSAM} by extending the SAM approach to medical images. They created a large-scale medical segmentation dataset with 33 segmentation tasks containing over 200,000 masks across 11 different data modalities. Then, they fine-tuned  SAM on the collected dataset for generalized medical image segmentation. In their fine-tuning approach, the image and prompt encoders are frozen, while the SAM decoder is trained on only medical datasets. MedSAM outperformed SAM on 21 3D segmentation tasks and 9 2D segmentation tasks.

\noindent \textbf{Adapting through Auxuliary Prompt Encoder:}  Authors of \cite{shaharabany2023autosam} present a fully automated solution for SAM's prompting for medical datasets, AutoSAM, and propose an auxiliary prompt encoder.
A surrogate prompt for SAM is generated by the \textbf{AutoSAM} auxiliary prompt encoder network based on the input image. In contrast to the prompt encoder provided by SAM, AutoSAM's encoder takes as input the image itself, rather than bounding boxes, points, or masks. A binary cross-entropy loss and a Dice loss are used to propagate gradients from the SAM network to the prompt encoder network during training. In comparison to SAM's decoders, the AutoSAM encoder network uses the Harmonic DenseNet \cite{chao2019hardnet} as its backbone and has fewer trainable parameters. By maintaining the integrity of the main SAM network, AutoSAM can be easily implemented and does not require SAM's fine-tuning to be determined according to an appropriate training schedule.

\noindent \textbf{Adapting Through Adapters:} 
Different multi-modal images bring different segmentation targets, so segmenting multiple targets in ophthalmology can be challenging, such as segmenting blood vessels based on color fundus imagery, or retinal layers based on optical coherence tomography imaging. While SAM can locate several blood vessels from OCTA images, it cannot segment them from color fundus images because blood vessels or lesions may not be distinctive enough to identify \cite{qiu2023learnable}. SAM is extended to segment blood vessels or lesions or retinal layers accurately after one-shot fine-tuning with a new learnable prompt layer \cite{qiu2023learnable}. In addition to learning its target automatically in different modal images, the proposed learnable prompt possesses generalization abilities between datasets. 

Since SAM was originally designed for 2D natural images, it cannot effectively extract 3D spatial information from volumetric medical data. In \textbf{3DSAM-adapter}~\citet{Gong20233DSAMadapterHA}, a modification scheme is devised for the image encoder at the input level to enable the original 2D transformer to accommodate volumetric inputs. This scheme ensures the maximum reusability of pre-trained weights while still allowing them to capture certain 3D spatial patterns through parameter-efficient fine-tuning. Secondly, at the prompt encoder level, an alternative approach is proposed to positional encoding by introducing a visual sampler derived from the image embedding as the representation of the point prompt. Additionally, a set of global queries is employed to filter out noisy prompts. As image token sizes increase with higher dimensions, this strategy mitigates over-smoothing issues caused by the substantial increase in image token size. It also enhances the model's resilience to inaccurate prompts. 
Thus, with minimal adaptations, the transformer, originally trained on natural images, can capture spatial patterns inherent in volumetric medical images despite the domain gap between natural and medical data as well as the disparity in the spatial arrangement between 2D and 3D.

In \textbf{Medical SAM Adapter \cite{wu2023medical}}, a general medical image segmentation adapter is proposed for SAM that is designed with domain-specific knowledge, such as the high dimensionality (3D) of medical data, as well as unique visual prompts, such as clicks and BBoxes. Adapter modules are inserted into the original fundamental model, and then only Adapter parameters are adjusted while the pre-trained parameters are frozen. After training, Medical SAM Adapter (MSA) has demonstrated superior performance on 19 medical image segmentation tasks.

\noindent \textbf{Adapting by Modifying SAM's Decoder:} Through SAM, automatic segmentation can be achieved in two ways. The first option is to use grid points as prompts, and the second is to use boxes that are the same size as the image as prompts. Despite complete fine-tuning, fully automated SAM tends to generate a lot of false positive masks, and its performance falls well short of what clinicians expect. In \textbf{DeSAM \cite{gao2023desam}}, authors argue that in the cross-attention transformer layer of the SAM mask decoder, image embeddings, and prompt tokens interact with each other, resulting in a highly dependent final output mask. Therefore, the model remains sensitive to incorrect prompts, even after fine-tuning. DeSAM separates the mask decoder of SAM into two subtasks: 1) prompt-relevant IoU regression, and 2) prompt-invariant mask learning. The prompt-relevant IoU module predicts IoU scores based on the given prompt and generates mask embeddings. The prompt-invariant mask module (PIMM) generates the mask by combining the embeddings of the image encoder and the mask embeddings from PRIM. The performance degradation of SAM caused by wrong prompts in segment-everything mode is minimized with DeSAM.

\noindent \textbf{SAM as a Medical Annotator:} 
\citet{Lei2023medlam} propose an annotation process for medical datasets using SAM and introduce a few-shot localization framework which is an extension of \cite{lei2021contrastive} and capable of locating any target anatomical part. \textbf{MedLAM} leverages a comprehensive dataset of 14,012 CT scans and incorporates two self-supervised tasks: relative distance regression (RDR) and multi-scale similarity (MSS). MedLAM significantly reduces the annotation burden by requiring annotations of only six extreme points on a few template images. These annotations then enable MedLAM to autonomously identify the target anatomical region across the entire dataset scheduled for annotation. As a result, MedLAM generates a 2D bounding box for each image slice, which is effectively utilized by SAM for subsequent segmentation tasks. Two 3D datasets containing 38 organs were evaluated by MedLAM, demonstrating that MedLSAM is comparable in performance to SAM and its medical adaptations, while minimizing the need to annotate extreme points across the dataset.

Similarly,  Segment Any Medical Model (\textbf{SAMM \cite{liu2023samm}}) is a medical image segmentation tool that combines 3D Slicer and SAM to assist with the development, assessment, and application of SAM. 3D Slicer \cite{fedorov20123d} is an open-source application that is capable of reading and writing a wide variety of file formats, manipulating 2D coordinate systems, and providing consistent user interfaces and tools to facilitate medical image analysis. Segmentation can be automated through prompts that can be applied automatically to subsequent slices. SAM's integration with 3D Slicer enables researchers to segment medical images with a state-of-the-art foundation model.

\subsubsection{SAM for Tracking}
\label{subsubsec:SAM for Tracking}
One of the most important computer vision tasks is to track arbitrary objects in generic scenes and to distinguish regions of interest in videos from the background (also known as video object segmentation or VOS). A bounding box or segmentation mask is typically used to initialize trackers and segmenters trained on large datasets with manually annotated annotations. A specific object mask ground truth is also required to initialize the model under current initialization settings, particularly the semi-supervised VOS. SAM \cite{kirillov2023segment}, a foundational model for segmentation can be used to segment across frames within a video but it leads to poor results due to a lack of temporal consistency. Track Anything (\textbf{TAM}) \cite{yang2023track} proposes to use SAM and the off-the-shelf tracker XMem \cite{cheng2022xmem} to segment and track anything within a video. A user can simply click on an object to initialize SAM and predict the mask. XMem then tracks the object using the initial mask prediction provided by SAM in video based on spatiotemporal correspondences. The user can pause the tracking process and correct any errors immediately. In a similar manner,  SAM-Track \cite{cheng2023segment} utilizes DeAOT \cite{yang2022decoupling} with SAM. 

TAM \cite{yang2023track} and SAM-Track \cite{cheng2023segment} perform well but they do not effectively preserve the original performance of SAM in zero-shot scenarios, which are more challenging. Similarly, semi-supervised methods \cite{cheng2021mask2former, cheng2022xmem} for video object segmentation (VOS) and video instance segmentation (VIS) have performance gaps when they are applied to unseen data, especially when utilizing zero-shot models, which are used to segment object categories outside of the training distribution of video domains where they have not been trained. 

To solve these issues, \textbf{SAM-PT \cite{sam-pt}} proposes to combine SAM sparse point tracking for video segmentation, thus a sparse point annotation is all that is needed for the first frame to denote the target object. The open-world UVO \cite{wang2021unidentified} benchmark shows its strength in generalization to unseen objects. Utilizing state-of-the-art point trackers, such as PIPS \cite{harley2022particle}, SAM-PT provides sparse point trajectory predictions for video segmentation. SAM-PT shows that K-Medoids cluster centers are the most compatible for initiating SAM by using mask labels as initial points to track. Further, to distinguish target objects from their backgrounds, SAM-PT tracks positive and negative points simultaneously. 

\noindent \textbf{SAM-DA \cite{sam_da}} is another approach that uses SAM auto-segment abilities for tracking. Specifically, it determines enormous high-quality target domain training samples automatically from every nighttime image for tracking nighttime UAVs using SAM auto segmentation.

\subsubsection{SAM for Remote Sensing}
\label{subsubsec:SAM for Remote Sensing}
Guided primarily by points, boxes, and coarse-grained masks, SAM \cite{kirillov2023segment} relies heavily on manual guidance due to its interactive nature. Therefore, it makes SAM ineffective when it comes to understanding remote-sensing images in a fully automatic way. The results of SAM are heavily dependent on the type, location, and quantity of prompts used to segment remote-sensing image targets. To achieve desired results, it is usually necessary to refine manual prompts. As a result, SAM presents considerable limitations when used for remote-sensing image segmentation. \textbf{RsPrompter \cite{chen2023rsprompter}} incorporates semantic categorization information with SAM for automated instance segmentation for remote sensing images. RsPrompter proposes learning to generate appropriate prompts for SAM input. It generates prompts containing information about semantic categories by analyzing the intermediate layers of the encoder,  generating prompt embeddings, which can be viewed as point or box embeddings.

\subsubsection{SAM for Captioning}
\label{subsubsec:SAM for Captioning}
An emerging multimodal topic, controlled image captioning uses natural language to explain an image according to human goals, such as examining certain regions of the image or describing it in a particular way. However, the scalability and usability of such interactive image captioning systems are greatly limited by the lack of well-annotated multimodal data.  Caption AnyThing (\textbf{CAT \cite{wang2023caption}}) is a zero-shot image captioning model that leverages pre-trained image captioners \cite{li2022blip, li2023blip, wang2022git} with segment anything (SAM) and large language models, ChatGPT. A user can define visual controls through visual prompts which are then converted into a mask using SAM to select the region of interest. Based on the original image and the mask provided, an image captioner predicts the raw caption. A text refiner (a large language model such as ChatGPT) then modifies user-defined language controls to tailor the language style to the user's preference, thus optimizing the raw descriptions.

\subsubsection{SAM for Mobile Applications}
\label{subsubsec:SAM for Mobile Applications}
A significant amount of attention has been directed toward SAM for its impressive zero-shot transfer performance as well as the ease with which it can be integrated with other models for advanced vision applications such as image editing. It is often necessary to run such use cases on edge devices with limited resources, such as mobile apps. In this section, we discuss efforts to make the SAM model mobile friendly with minimal impact on its generalizability.

One can get MobileSAM by following \cite{kirillov2023segment} pipeline for retraining a new SAM with a smaller image encoder such as replacing the ViT-H with a smaller ViT-L or even ViT-B. However, it can take multiple days and 128 GPUs to train a new SAM using ViT-L or ViT-B as its image encoder. Such resource-intensive retraining can be a non-trivial burden to reproduce or improve their results. A major reason for this optimization difficulty is that the mask decoder and image encoder are coupled together. In \textbf{FasterSAM \cite{zhang2023faster}}, the heavyweight image encoder is replaced by a lightweight one, making SAM more mobile-friendly. The first step is to distill the knowledge from the default image encoder, ViT-H, into a tiny version, ViT. Afterward, FasterSAM aligns the original SAM's mask decoder with the distilled image encoder more closely. MobileSAM, which is more than 60 times smaller than the original SAM, can be trained on a single GPU within less than one day.
In a similar manner, \textbf{Fast Segment Anything \cite{zhao2023fast}} also tries to bring the SAM \cite{kirillov2023segment} like capabilities to the mobile applications. It advocates the use of CNN rather than a vision transformer for image encoding and achieves segmenting anything pipeline into two stages. The first stage consists of implementing a Convolutional Neural Network (CNN)-based detector. Specifically, this method is based on YOLOv8-seg \cite{Jocher2023yolo}, a detector that uses the YOLACT \cite{bolya2019yolact} method for instance segmentation. It then outputs the region of interest that corresponds to the prompt in the second stage. In comparison to SAM, it provides comparable performance with dramatically reduced computational and resource demands, thus enabling real-time applications using only 2\% (1/50) of the SA-1B dataset.

\noindent \textbf{RefSAM \cite{refsam}} is also an efficient and end-to-end  SAM-based framework for the task of referring video object segmentation (RVOS), which performs accurate target object segmentation in videos based on the powerful foundation model SAM. An efficient and lightweight CrossModal MLP that converts the textual features of the referring expression into dense and sparse feature representations is employed. Further, the authors propose to align vision and language features using a parameter-efficient strategy. 

\subsection{Generalist Models}
\label{subsec:Generalist Models}

Using contextual learning, the model can be quickly adapted to a variety of tasks with only a few prompts and examples. The difficulty with in-context learning in computer vision comes from the fact that output representations vary greatly from task to task (requiring different loss functions and architectures.), therefore it's unclear how to define general-purpose task prompts or instructions for a visual model that can reconfigure the model for out-of-domain tasks. 
\textbf{Painter \cite{wang2023images}} is a generalist model that can perform different tasks simultaneously and even adapt to a new task given a prompt and very few task-specific examples. Given an input and output image for a certain task, the pixels of the output image are masked. The objective of the Painter model is then to inpaint the masked output image. This simple training objective allows the unification of several vision tasks (without modifications to model architecture or loss function) including depth estimation, human keypoint detection, semantic segmentation, instance segmentation, image denoising, image deraining, and image enhancement. After training, the Painter can determine which task to perform during inference using the input/output paired images from the same task as the input condition.

\noindent \textbf{VisionLLM \cite{wang2023visionllm}} is another generalist model that aligns vision and language modalities to solve open-ended tasks. Given an image, VisionLLM learns image features using a vision model; these image features along with a language instruction e.g., "describe the image in detail" are passed through a language-guided image tokenizer. The output of the image tokenizer along with language instructions is provided to an open-ended LLM-based task decoder designed to orchestrate various tasks in accordance with language instructions. \textbf{Prismer} \cite{liu2023prismer} is also a vision-language model that leverages diverse pre-trained domain experts in semantic segmentation, object, text, and edge detection, surface normal and depth estimation to perform multiple reasoning tasks such as image captioning and visual question answering.

\section{Heterogeneous Modalities based Models}
\label{sec:Heterogeneous Modalities based Models}
In this section, we discuss foundational models that align multiple paired modalities e.g., image-text, video-audio, or image-depth etc., to learn meaningful representations.

\noindent \textbf{Aligning CLIP with Heterogeneous Modalities:}  \cite{fang2021clip2video} extends the CLIP model for videos. There are two aspects of video-and-language understanding, spatial representation of multimodal image-text training, and temporal relationships of video frames and video language.  \textbf{CLIP2Video} \cite{fang2021clip2video} transfers the spatial semantics of the image-text CLIP model to the video-text retrieval problem by introducing temporal consistency with CLIP using the proposed Temporal Difference Block (TDB) and Temporal Alignment Block (TAB). They are designed to deal with the temporal relationships between video frames and video language. Adding the difference between two frames of an image to the sequence simulates motion change through Temporal Difference Blocks. The Temporal Alignment Block enhances the correlation between video and language by aligning them in the same feature space. 

Similarly, the \textbf{AudioCLIP} \cite{guzhov2021audioclip} model extends the CLIP model to also handle audio. AudioCLIP is therefore a tri-modal hybrid architecture that incorporates the ESResNeXt audio model into the CLIP framework using the AudioSet dataset. New text-to-audio- and image-to-audio loss terms are introduced along with the existing text-to-image-similarity loss term. After training, AudioCLIP is capable of processing all three modalities and any pair of them simultaneously. AudioCLIP outperforms previous approaches in environmental sound classification tasks and extends CLIP zero-shot capabilities to audio modality. Therefore AudioCLIP is capable of cross-modal querying using text, image, and audio or any combination of these modalities.

Both CLIP2Video \cite{fang2021clip2video} and AudioCLIP \cite{guzhov2021audioclip} extend CLIP to include one additional modality, however, there can be multiple types of paired modalities available in practice. \textbf{Image Bind \cite{girdhar2023imagebind}} includes multiple modalities by learning common representations for the paired data modalities e.g., (video, audio) or (image, depth). By aligning visual features with any of the sensory experiences associated with them, images have this 'binding' property that can offer many sources of supervision to learn visual features. For better representation learning, different sensors should be aligned to a single joint embedding space to learn visual features. The problem, however, is that it is not feasible to obtain every type and combination of paired data with the same set of images. The lack of multimodal data across all modalities is one of the major obstacles to learning a joint embedding. Using multiple types of image-paired data, ImageBind learns a single shared representation space, therefore it is independent of the constraint where datasets from all modalities need to co-occur with each other for joint learning. ImageBind combines large-scale paired data (image, text) with other paired data modalities (video, audio) or (image, depth) to develop a joint feature representation, thereby aligning each other modality (audio, depth) with textual embedding. ImageBind expands zero-shot capabilities across four modalities including audio, depth, thermal, and Inertial Measurement Unit (IMU) readings.

\begin{figure*}[!t]
    \centering
    \includegraphics[width=1\textwidth]{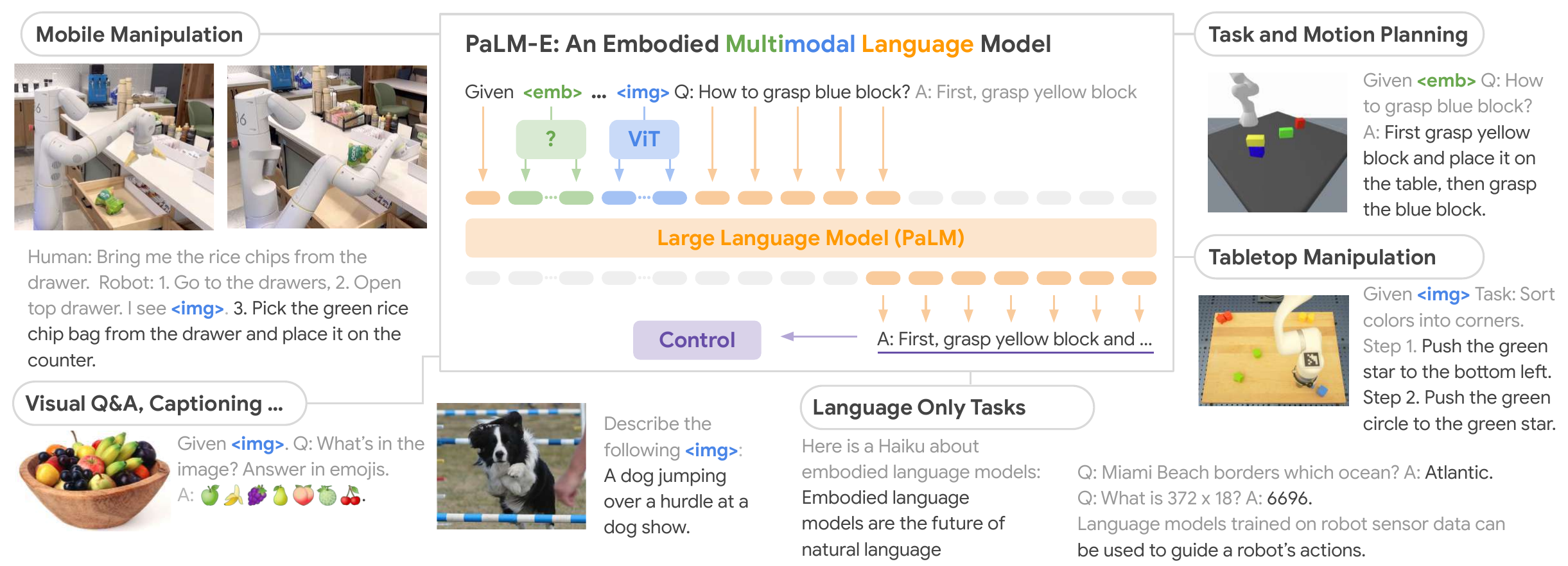}\vspace{-1em}
    \caption{PaLM-E \cite{driess2023palme} applies visual-language knowledge to embodied reasoning, to plan robots within dynamic environments, and to answer questions related to observable reality. PAL-E performs end-to-end training using multimodal sentences, where inputs from a variety of modalities are added to text tokens as input. Figure from Reference \cite{driess2023palme}. 
    }
    \label{fig:palm}
\end{figure*}

\begin{table*}[t]
	\centering\setlength{\tabcolsep}{10pt}
			\resizebox{1\textwidth}{!}{
			\begin{tabular}{clccccx{5cm}x{5cm}}
				\toprule
				\textbf{Type} & \textbf{Foundational Model} &  
                    \textbf{Public} & \textbf{[\url{Link}]}  & \textbf{Prompts} & \textbf{Training} & \textbf{Data Type} & \textbf{Data Size}\\
                \toprule
                \multirow{45}{*}{\rotatebox{90}{\shortstack{Visually Prompted Models}}} & CLIPSeg \cite{luddecke2022image} & \cmark & \href{https://github.com/timojl/clipseg}{Link} & Text, Image & \cmark & PhraseCut \cite{wu2020phrasecut} & 0.34M Images\\
                \cmidrule{2-8}
                &  SegGPT \cite{wang2023seggpt} & \cmark & \href{https://github.com/baaivision/Painter/tree/main/SegGPT}{Link} & Learnable Image Prompt & \cmark & Curation of Twelve Modalities  \cite{wang2023seggpt} & $\approx$ 0.3M Images\\
                \cmidrule{2-8}
                &  SAM \cite{kirillov2023segment} & \cmark & \href{https://github.com/facebookresearch/segment-anything}{Link}& Points, Box, Text & \cmark & SA-1B \cite{kirillov2023segment} & 11M Images, 1.1B Masks\\
                \cmidrule{2-8}
                &  SEEM \cite{zou2023segment} & \cmark & \href{https://github.com/UX-Decoder/Segment-Everything-Everywhere-All-At-Once}{Link} & Text, Points, Scribbles, Boxes, Images & \cmark & COCO \cite{lin2014microsoft} & 118K Images\\
                \cmidrule{2-8}
                &  CAT \cite{wang2023caption} & \cmark & \href{Caption
anything: Interactive image description with diverse mul-
timodal control}{Link} & Text, Points, Boxes, Mask & \xmark & -- & -- \\
                \cmidrule{2-8}
                &  TAM \cite{yang2023track} & \cmark & \href{https://github.com/gaomingqi/Track-Anything}{Link} & Points, Boxes, Clicks & \xmark & -- & -- \\
                \cmidrule{2-8}
                &  SAM-Track \cite{cheng2023segment} & \cmark & \href{https://github.com/z-x-yang/Segment-and-Track-Anything}{Link} & Text & \xmark & --& -- \\
                \cmidrule{2-8}
                &  SAM-PT \cite{sam-pt} & \cmark & \href{https://github.com/SysCV/sam-pt}{Link} & Points & \xmark & --& -- \\
                \cmidrule{2-8}
                &  SAM-DA \cite{sam_da} & \cmark & \href{https://github.com/vision4robotics/SAM-DA}{Link} & Boxes & \cmark & NAT2021 \cite{ye2022unsupervised}  & 0.27M Images \\
                \cmidrule{2-8}
                &  RsPrompter \cite{chen2023rsprompter} & \cmark & \href{https://github.com/KyanChen/RSPrompter}{Link} & Learnable Prompter & \cmark & WHU  \cite{ji2018fully}, NWPU \cite{cheng2014multi}, and SSDD \cite{zhang2021sar} & $\approx$ 6.6K Images\\
                \cmidrule{2-8}
                & MedSAM \cite{ma2023segment} & \cmark & \href{https://github.com/bowang-lab/MedSAM}{Link} & Boxes & \cmark & Curation of Eleven Modalities \cite{ma2023segment} & 200k Masks  \\
                \cmidrule{2-8}
                &  AutoSAM \cite{shaharabany2023autosam} & \xmark & -- & Learnable Prompt &  \cmark & MoNuSeg \cite{kumar2019multi}, GlaS \cite{sirinukunwattana2017gland}, Polyp \cite{jha2020kvasir, bernal2015wm, tajbakhsh2015automated, silva2014toward}, Sun-Seg \cite{ji2022video} & $\approx$ 1.5K Images, 1.1K Videos\\
                \cmidrule{2-8}
                & DSAM-adapter \cite{Gong20233DSAMadapterHA} & \cmark & \href{https://github.com/med-air/3DSAM-adapter}{Link}& Points &  \cmark & KiTS21 \cite{heller2021state}, LiTS17 \cite{bilic2023liver}, Pancreas \cite{antonelli2022medical}, Colon \cite{antonelli2022medical}   & $\approx$ 600 CT scans \\
                \cmidrule{2-8}
                &  DeSAM \cite{gao2023desam} & \cmark & \href{https://github.com/yifangao112/DeSAM}{Link} & Points, Boxes & \cmark & Prostate datasets (NCI-ISBI, I2CVB,  PROMISE12) \cite{gao2023desam} &   --   \\
                \cmidrule{2-8}
                &  MedLAM \cite{Lei2023medlam} & \cmark & \href{https://github.com/openmedlab/MedLSAM}{Link} & Boxes & \cmark & Curation of Sixteen Modalities  \cite{Lei2023medlam}  & $\approx$ 14l CT Sans\\
                \cmidrule{2-8}
                &  FasterSAM \cite{zhang2023faster} & \cmark & \href{https://github.com/ChaoningZhang/MobileSAM}{Link} & Points, Box, Text & \cmark & Subset of SA-1B \cite{kirillov2023segment} & 11K Images   \\
                \cmidrule{2-8}
                &  Fast Segment \cite{zhao2023fast} & \cmark & \href{https://github.com/CASIA-IVA-Lab/FastSAM}{Link} &  Points, Box, Text & \cmark & Subset of SA-1B \cite{kirillov2023segment} & 22K Images  \\
                \cmidrule{2-8}
                &  RefSAM \cite{refsam} & \cmark & \href{https://github.com/LancasterLi/RefSAM}{Link}& Points, Boxes &\cmark & Ref-Youtu-VOS \cite{seo2020urvos}, Ref-DAVIS17 \cite{khoreva2019video} & $\approx$ 4k Videos\\
                \cmidrule{2-8}
                &  Painter \cite{wang2023images} & \cmark & \href{https://github.com/baaivision/Painter/tree/main/Painter}{Link} & Task-specific Prompts & \cmark & NYUv2 \cite{silberman2012indoor}, ADE20K \cite{zhou2019semantic}, COCO \cite{lin2014microsoft}, SIDD \cite{abdelhamed2018high}, LOL \cite{wei2018deep} & $\approx$ 162K Images \\
                \cmidrule{2-8}
                &  VisionLLM \cite{wang2023visionllm} & \cmark & \href{https://github.com/OpenGVLab/VisionLLM}{Link} & Task Descriptions and Categories & \cmark &  COCO \cite{lin2014microsoft}, RefCOCO+ \cite{yu2016modeling}, RefCOCOg \cite{mao2016generation} & 238K Images \\
                \cmidrule{2-8}
                &  Prismer \cite{liu2023prismer} & \cmark & \href{https://github.com/NVlabs/prismer}{Link} & Text & \cmark & COCO \cite{lin2014microsoft}, Visual Genome \cite{krishna2017visual}, Captions \cite{sharma2018conceptual}, SBU-captions \cite{ordonez2011im2text}, Conceptual12M \cite{changpinyo2021conceptual} & 11M Images, 12.7M (Image, Text) \\
                \midrule
                \multirow{6}{*}{\rotatebox{90}{\shortstack{Heterogeneous Modalities \\ based Models}}} &  CLIP2Video \cite{fang2021clip2video} & \cmark & \href{https://github.com/CryhanFang/CLIP2Video}{Link} &  Text & \cmark & MSR-VTT \cite{xu2016msr}, MSVD \cite{chen2011collecting}, VATEX \cite{wang2019vatex} & $\approx$ 33.7K Videos \\
                \cmidrule{2-8}
                &  AudioCLIP \cite{guzhov2021audioclip} & \cmark & \href{https://github.com/AndreyGuzhov/AudioCLIP}{Link} & Text, Audio & \cmark & AudioSet \cite{gemmeke2017audio} & 1.8M (Video, Audio)  \\
                \cmidrule{2-8}
                &  Image Bind \cite{girdhar2023imagebind} & \cmark & \href{https://github.com/facebookresearch/ImageBind}{Link} & Text, Depth, Audio, Thermal & \cmark & Audioset \cite{gemmeke2017audio},  SUN RGB-D \cite{song2015sun}, LLVIP \cite{jia2021llvip},  Ego4D \cite{grauman2022ego4d} &   1.8M (Video, Audio),  10.3K (Image, Depth), 15.4K (Image, Thermal),  3,670 hours of (Video, IMU) \\
                \cmidrule{2-8}
                &  MACAW-LLM  \cite{Macaw-LLM} & \cmark & \href{https://github.com/lyuchenyang/Macaw-LLM}{Link} & Text generated using GPT4  & \cmark & Multi-turn Dialogue & 69K (Image, Text), 50k (Video, Text) \\
                \cmidrule{2-8}
                &  COSA  \cite{chen2023cosa} & \cmark & \href{https://github.com/TXH-mercury/COSA}{Link} & Text & \cmark & CC14M \cite{chen2023cosa}, LAION \cite{schuhmann2021laion}, WebVid \cite{wang2022object} & From 5M to 514M (Image, Text)  \\
                \cmidrule{2-8}
                &  Valley \cite{luo2023valley} & \cmark & \href{https://github.com/RupertLuo/Valley}{Link} & Text generated using ChatGPT & \cmark & Video Conversation and QA \cite{luo2023valley} & 42K Conversations and 5.8K QA about Videos \\
				\midrule
                \multirow{5}{*}{\rotatebox{90}{\shortstack{Embodied Foundational \\ Agents}}} & Palm-E \cite{driess2023palme} & \xmark & -- & Multi-modal (Image, Text) & \cmark & Language-Table \cite{lynch2022interactive}, SayCan \cite{ahn2022can} &  Pushing Dynamics, Tasks in a Kitchen Environment,\\
                \cmidrule{2-8}
                &  ViMA \cite{jiang2022vima} & \cmark & \href{https://github.com/vimalabs/VIMA}{Link} & Multi-modal (Image, Text) & \cmark & VIMA-BENCH \cite{jiang2022vima} & 600K+ expert trajectories for learning\\
                \cmidrule{2-8}
                &  MineDojo \cite{fan2022minedojo} & \cmark & \href{https://github.com/MineDojo/MineDojo}{Link} & Text & \cmark &  Data collected from Minecraft & 730k Videos, 6k Transcripts, 340K Reddit Posts \\
                \cmidrule{2-8}
                &  VOYAGER \cite{wang2023voyager} & \cmark & \href{https://github.com/MineDojo/Voyager}{Link} & Iterative Prompting with Feedback & \cmark & Collected from within Minecraft & Minecraft items, Library storing behavior \\
                \cmidrule{2-8}
                &  LM-Nav \cite{shah2022lmnav} & \cmark & \href{https://github.com/blazejosinski/lm_nav}{Link} & LandMarks, Text & \xmark & -- & -- \\
			\bottomrule
		\end{tabular}}
		\vspace{0.3em}
	\caption{A summary of publicly available information about visually prompted, heterogeneous modality-based models and embodied foundational agents, their prompt design differences, and the nature of their training data type and size.}
    \label{tab:visually_prompted_models_insights}
\end{table*}

\noindent \textbf{Aligning LLM with Heterogeneous Modalities:}  \textbf{MACAW-LLM}  \cite{Macaw-LLM} is an instruction-tuned multi-modal LLM that integrates four different modalities including image, video, audio, and text into a single model. It aligns the feature representation of different data modalities into the embeddings of an LLM thereby bringing those features closer to the textual representation of a large language model. A large-scale multimodal instruction dataset that combines image and video modalities is used to train the MACAW-LLM, which facilitates future work on this type of model learning. MACAW-LLM consists of three modules including Modality Module integrates different modality (e.g., visual and audio data) encoders into  MACAW-LLM, Alignment Module which unifies different modality encoders which are trained independently, and Cognitive Module which is pre-trained LLM.

For video-language reasoning, temporal consistency and context are important. The current image-text foundation models which are trained exclusively on image-text corpora have gained a comprehensive understanding of the semantic correlations between visual concepts and language, however, they lack the temporal contexts required for videos. A solution to this problem is to train on large-scale video-text corpora which is costly to obtain. \textbf{COSA} \cite{chen2023cosa} proposes to generate a video paragraph corpus from the image-text corpus by dynamically transforming it into long-form video paragraph samples. It randomly concatenates a certain number of image-text training samples together from the same batch at each training step. The images and texts are concatenated, ensuring that events and sentences correspond explicitly. The on-the-fly concatenated corpus is superior to short-form video-text corpora \cite{bain2021frozen} due to its richer scene transformations, reduced vision redundancy, and finer-grained captions that describe each frame sequentially.  Concatenated samples are used as input during pretraining by COSA, which has a simple architecture. In addition to visual retrieval, captioning, and answering questions, this model is capable of handling both discriminative and generative tasks.

\textbf{Valley} (Video Assistant with a Large Language model) \cite{luo2023valley} is another multi-modal framework capable of integrating video, image, and language perceptions. A simple projection module is used in the Valley model to bridge video, image, and language modalities, and is further unified with a multi-lingual LLM through the instructions-tuned pipeline. To obtain a unified vision encoding of video and image input, Valley also employs a spatiotemporal pooling strategy. Various video tasks including video question-answering, long description, casual relationship inference, and action recognition are used to collect instructions-following data. This data is then used for instruction fine-tuning to obtain a foundational model for videos.

\section{Embodied Foundational Agents}
\label{sec:Embodied Foundational Agents}
The training of LLMs on massive textual data may result in representations that are relevant to the real world, but to solve a wider range of computer vision and robotics problems grounded in reality, connecting these representations to real-world visual and physical sensor modalities is essential. In this section, we discuss foundational embodied agents for robot manipulation \cite{driess2023palme, jiang2022vima}.

\noindent \textbf{For Robot Manipulation:} By incorporating continuous inputs from sensor modalities of the embodied agent directly into embodied language models, \textbf{Palm-E} \cite{driess2023palme} (see Fig.~\ref{fig:palm})solves this challenging task, enabling the language model itself to make more grounded inferences about sequential decision-making. An LLM based on a Transformer embeds inputs such as images and state estimates into the same latent embedding as language tokens and processes them the same way as texts. The continuous inputs are injected through an encoder into a pre-trained LLM.  These encodings are trained from end to end by enforcing output sequential decisions in terms of natural text that can be understood by the embodied agents.

There are various forms of robotic task specification, including imitating one-shot demonstrations, following linguistic instructions, and reaching visual targets. Different models are often used for these different tasks. \textbf{ViMA} \cite{jiang2022vima} showed that multimodal prompting, interleaving textual and visual tokens, is effective at expressing a wide range of robot manipulation tasks thus learning robot manipulation through multimodal prompts. ViMA follows the transformer-based encoder-decoder network proposed by \cite{raffel2020exploring}. By encoding textual and visual prompt tokens using a pre-trained language model \cite{tsimpoukelli2021multimodal} and decoding robot control actions auto-regressively, VIMA encodes and decodes robot actions for every set of environment interactions. Further, they developed a new simulation benchmark with multimodal prompts that contains more than 600K expert trajectories for imitation learning. They also developed a four-level evaluation protocol for systematic generalization.

\noindent \textbf{For Continual Learners:} For such agents,  \cite{fan2022minedojo} provides a convenient API that makes it easy for users to specify task specifications, change world settings, and observe and act on tasks in Minecraft. There are thousands of open-ended tasks prompted by natural language. As a part of MineDojo, they collected a wealth of data from Minecraft including 30K+ YouTube videos with time-aligned transcripts, 6K+ free-form Wiki pages, and 340K+ Reddit posts with multimedia content. \textbf{MineDojo} \cite{fan2022minedojo} then devised a novel learning algorithm for embodied agents using such data. Their video-text model associates natural language subtitles with time-aligned YouTube videos from MineDojo. Further, they proposed a method for evaluating agents by using such a large pre-trained video-language model that was developed based on Minecraft YouTube videos. This complements human scoring \cite{shah2021minerl}, which is expensive.

Inspired by how humans play Minecraft, \textbf{VOYAGER}  \cite{wang2023voyager} argues that lifelong learning agents should be able to perform similar tasks to humans e.g., it should solve tasks based on the current level of their skill, adapt to environmental feedback to refine skills and actively seek out new tasks and explore the world.  VOYAGER is one of the first embodied lifelong learning agents powered by LLM. It is designed to drive exploration, hone a wide range of skills, and continuously discover new things in Minecraft. Based on the automatic curriculum generated by GPT-4, VOYAGER is designed to solve increasingly difficult tasks so that it discovers as much variety as possible. This is similar to novelty search  \cite{eysenbach2018diversity, conti2018improving}. As VOYAGER stores action programs to complete tasks, a skill library is incrementally built. VOYAGER's capability is compounded over time by composing smaller programs, alleviating the catastrophic forgetting associated with other continual learning methods \cite{parisi2019continual, wang2023comprehensive}.

\noindent \textbf{For Navigation Planning:} An actionable plan can be derived without prior fine-tuning in the target environment using \textbf{LM-Nav} \cite{shah2022lmnav}, which combines pre-trained vision and language models with a goal-conditioned controller. A pre-trained navigation model is combined with two robot-agnostic pre-trained models to achieve this. Using the robot's observations, we construct a topological "mental map" of the environment using a visual navigation model,  ViNG \cite{shah2021ving}. Then to decode free-form textual instructions into textual landmarks, LM-Nav uses a large language model, GPT-3 \cite{brown2020language}. To ground these textual landmarks in the topological map, it employs a vision-language model such as CLIP \cite{radford2021learning}, which infers a joint likelihood over the landmarks and nodes. Finally, to find a plan for the robot, VNM uses a novel search algorithm, which maximizes a probabilistic objective. In a nutshell, LM-Nav provides long-horizon instruction following in complex, real-world environments by combining three large pre-trained models; a self-supervised robotic control model, a vision-language model, and a large language model. 

\section{Open Challenges \& Research Directions}
\label{sec:Open Challenges and Research Directions}
While individual models discussed in this survey may have respective shortcomings and open challenges, this section aims to provide a holistic overview of the shared challenges these approaches (or their subset) face. We also highlight research directions that can help address these challenges.

\noindent \textbf{Multimodal Open-source Models:} 
In NLP tasks, the transition from GPT3 to ChatGPT shows the importance of instruction-following and human-feedback-based reinforcement learning.  For multimodal (text and image) inputs, a similar capability is claimed by GPT4 \cite{openai2023gpt4} to allow reasoning and understanding based on vision-language inputs. However, GPT4 is a closed-source model with restricted access to date and its training details also remain unknown. To bridge this gap, multimodal open-source foundation models such as BLIP2 \cite{li2023blip}, GIT \cite{wang2022git} and Flamingo \cite{alayrac2022flamingo} can be extended with the instruction following and human intent alignment to have ChatGPT-like capabilities in multimodal spaces. To this end, initial efforts have been reported such as Intruct-BLIP, miniGPT4, LLaVA, and Video-ChatGPT.  However, matching the capabilities of GPT4 with open-source public models is still a major challenge for multi-modal foundational models.

\noindent \textbf{Evaluation and Benchmarking:} The open-ended nature of large-scale conversational Vision-Language Models make their comprehensive evaluation challenging. This challenge is shared with the progress in LLM but more severe for visual inputs since the possible tasks and reasoning capabilities become quite diverse for a broad and extensive evaluation. One \emph{quantitative} approach taken is to define a set of instructions covering multiple reasoning aspects and forward the responses from two competing chatbot VLMs to GPT4 to rate them on a scale of 1 to 10. This `LLM-as-a-judge' approach was introduced by Vicuna-Instruction-80 \cite{vicuna2023,zheng2023judging} benchmark for LLM that comprises of 9 instruction categories: \emph{generic, knowledge, math, counterfactual, Fermi, coding, writing, roleplay, common-sense}. This approach has also been extended for VLMs e.g., \cite{maaz2023videochatgpt} uses four criteria (\emph{correctness of information, detail orientation, contextual understanding, temporal understanding, consistency}) scored by GPT4 for benchmarking a VLM tailored for videos. However, the use of an external GPT4 model as a gold standard is still debatable and new efforts in LLM for benchmarking and identifying the corner cases have been reported to address the limitations of existing evaluation measures e.g., \cite{jain2023bring, li2023evaluating, schaeffer2023emergent, bubeck2023sparks, lin2023llm}. Such efforts are likely to be extended to VLMs with even more attention to the peculiar visual aspect of VLMs.

\noindent \textbf{Hallucination:} 
Hallucination refers to the phenomenon where the output generated from a large VLM/LLM  is unreal or nonsensical, often based on a hypothetical scenario.
Foundational language and vision models, specifically those based on Generative Pretrained Models for open-ended conversations \cite{openai2023gpt4,zhu2023minigpt,maaz2023videochatgpt,liu2023llava}, can sometimes fabricate answers, even if they are technically correct in specific contexts. This is because they are trained on massive datasets of text and/or images, that are often noisy, and they may not be able to distinguish between what is real and what is not. Specifically for VLMs where visual data is provided as an input condition, e.g., image-based visual question answering, one form of hallucination ignores the visual input and can only provide an answer based on the text prompt. As an example, an image with a green-colored apple together with a corresponding question \emph{what color is the apple in this image?} may result in an answer \emph{red} due to over-reliance on training data and ignoring the prompt context. One way to control hallucinations is to provide explicit instructions (or so-called \emph{system commands}) to a conversational LLM to provide its answers based on the provided context e.g., asking the chatbot to provide the missing information in patient health records while being strictly grounded in the facts available in the patient data. Other strategies to mitigate hallucination include the chain of thought prompting \cite{lu2023chameleon,zhang2023multimodal}, self-consistency (voting) \cite{mundler2023self,li2023evaluatingb} and use of knowledge bases for retrieval augmented generation \cite{ranjit2023retrieval,liu2023reta,pan2023retrieving}. 

\noindent \textbf{Multimodal Alignment:} Existing VLMs sometimes also suffer from poor alignment between vision-language (or other modalities). For instance, Segment anything \cite{kirillov2023segment} performance with text prompts is weaker as compared to the visual prompts (points, boxes, or masks). For heterogeneous modalities, such an alignment can be even more challenging. Approaches like ImageBind \cite{girdhar2023imagebind}  demonstrate viable ways in which the alignment between several modalities can be achieved, however, there is still much room to demonstrate strong alignment capabilities for even a wider range of related inputs that have the shared semantic space. For example, when a human sees a picture of a food item, it not only recognizes tcomputing item's category but also remembers the taste, the recipe used to cook it, and the crunchy sound each bite generates while eating the food. For a unified representation space that can provide a complete understanding of the world around us, foundational models targeted at learning joint embedding spaces will be crucial for further development.

\noindent \textbf{Large Data and Compute Requirements:}
Training large-scale vision and language models are data and compute-intensive. Labeled data on a large scale can be costly and time-consuming to acquire, especially for specialized visual domains or low-resource languages. Similarly, their inference is also costly due to the many parameters involved. The computational demands of these models limit their accessibility and scalability in many real-world applications. As an example, applications requiring real-time inference capability or ones requiring deployment on the edge and mobile devices with limited on-device compute and restricted battery times. Similarly, visual prompt-based models like Segment Anything \cite{kirillov2023segment} would benefit from a real-time speed to ensure intractability \cite{mobile_sam,zhao2023fast}. However, the current versions with a high-performing image encoder do not offer real-time overall processing. Efforts, such as retentive networks \cite{sun2023retentive}, can be integrated within VLMs for high throughput processing.  

\noindent \textbf{Adaptation of FMs:} Foundational model training often consumes long training times and massive compute resources. Therefore, FMs are adapted to several downstream tasks and applications. The efficient adaptation of FMs without damaging the extensive knowledge learned by the model is an open-research question with many interesting initial efforts reported in recent years. Due to great interest in LLMs and Diffusion models, such Parameter-Efficient Fine-tuning (PEFT) approaches are primarily explored for these two model classes but are directly applicable to adapt other vision foundational models as well. Some representative approaches include Low-rank Adaptation (LoRA) \cite{hu2021lora} and its variants such as QLoRA \cite{dettmers2023qlora} and GLoRA \cite{chavan2023one}, Prefix tuning \cite{li2021prefix,liu2021p}, Adapters \cite{zhang2023llama,gao2023llamaadapterv2}, Prompt tuning \cite{lester2021power,chen2023sam}. Reducing the compute and memory footprint for quick adaptation of textually and visually prompted foundational models is still an open research direction since the existing approaches require careful selection of hyper-parameters (e.g., rank in LoRA or the placement and dimensions of bottleneck adapters) and can result in loss of generalization performance.

\noindent \textbf{Vulnerability to Adversarial Attacks:}
Foundation models, similar to other neural network-based models, can be fooled by adversarial attacks. These attacks involve carefully crafted inputs that can cause the model to generate incorrect or harmful output. However, there are specific ways for adversarial attacks on foundational models that make them susceptible to unwanted behavior. As an example, models based on conversational LLMs have been shown vulnerable to adversarially prompt injection which needs a direct interaction between the adversary and the LLM-based conversational agent \cite{ChatgptJailbreak,perez2022ignore,RedteamingLLMs}. Greshake et al. \cite{greshake2023more} demonstrate that even a direct interaction between the model and adversary is not needed in LLM-integrated applications, and an adversary can remotely poison the information retrieved by the conversational agent via indirect prompt injection. This leads to a spectrum of vulnerabilities for LLMs and VLMs including manipulated content, fraud, malware, intrusion, personal information leakage, and denial of services via language and visual prompt injections. Carlini et al. \cite{carlini2023aligned} recently showed that NLP-based optimization attacks are weak and their failure on foundational models should not be taken as a certification of robustness. They further demonstrate that in conversational VLMs, an attack can be easily launched by adversarially perturbing the inputs to get harmful responses from the model. Maus et al. \cite{maus2023adversarial} show how non-meaningful text can be appended with the prompts to fool generative text and image models. The exemplars (input-output pairs) provided for In-Context Learning of VLMs can also be changed to fool the model \cite{wang2023adversarial}. Visual prompt-based models e.g., SAM have also been attacked by corrupting their inputs and associated prompts \cite{zhang2023attack, huang2023robustness}. Strategies to robustify foundational VLMs against such attacks is an open research question of high significance.

\noindent \textbf{Bias and Fairness:}
Foundational models in vision and language can inherit and amplify biases present in the data used for training them. Biases, stereotypes, and prejudices related to race, underrepresented groups, minority cultures, and gender can make the models output biased predictions or exhibit skewed behavior.  For example, recent work \cite{shtedritski2023does} shows the sensitivity of the CLIP model towards red circles, where simply drawing a red circle around a person's face increases its chances of being misclassified as a murderer, suspect, or a missing person. This behavior potentially emerges from the data that contain examples from news media outlets that typically put a red circle around criminals in their broadcasts. In concurrent efforts, new benchmarks have been developed to assess the capability of existing VLMs towards certain biases \cite{hall2023visogender}.
Addressing biases in foundational AI models is crucial to ensure fairness, inclusivity, and ethical deployment of these systems.

\noindent \textbf{Interpretablity:} 
Foundation models are often difficult to interpret, which can make it difficult to understand how they work and why they generate the output they do. In this direction, existing methods have investigated chain-of-thought reasoning to explain the outputs generated by vision and language models \cite{wei2022chain}. New benchmarks are also developed to evaluate and train explicitly focusing on providing detailed step-wise rationales for model choices e.g., ScienceQA \cite{lu2022learn}. An interesting idea, called Visual Programming, is to use interpretable neuro-symbolic representation to break down a complex task into simpler steps that explain the rationale of a particular output from GPT-3 \cite{gupta2023visual}. While these efforts are promising, they have several failure cases and can be further improved e.g., by allowing user feedback to improve the interpretation generated by the model. 

\noindent \textbf{Limited Contextual Understanding:}
While transformer-based foundational models have shown impressive language and vision understanding capabilities, they can still struggle with certain contextual nuances. Understanding sarcasm, irony, or other forms of figurative image and language inputs (e.g., memes) can be challenging for these models, leading to inaccurate interpretations or responses. While there have been initial efforts on this topic in language-only models, similar efforts for large multi-modal models are much needed and remain an open problem. 

\noindent \textbf{Lack of Real-world Understanding:}
Foundational models for language and vision lack a deep understanding of the world. They can only generate prompt-conditioned text, code, or visual outputs that are consistent with the data they have been trained on. This training is focused on outputting the next sequence element given previous sequence elements or learning multimodal alignment. However, such training is much different from human learning and reasoning which is grounded in physical reality \cite{hu2023neural}. This means that large language and vision models may not be able to understand the meaning of what they are generating or to reason about complex concepts grounded in the real world. Egocentric perception and embodied AI agents based on foundational multimodal models need to develop world models and the alignment of heterogeneous modalities for improved physically grounded reasoning.  In this direction, works on embodied foundational models such as MineDojo \cite{fan2022minedojo}, VPT \cite{baker2022learning} and Voyager \cite{wang2023voyager} use open-ended nature of Minecraft game as a testbed for embodied agents based on GPT models. However, taking these agents to real-world tasks and complex environments is a challenging problem requiring much further work. Google's Palm-E \cite{driess2023palme} is a step in this direction that combines ViTs with Palm \cite{chowdhery2022palm,anil2023palm} LLM model trained with language and sensory inputs to allow embodied agents to understand commands and take informed actions. 

\section{Conclusion}
Models with a foundational understanding of multiple modalities, including natural language and vision, are essential for developing AI systems that can effectively perceive and reason about the real world. This survey reviews vision and language foundational models focusing on their architecture types, training objectives, downstream task adaption, and their prompting designs. We provide systematic categorizations of textually-prompted, visually-prompted, and heterogeneous modality models. We provide a broad coverage of their applications in a variety of visual tasks including zero-shot recognition and localization abilities, visual dialogue about an image or a video, cross-modal, and medical data understanding. We summarize how foundational models in vision can act as generalist models solving multiple tasks simultaneously and their combination with large language models gives rise to foundational embodied agents that can continually learn and navigate in a complex environment. We hope that this effort will spur further research in leveraging the potential of foundational models along with addressing their limitations, e.g. limited contextual understanding, biases, and vulnerability to malicious uses.

{
\small
\bibliographystyle{plainnat}
\bibliography{refs}

\begin{thebibliography}{366}
\providecommand{\natexlab}[1]{#1}
\providecommand{\url}[1]{\texttt{#1}}
\expandafter\ifx\csname urlstyle\endcsname\relax
  \providecommand{\doi}[1]{doi: #1}\else
  \providecommand{\doi}{doi: \begingroup \urlstyle{rm}\Url}\fi

\bibitem[cha()]{chatgpt}
Introducing chatgpt.
\newblock \url{https://openai.com/blog/chatgpt}.
\newblock Accessed: 2023-06-30.

\bibitem[Cha(2023)]{ChatgptJailbreak}
How to jailbreak chatgpt, 2023.
\newblock URL \url{https://watcher.guru/news/how-to-jailbreak-chatgpt}.

\bibitem[Red(2023)]{RedteamingLLMs}
Red-teaming large-language models, 2023.
\newblock URL \url{https://huggingface.co/blog/red-teaming}.

\bibitem[Abdelhamed et~al.(2018)Abdelhamed, Lin, and Brown]{abdelhamed2018high}
Abdelrahman Abdelhamed, Stephen Lin, and Michael~S Brown.
\newblock A high-quality denoising dataset for smartphone cameras.
\newblock In \emph{Proceedings of the IEEE conference on computer vision and
  pattern recognition}, pages 1692--1700, 2018.

\bibitem[Agrawal et~al.(2019)Agrawal, Desai, Wang, Chen, Jain, Johnson, Batra,
  Parikh, Lee, and Anderson]{agrawal2019nocaps}
Harsh Agrawal, Karan Desai, Yufei Wang, Xinlei Chen, Rishabh Jain, Mark
  Johnson, Dhruv Batra, Devi Parikh, Stefan Lee, and Peter Anderson.
\newblock Nocaps: Novel object captioning at scale.
\newblock In \emph{Proceedings of the IEEE/CVF international conference on
  computer vision}, pages 8948--8957, 2019.

\bibitem[Ahn et~al.(2022)Ahn, Brohan, Brown, Chebotar, Cortes, David, Finn, Fu,
  Gopalakrishnan, Hausman, et~al.]{ahn2022can}
Michael Ahn, Anthony Brohan, Noah Brown, Yevgen Chebotar, Omar Cortes, Byron
  David, Chelsea Finn, Chuyuan Fu, Keerthana Gopalakrishnan, Karol Hausman,
  et~al.
\newblock Do as i can, not as i say: Grounding language in robotic affordances.
\newblock \emph{arXiv preprint arXiv:2204.01691}, 2022.

\bibitem[Alayrac et~al.(2022)Alayrac, Donahue, Luc, Miech, Barr, Hasson, Lenc,
  Mensch, Millican, Reynolds, et~al.]{alayrac2022flamingo}
Jean-Baptiste Alayrac, Jeff Donahue, Pauline Luc, Antoine Miech, Iain Barr,
  Yana Hasson, Karel Lenc, Arthur Mensch, Katherine Millican, Malcolm Reynolds,
  et~al.
\newblock Flamingo: a visual language model for few-shot learning.
\newblock \emph{Advances in Neural Information Processing Systems},
  35:\penalty0 23716--23736, 2022.

\bibitem[Anil et~al.(2023)Anil, Dai, Firat, Johnson, Lepikhin, Passos, Shakeri,
  Taropa, Bailey, Chen, et~al.]{anil2023palm}
Rohan Anil, Andrew~M Dai, Orhan Firat, Melvin Johnson, Dmitry Lepikhin,
  Alexandre Passos, Siamak Shakeri, Emanuel Taropa, Paige Bailey, Zhifeng Chen,
  et~al.
\newblock Palm 2 technical report.
\newblock \emph{arXiv preprint arXiv:2305.10403}, 2023.

\bibitem[Antol et~al.(2015)Antol, Agrawal, Lu, Mitchell, Batra, Zitnick, and
  Parikh]{antol2015vqa}
Stanislaw Antol, Aishwarya Agrawal, Jiasen Lu, Margaret Mitchell, Dhruv Batra,
  C~Lawrence Zitnick, and Devi Parikh.
\newblock Vqa: Visual question answering.
\newblock In \emph{Proceedings of the IEEE international conference on computer
  vision}, pages 2425--2433, 2015.

\bibitem[Antonelli et~al.(2022)Antonelli, Reinke, Bakas, Farahani,
  Kopp-Schneider, Landman, Litjens, Menze, Ronneberger, Summers,
  et~al.]{antonelli2022medical}
Michela Antonelli, Annika Reinke, Spyridon Bakas, Keyvan Farahani, Annette
  Kopp-Schneider, Bennett~A Landman, Geert Litjens, Bjoern Menze, Olaf
  Ronneberger, Ronald~M Summers, et~al.
\newblock The medical segmentation decathlon.
\newblock \emph{Nature communications}, 13\penalty0 (1):\penalty0 4128, 2022.

\bibitem[Awadalla et~al.(2023)Awadalla, Gao, Gardner, Hessel, Hanafy, Zhu,
  Marathe, Bitton, Gadre, Jitsev, Kornblith, Koh, Ilharco, Wortsman, and
  Schmidt]{anas_awadalla_2023_7733589}
Anas Awadalla, Irena Gao, Joshua Gardner, Jack Hessel, Yusuf Hanafy, Wanrong
  Zhu, Kalyani Marathe, Yonatan Bitton, Samir Gadre, Jenia Jitsev, Simon
  Kornblith, Pang~Wei Koh, Gabriel Ilharco, Mitchell Wortsman, and Ludwig
  Schmidt.
\newblock Openflamingo, March 2023.
\newblock URL \url{https://doi.org/10.5281/zenodo.7733589}.

\bibitem[Bain et~al.(2021)Bain, Nagrani, Varol, and Zisserman]{bain2021frozen}
Max Bain, Arsha Nagrani, G{\"u}l Varol, and Andrew Zisserman.
\newblock Frozen in time: A joint video and image encoder for end-to-end
  retrieval.
\newblock In \emph{Proceedings of the IEEE/CVF International Conference on
  Computer Vision}, pages 1728--1738, 2021.

\bibitem[Baker et~al.(2022)Baker, Akkaya, Zhokhov, Huizinga, Tang, Ecoffet,
  Houghton, Sampedro, and Clune]{baker2022learning}
Bowen Baker, Ilge Akkaya, Peter Zhokhov, Joost Huizinga, Jie Tang, Adrien
  Ecoffet, Brandon Houghton, Raul Sampedro, and Jeff Clune.
\newblock Learning to play minecraft with video pretraining (vpt).
\newblock \emph{OpenAI Blog, June}, 23, 2022.

\bibitem[Bernal et~al.(2015)Bernal, S{\'a}nchez, Fern{\'a}ndez-Esparrach, Gil,
  Rodr{\'\i}guez, and Vilari{\~n}o]{bernal2015wm}
Jorge Bernal, F~Javier S{\'a}nchez, Gloria Fern{\'a}ndez-Esparrach, Debora Gil,
  Cristina Rodr{\'\i}guez, and Fernando Vilari{\~n}o.
\newblock Wm-dova maps for accurate polyp highlighting in colonoscopy:
  Validation vs. saliency maps from physicians.
\newblock \emph{Computerized medical imaging and graphics}, 43:\penalty0
  99--111, 2015.

\bibitem[Berrios et~al.(2023)Berrios, Mittal, Thrush, Kiela, and
  Singh]{berrios2023towards}
William Berrios, Gautam Mittal, Tristan Thrush, Douwe Kiela, and Amanpreet
  Singh.
\newblock Towards language models that can see: Computer vision through the
  lens of natural language.
\newblock \emph{arXiv preprint arXiv:2306.16410}, 2023.

\bibitem[Bilic et~al.(2023)Bilic, Christ, Li, Vorontsov, Ben-Cohen, Kaissis,
  Szeskin, Jacobs, Mamani, Chartrand, et~al.]{bilic2023liver}
Patrick Bilic, Patrick Christ, Hongwei~Bran Li, Eugene Vorontsov, Avi
  Ben-Cohen, Georgios Kaissis, Adi Szeskin, Colin Jacobs, Gabriel
  Efrain~Humpire Mamani, Gabriel Chartrand, et~al.
\newblock The liver tumor segmentation benchmark (lits).
\newblock \emph{Medical Image Analysis}, 84:\penalty0 102680, 2023.

\bibitem[Bolya et~al.(2019)Bolya, Zhou, Xiao, and Lee]{bolya2019yolact}
Daniel Bolya, Chong Zhou, Fanyi Xiao, and Yong~Jae Lee.
\newblock Yolact: Real-time instance segmentation.
\newblock In \emph{Proceedings of the IEEE/CVF international conference on
  computer vision}, pages 9157--9166, 2019.

\bibitem[Bommasani et~al.(2021)Bommasani, Hudson, Adeli, Altman, Arora, von
  Arx, Bernstein, Bohg, Bosselut, Brunskill,
  et~al.]{bommasani2021opportunities}
Rishi Bommasani, Drew~A Hudson, Ehsan Adeli, Russ Altman, Simran Arora, Sydney
  von Arx, Michael~S Bernstein, Jeannette Bohg, Antoine Bosselut, Emma
  Brunskill, et~al.
\newblock On the opportunities and risks of foundation models.
\newblock \emph{arXiv preprint arXiv:2108.07258}, 2021.

\bibitem[Brock et~al.(2021)Brock, De, Smith, and Simonyan]{brock2021high}
Andy Brock, Soham De, Samuel~L Smith, and Karen Simonyan.
\newblock High-performance large-scale image recognition without normalization.
\newblock In \emph{International Conference on Machine Learning}, pages
  1059--1071. PMLR, 2021.

\bibitem[Brown et~al.(2020)Brown, Mann, Ryder, Subbiah, Kaplan, Dhariwal,
  Neelakantan, Shyam, Sastry, Askell, et~al.]{brown2020language}
Tom Brown, Benjamin Mann, Nick Ryder, Melanie Subbiah, Jared~D Kaplan, Prafulla
  Dhariwal, Arvind Neelakantan, Pranav Shyam, Girish Sastry, Amanda Askell,
  et~al.
\newblock Language models are few-shot learners.
\newblock \emph{Advances in neural information processing systems},
  33:\penalty0 1877--1901, 2020.

\bibitem[Bubeck et~al.(2023)Bubeck, Chandrasekaran, Eldan, Gehrke, Horvitz,
  Kamar, Lee, Lee, Li, Lundberg, et~al.]{bubeck2023sparks}
S{\'e}bastien Bubeck, Varun Chandrasekaran, Ronen Eldan, Johannes Gehrke, Eric
  Horvitz, Ece Kamar, Peter Lee, Yin~Tat Lee, Yuanzhi Li, Scott Lundberg,
  et~al.
\newblock Sparks of artificial general intelligence: Early experiments with
  gpt-4.
\newblock \emph{arXiv preprint arXiv:2303.12712}, 2023.

\bibitem[Byeon et~al.(2022)Byeon, Park, Kim, Lee, Baek, and
  Kim]{kakaobrain2022coyo-700m}
Minwoo Byeon, Beomhee Park, Haecheon Kim, Sungjun Lee, Woonhyuk Baek, and
  Saehoon Kim.
\newblock Coyo-700m: Image-text pair dataset.
\newblock \url{https://github.com/kakaobrain/coyo-dataset}, 2022.

\bibitem[Caesar et~al.(2018)Caesar, Uijlings, and Ferrari]{caesar2018coco}
Holger Caesar, Jasper Uijlings, and Vittorio Ferrari.
\newblock Coco-stuff: Thing and stuff classes in context.
\newblock In \emph{Proceedings of the IEEE conference on computer vision and
  pattern recognition}, pages 1209--1218, 2018.

\bibitem[Cao et~al.(2022)Cao, Tan, Gao, Chen, Heng, and Li]{cao2022survey}
Hanqun Cao, Cheng Tan, Zhangyang Gao, Guangyong Chen, Pheng-Ann Heng, and
  Stan~Z Li.
\newblock A survey on generative diffusion model.
\newblock \emph{arXiv preprint arXiv:2209.02646}, 2022.

\bibitem[Carion et~al.(2020)Carion, Massa, Synnaeve, Usunier, Kirillov, and
  Zagoruyko]{carion2020end}
Nicolas Carion, Francisco Massa, Gabriel Synnaeve, Nicolas Usunier, Alexander
  Kirillov, and Sergey Zagoruyko.
\newblock End-to-end object detection with transformers.
\newblock In \emph{Computer Vision--ECCV 2020: 16th European Conference,
  Glasgow, UK, August 23--28, 2020, Proceedings, Part I 16}, pages 213--229.
  Springer, 2020.

\bibitem[Carlini et~al.(2023)Carlini, Nasr, Choquette-Choo, Jagielski, Gao,
  Awadalla, Koh, Ippolito, Lee, Tramer, et~al.]{carlini2023aligned}
Nicholas Carlini, Milad Nasr, Christopher~A Choquette-Choo, Matthew Jagielski,
  Irena Gao, Anas Awadalla, Pang~Wei Koh, Daphne Ippolito, Katherine Lee,
  Florian Tramer, et~al.
\newblock Are aligned neural networks adversarially aligned?
\newblock \emph{arXiv preprint arXiv:2306.15447}, 2023.

\bibitem[Caron et~al.(2021)Caron, Touvron, Misra, Jégou, Mairal, Bojanowski,
  and Joulin]{caron2021emerging}
Mathilde Caron, Hugo Touvron, Ishan Misra, Hervé Jégou, Julien Mairal, Piotr
  Bojanowski, and Armand Joulin.
\newblock Emerging properties in self-supervised vision transformers.
\newblock \emph{arXiv preprint arXiv: 2104.14294}, 2021.

\bibitem[Carpenter et~al.(1990)Carpenter, Just, and Shell]{carpenter1990one}
Patricia~A Carpenter, Marcel~A Just, and Peter Shell.
\newblock What one intelligence test measures: a theoretical account of the
  processing in the raven progressive matrices test.
\newblock \emph{Psychological review}, 97\penalty0 (3):\penalty0 404, 1990.

\bibitem[Chakraborty et~al.(2023)Chakraborty, Weerakoon, Poddar, Tokekar, Bedi,
  and Manocha]{chakraborty2023re}
Souradip Chakraborty, Kasun Weerakoon, Prithvi Poddar, Pratap Tokekar,
  Amrit~Singh Bedi, and Dinesh Manocha.
\newblock Re-move: An adaptive policy design approach for dynamic environments
  via language-based feedback.
\newblock \emph{arXiv preprint arXiv:2303.07622}, 2023.

\bibitem[Changpinyo et~al.(2021)Changpinyo, Sharma, Ding, and
  Soricut]{changpinyo2021conceptual}
Soravit Changpinyo, Piyush Sharma, Nan Ding, and Radu Soricut.
\newblock Conceptual 12m: Pushing web-scale image-text pre-training to
  recognize long-tail visual concepts.
\newblock In \emph{Proceedings of the IEEE/CVF Conference on Computer Vision
  and Pattern Recognition}, pages 3558--3568, 2021.

\bibitem[Chao et~al.(2019)Chao, Kao, Ruan, Huang, and Lin]{chao2019hardnet}
Ping Chao, Chao-Yang Kao, Yu-Shan Ruan, Chien-Hsiang Huang, and Youn-Long Lin.
\newblock Hardnet: A low memory traffic network.
\newblock In \emph{Proceedings of the IEEE/CVF international conference on
  computer vision}, pages 3552--3561, 2019.

\bibitem[Chavan et~al.(2023)Chavan, Liu, Gupta, Xing, and Shen]{chavan2023one}
Arnav Chavan, Zhuang Liu, Deepak Gupta, Eric Xing, and Zhiqiang Shen.
\newblock One-for-all: Generalized lora for parameter-efficient fine-tuning.
\newblock \emph{arXiv preprint arXiv:2306.07967}, 2023.

\bibitem[Chen and Dolan(2011)]{chen2011collecting}
David Chen and William~B Dolan.
\newblock Collecting highly parallel data for paraphrase evaluation.
\newblock In \emph{Proceedings of the 49th annual meeting of the association
  for computational linguistics: human language technologies}, pages 190--200,
  2011.

\bibitem[Chen et~al.(2023{\natexlab{a}})Chen, Zhu, Haydarov, Li, and
  Elhoseiny]{chen2023video}
Jun Chen, Deyao Zhu, Kilichbek Haydarov, Xiang Li, and Mohamed Elhoseiny.
\newblock Video chatcaptioner: Towards the enriched spatiotemporal
  descriptions.
\newblock \emph{arXiv preprint arXiv:2304.04227}, 2023{\natexlab{a}}.

\bibitem[Chen et~al.(2023{\natexlab{b}})Chen, Liu, Chen, Zhang, Li, Zou, and
  Shi]{chen2023rsprompter}
Keyan Chen, Chenyang Liu, Hao Chen, Haotian Zhang, Wenyuan Li, Zhengxia Zou,
  and Zhenwei Shi.
\newblock Rsprompter: Learning to prompt for remote sensing instance
  segmentation based on visual foundation model, 2023{\natexlab{b}}.

\bibitem[Chen et~al.(2021)Chen, Tworek, Jun, Yuan, Pinto, Kaplan, Edwards,
  Burda, Joseph, Brockman, et~al.]{chen2021evaluating}
Mark Chen, Jerry Tworek, Heewoo Jun, Qiming Yuan, Henrique Ponde de~Oliveira
  Pinto, Jared Kaplan, Harri Edwards, Yuri Burda, Nicholas Joseph, Greg
  Brockman, et~al.
\newblock Evaluating large language models trained on code.
\newblock \emph{arXiv preprint arXiv:2107.03374}, 2021.

\bibitem[Chen et~al.(2023{\natexlab{c}})Chen, He, Li, Jin, Feng, and
  Liu]{chen2023cosa}
Sihan Chen, Xingjian He, Handong Li, Xiaojie Jin, Jiashi Feng, and Jing Liu.
\newblock Cosa: Concatenated sample pretrained vision-language foundation
  model.
\newblock \emph{arXiv preprint arXiv: 2306.09085}, 2023{\natexlab{c}}.

\bibitem[Chen et~al.(2023{\natexlab{d}})Chen, Zhu, Ding, Cao, Zhang, Wang, Li,
  Sun, Mao, and Zang]{chen2023sam}
Tianrun Chen, Lanyun Zhu, Chaotao Ding, Runlong Cao, Shangzhan Zhang, Yan Wang,
  Zejian Li, Lingyun Sun, Papa Mao, and Ying Zang.
\newblock Sam fails to segment anything?--sam-adapter: Adapting sam in
  underperformed scenes: Camouflage, shadow, and more.
\newblock \emph{arXiv preprint arXiv:2304.09148}, 2023{\natexlab{d}}.

\bibitem[Chen et~al.(2020{\natexlab{a}})Chen, Kornblith, Norouzi, and
  Hinton]{chen2020simple}
Ting Chen, Simon Kornblith, Mohammad Norouzi, and Geoffrey Hinton.
\newblock A simple framework for contrastive learning of visual
  representations.
\newblock In \emph{International conference on machine learning}, pages
  1597--1607. PMLR, 2020{\natexlab{a}}.

\bibitem[Chen et~al.(2020{\natexlab{b}})Chen, Kornblith, Swersky, Norouzi, and
  Hinton]{chen2020big}
Ting Chen, Simon Kornblith, Kevin Swersky, Mohammad Norouzi, and Geoffrey~E
  Hinton.
\newblock Big self-supervised models are strong semi-supervised learners.
\newblock \emph{Advances in neural information processing systems},
  33:\penalty0 22243--22255, 2020{\natexlab{b}}.

\bibitem[Chen et~al.(2022{\natexlab{a}})Chen, Saxena, Li, Lin, Fleet, and
  Hinton]{chen2022unified}
Ting Chen, Saurabh Saxena, Lala Li, Tsung-Yi Lin, David~J Fleet, and Geoffrey~E
  Hinton.
\newblock A unified sequence interface for vision tasks.
\newblock \emph{Advances in Neural Information Processing Systems},
  35:\penalty0 31333--31346, 2022{\natexlab{a}}.

\bibitem[Chen et~al.(2022{\natexlab{b}})Chen, Wang, Changpinyo, Piergiovanni,
  Padlewski, Salz, Goodman, Grycner, Mustafa, Beyer, et~al.]{chen2022pali}
Xi~Chen, Xiao Wang, Soravit Changpinyo, AJ~Piergiovanni, Piotr Padlewski,
  Daniel Salz, Sebastian Goodman, Adam Grycner, Basil Mustafa, Lucas Beyer,
  et~al.
\newblock Pali: A jointly-scaled multilingual language-image model.
\newblock \emph{arXiv preprint arXiv:2209.06794}, 2022{\natexlab{b}}.

\bibitem[Chen et~al.(2015)Chen, Fang, Lin, Vedantam, Gupta, Doll{\'a}r, and
  Zitnick]{chen2015microsoft}
Xinlei Chen, Hao Fang, Tsung-Yi Lin, Ramakrishna Vedantam, Saurabh Gupta, Piotr
  Doll{\'a}r, and C~Lawrence Zitnick.
\newblock Microsoft coco captions: Data collection and evaluation server.
\newblock \emph{arXiv preprint arXiv:1504.00325}, 2015.

\bibitem[Chen et~al.(2020{\natexlab{c}})Chen, Li, Yu, El~Kholy, Ahmed, Gan,
  Cheng, and Liu]{chen2020uniter}
Yen-Chun Chen, Linjie Li, Licheng Yu, Ahmed El~Kholy, Faisal Ahmed, Zhe Gan,
  Yu~Cheng, and Jingjing Liu.
\newblock Uniter: Universal image-text representation learning.
\newblock In \emph{Computer Vision--ECCV 2020: 16th European Conference,
  Glasgow, UK, August 23--28, 2020, Proceedings, Part XXX}, pages 104--120.
  Springer, 2020{\natexlab{c}}.

\bibitem[Cheng et~al.(2021)Cheng, Choudhuri, Misra, Kirillov, Girdhar, and
  Schwing]{cheng2021mask2former}
Bowen Cheng, Anwesa Choudhuri, Ishan Misra, Alexander Kirillov, Rohit Girdhar,
  and Alexander~G Schwing.
\newblock Mask2former for video instance segmentation.
\newblock \emph{arXiv preprint arXiv:2112.10764}, 2021.

\bibitem[Cheng et~al.(2014)Cheng, Han, Zhou, and Guo]{cheng2014multi}
Gong Cheng, Junwei Han, Peicheng Zhou, and Lei Guo.
\newblock Multi-class geospatial object detection and geographic image
  classification based on collection of part detectors.
\newblock \emph{ISPRS Journal of Photogrammetry and Remote Sensing},
  98:\penalty0 119--132, 2014.

\bibitem[Cheng and Schwing(2022)]{cheng2022xmem}
Ho~Kei Cheng and Alexander~G Schwing.
\newblock Xmem: Long-term video object segmentation with an atkinson-shiffrin
  memory model.
\newblock In \emph{Computer Vision--ECCV 2022: 17th European Conference, Tel
  Aviv, Israel, October 23--27, 2022, Proceedings, Part XXVIII}, pages
  640--658. Springer, 2022.

\bibitem[Cheng et~al.(2023)Cheng, Li, Xu, Li, Yang, Wang, and
  Yang]{cheng2023segment}
Yangming Cheng, Liulei Li, Yuanyou Xu, Xiaodi Li, Zongxin Yang, Wenguan Wang,
  and Yi~Yang.
\newblock Segment and track anything.
\newblock \emph{arXiv preprint arXiv:2305.06558}, 2023.

\bibitem[Cherti et~al.(2023)Cherti, Beaumont, Wightman, Wortsman, Ilharco,
  Gordon, Schuhmann, Schmidt, and Jitsev]{cherti2023reproducible}
Mehdi Cherti, Romain Beaumont, Ross Wightman, Mitchell Wortsman, Gabriel
  Ilharco, Cade Gordon, Christoph Schuhmann, Ludwig Schmidt, and Jenia Jitsev.
\newblock Reproducible scaling laws for contrastive language-image learning.
\newblock In \emph{Proceedings of the IEEE/CVF Conference on Computer Vision
  and Pattern Recognition}, pages 2818--2829, 2023.

\bibitem[Chiang et~al.(2023)Chiang, Li, Lin, Sheng, Wu, Zhang, Zheng, Zhuang,
  Zhuang, Gonzalez, Stoica, and Xing]{vicuna2023}
Wei-Lin Chiang, Zhuohan Li, Zi~Lin, Ying Sheng, Zhanghao Wu, Hao Zhang, Lianmin
  Zheng, Siyuan Zhuang, Yonghao Zhuang, Joseph~E. Gonzalez, Ion Stoica, and
  Eric~P. Xing.
\newblock Vicuna: An open-source chatbot impressing gpt-4 with 90\%* chatgpt
  quality, March 2023.
\newblock URL \url{https://lmsys.org/blog/2023-03-30-vicuna/}.

\bibitem[Cho et~al.(2021)Cho, Lei, Tan, and Bansal]{cho2021unifying}
Jaemin Cho, Jie Lei, Hao Tan, and Mohit Bansal.
\newblock Unifying vision-and-language tasks via text generation.
\newblock In \emph{International Conference on Machine Learning}, pages
  1931--1942. PMLR, 2021.

\bibitem[Chowdhery et~al.(2022)Chowdhery, Narang, Devlin, Bosma, Mishra,
  Roberts, Barham, Chung, Sutton, Gehrmann, et~al.]{chowdhery2022palm}
Aakanksha Chowdhery, Sharan Narang, Jacob Devlin, Maarten Bosma, Gaurav Mishra,
  Adam Roberts, Paul Barham, Hyung~Won Chung, Charles Sutton, Sebastian
  Gehrmann, et~al.
\newblock Palm: Scaling language modeling with pathways.
\newblock \emph{arXiv preprint arXiv:2204.02311}, 2022.

\bibitem[Christiano et~al.(2017)Christiano, Leike, Brown, Martic, Legg, and
  Amodei]{christiano2017deep}
Paul~F Christiano, Jan Leike, Tom Brown, Miljan Martic, Shane Legg, and Dario
  Amodei.
\newblock Deep reinforcement learning from human preferences.
\newblock \emph{Advances in neural information processing systems}, 30, 2017.

\bibitem[Chung et~al.(2022)Chung, Hou, Longpre, Zoph, Tay, Fedus, Li, Wang,
  Dehghani, Brahma, et~al.]{chung2022scaling}
Hyung~Won Chung, Le~Hou, Shayne Longpre, Barret Zoph, Yi~Tay, William Fedus,
  Eric Li, Xuezhi Wang, Mostafa Dehghani, Siddhartha Brahma, et~al.
\newblock Scaling instruction-finetuned language models.
\newblock \emph{arXiv preprint arXiv:2210.11416}, 2022.

\bibitem[Conti et~al.(2018)Conti, Madhavan, Petroski~Such, Lehman, Stanley, and
  Clune]{conti2018improving}
Edoardo Conti, Vashisht Madhavan, Felipe Petroski~Such, Joel Lehman, Kenneth
  Stanley, and Jeff Clune.
\newblock Improving exploration in evolution strategies for deep reinforcement
  learning via a population of novelty-seeking agents.
\newblock \emph{Advances in neural information processing systems}, 31, 2018.

\bibitem[Croitoru et~al.(2023)Croitoru, Hondru, Ionescu, and
  Shah]{croitoru2023diffusion}
Florinel-Alin Croitoru, Vlad Hondru, Radu~Tudor Ionescu, and Mubarak Shah.
\newblock Diffusion models in vision: A survey.
\newblock \emph{IEEE Transactions on Pattern Analysis and Machine
  Intelligence}, 2023.

\bibitem[Dai et~al.(2023)Dai, Li, Li, Tiong, Zhao, Wang, Li, Fung, and
  Hoi]{dai2023instructblip}
Wenliang Dai, Junnan Li, Dongxu Li, Anthony Meng~Huat Tiong, Junqi Zhao,
  Weisheng Wang, Boyang Li, Pascale Fung, and Steven Hoi.
\newblock Instructblip: Towards general-purpose vision-language models with
  instruction tuning.
\newblock \emph{arXiv preprint arXiv: 2305.06500}, 2023.

\bibitem[Das et~al.(2017)Das, Kottur, Gupta, Singh, Yadav, Moura, Parikh, and
  Batra]{das2017visual}
Abhishek Das, Satwik Kottur, Khushi Gupta, Avi Singh, Deshraj Yadav,
  Jos{\'e}~MF Moura, Devi Parikh, and Dhruv Batra.
\newblock Visual dialog.
\newblock In \emph{Proceedings of the IEEE conference on computer vision and
  pattern recognition}, pages 326--335, 2017.

\bibitem[Desai et~al.(2021)Desai, Kaul, Aysola, and
  Johnson]{DBLP:conf/nips/DesaiKA021}
Karan Desai, Gaurav Kaul, Zubin Aysola, and Justin Johnson.
\newblock Redcaps: Web-curated image-text data created by the people, for the
  people.
\newblock In Joaquin Vanschoren and Sai{-}Kit Yeung, editors, \emph{Proceedings
  of the Neural Information Processing Systems Track on Datasets and Benchmarks
  1, NeurIPS Datasets and Benchmarks 2021, December 2021, virtual}, 2021.
\newblock URL
  \url{https://datasets-benchmarks-proceedings.neurips.cc/paper/2021/hash/e00da03b685a0dd18fb6a08af0923de0-Abstract-round1.html}.

\bibitem[Dettmers et~al.(2023)Dettmers, Pagnoni, Holtzman, and
  Zettlemoyer]{dettmers2023qlora}
Tim Dettmers, Artidoro Pagnoni, Ari Holtzman, and Luke Zettlemoyer.
\newblock Qlora: Efficient finetuning of quantized llms.
\newblock \emph{arXiv preprint arXiv:2305.14314}, 2023.

\bibitem[Devlin et~al.(2018)Devlin, Chang, Lee, and Toutanova]{devlin2018bert}
Jacob Devlin, Ming-Wei Chang, Kenton Lee, and Kristina Toutanova.
\newblock Bert: Pre-training of deep bidirectional transformers for language
  understanding.
\newblock \emph{arXiv preprint arXiv:1810.04805}, 2018.

\bibitem[Ding et~al.(2022)Ding, Xiao, Codella, Luo, Wang, and
  Yuan]{ding2022davit}
Mingyu Ding, Bin Xiao, Noel Codella, Ping Luo, Jingdong Wang, and Lu~Yuan.
\newblock Davit: Dual attention vision transformers.
\newblock In \emph{European Conference on Computer Vision}, pages 74--92.
  Springer, 2022.

\bibitem[Ding et~al.(2023)Ding, Zhang, Paxton, and Zhang]{ding2023task}
Yan Ding, Xiaohan Zhang, Chris Paxton, and Shiqi Zhang.
\newblock Task and motion planning with large language models for object
  rearrangement.
\newblock \emph{arXiv preprint arXiv: 2303.06247}, 2023.

\bibitem[Djolonga et~al.(2021)Djolonga, Yung, Tschannen, Romijnders, Beyer,
  Kolesnikov, Puigcerver, Minderer, D'Amour, Moldovan,
  et~al.]{djolonga2021robustness}
Josip Djolonga, Jessica Yung, Michael Tschannen, Rob Romijnders, Lucas Beyer,
  Alexander Kolesnikov, Joan Puigcerver, Matthias Minderer, Alexander D'Amour,
  Dan Moldovan, et~al.
\newblock On robustness and transferability of convolutional neural networks.
\newblock In \emph{Proceedings of the IEEE/CVF Conference on Computer Vision
  and Pattern Recognition}, pages 16458--16468, 2021.

\bibitem[Dong et~al.(2022{\natexlab{a}})Dong, Li, Dai, Zheng, Wu, Chang, Sun,
  Xu, and Sui]{dong2022survey}
Qingxiu Dong, Lei Li, Damai Dai, Ce~Zheng, Zhiyong Wu, Baobao Chang, Xu~Sun,
  Jingjing Xu, and Zhifang Sui.
\newblock A survey for in-context learning.
\newblock \emph{arXiv preprint arXiv:2301.00234}, 2022{\natexlab{a}}.

\bibitem[Dong et~al.(2022{\natexlab{b}})Dong, Bao, Chen, Zhang, Yu, Yuan, Chen,
  and Guo]{dong2022cswin}
Xiaoyi Dong, Jianmin Bao, Dongdong Chen, Weiming Zhang, Nenghai Yu, Lu~Yuan,
  Dong Chen, and Baining Guo.
\newblock Cswin transformer: A general vision transformer backbone with
  cross-shaped windows.
\newblock In \emph{Proceedings of the IEEE/CVF Conference on Computer Vision
  and Pattern Recognition}, pages 12124--12134, 2022{\natexlab{b}}.

\bibitem[Dong et~al.(2023)Dong, Bao, Zheng, Zhang, Chen, Yang, Zeng, Zhang,
  Yuan, Chen, et~al.]{dong2023maskclip}
Xiaoyi Dong, Jianmin Bao, Yinglin Zheng, Ting Zhang, Dongdong Chen, Hao Yang,
  Ming Zeng, Weiming Zhang, Lu~Yuan, Dong Chen, et~al.
\newblock Maskclip: Masked self-distillation advances contrastive
  language-image pretraining.
\newblock In \emph{Proceedings of the IEEE/CVF Conference on Computer Vision
  and Pattern Recognition}, pages 10995--11005, 2023.

\bibitem[Dosovitskiy et~al.(2020)Dosovitskiy, Beyer, Kolesnikov, Weissenborn,
  Zhai, Unterthiner, Dehghani, Minderer, Heigold, Gelly,
  et~al.]{dosovitskiy2020image}
Alexey Dosovitskiy, Lucas Beyer, Alexander Kolesnikov, Dirk Weissenborn,
  Xiaohua Zhai, Thomas Unterthiner, Mostafa Dehghani, Matthias Minderer, Georg
  Heigold, Sylvain Gelly, et~al.
\newblock An image is worth 16x16 words: Transformers for image recognition at
  scale.
\newblock \emph{arXiv preprint arXiv:2010.11929}, 2020.

\bibitem[Dou et~al.(2022{\natexlab{a}})Dou, Kamath, Gan, Zhang, Wang, Li, Liu,
  Liu, LeCun, Peng, et~al.]{dou2022coarse}
Zi-Yi Dou, Aishwarya Kamath, Zhe Gan, Pengchuan Zhang, Jianfeng Wang, Linjie
  Li, Zicheng Liu, Ce~Liu, Yann LeCun, Nanyun Peng, et~al.
\newblock Coarse-to-fine vision-language pre-training with fusion in the
  backbone.
\newblock \emph{arXiv preprint arXiv:2206.07643}, 2022{\natexlab{a}}.

\bibitem[Dou et~al.(2022{\natexlab{b}})Dou, Xu, Gan, Wang, Wang, Wang, Zhu,
  Zhang, Yuan, Peng, et~al.]{dou2022empirical}
Zi-Yi Dou, Yichong Xu, Zhe Gan, Jianfeng Wang, Shuohang Wang, Lijuan Wang,
  Chenguang Zhu, Pengchuan Zhang, Lu~Yuan, Nanyun Peng, et~al.
\newblock An empirical study of training end-to-end vision-and-language
  transformers.
\newblock In \emph{Proceedings of the IEEE/CVF Conference on Computer Vision
  and Pattern Recognition}, pages 18166--18176, 2022{\natexlab{b}}.

\bibitem[Doveh et~al.(2023)Doveh, Arbelle, Harary, Alfassy, Herzig, Kim,
  Giryes, Feris, Panda, Ullman, et~al.]{doveh2023dense}
Sivan Doveh, Assaf Arbelle, Sivan Harary, Amit Alfassy, Roei Herzig, Donghyun
  Kim, Raja Giryes, Rogerio Feris, Rameswar Panda, Shimon Ullman, et~al.
\newblock Dense and aligned captions (dac) promote compositional reasoning in
  vl models.
\newblock \emph{arXiv preprint arXiv:2305.19595}, 2023.

\bibitem[Driess et~al.(2023)Driess, Xia, Sajjadi, Lynch, Chowdhery, Ichter,
  Wahid, Tompson, Vuong, Yu, Huang, Chebotar, Sermanet, Duckworth, Levine,
  Vanhoucke, Hausman, Toussaint, Greff, Zeng, Mordatch, and
  Florence]{driess2023palme}
Danny Driess, Fei Xia, Mehdi S.~M. Sajjadi, Corey Lynch, Aakanksha Chowdhery,
  Brian Ichter, Ayzaan Wahid, Jonathan Tompson, Quan Vuong, Tianhe Yu, Wenlong
  Huang, Yevgen Chebotar, Pierre Sermanet, Daniel Duckworth, Sergey Levine,
  Vincent Vanhoucke, Karol Hausman, Marc Toussaint, Klaus Greff, Andy Zeng,
  Igor Mordatch, and Pete Florence.
\newblock Palm-e: An embodied multimodal language model.
\newblock \emph{arXiv preprint arXiv: 2303.03378}, 2023.

\bibitem[Du et~al.(2022)Du, Liu, Li, and Zhao]{Yifan2022VLM}
Yifan Du, Zikang Liu, Junyi Li, and Wayne~Xin Zhao.
\newblock A survey of vision-language pre-trained models.
\newblock In \emph{IJCAI}, 2022.

\bibitem[Everingham et~al.(2015)Everingham, Eslami, Van~Gool, Williams, Winn,
  and Zisserman]{everingham2015pascal}
Mark Everingham, SM~Ali Eslami, Luc Van~Gool, Christopher~KI Williams, John
  Winn, and Andrew Zisserman.
\newblock The pascal visual object classes challenge: A retrospective.
\newblock \emph{International journal of computer vision}, 111:\penalty0
  98--136, 2015.

\bibitem[Eysenbach et~al.(2018)Eysenbach, Gupta, Ibarz, and
  Levine]{eysenbach2018diversity}
Benjamin Eysenbach, Abhishek Gupta, Julian Ibarz, and Sergey Levine.
\newblock Diversity is all you need: Learning skills without a reward function.
\newblock \emph{arXiv preprint arXiv:1802.06070}, 2018.

\bibitem[Fan et~al.(2022)Fan, Wang, Jiang, Mandlekar, Yang, Zhu, Tang, Huang,
  Zhu, and Anandkumar]{fan2022minedojo}
Linxi Fan, Guanzhi Wang, Yunfan Jiang, Ajay Mandlekar, Yuncong Yang, Haoyi Zhu,
  Andrew Tang, De-An Huang, Yuke Zhu, and Anima Anandkumar.
\newblock Minedojo: Building open-ended embodied agents with internet-scale
  knowledge.
\newblock \emph{arXiv preprint arXiv: 2206.08853}, 2022.

\bibitem[Fan et~al.(2023)Fan, Li, Ma, Lee, Yu, and
  Hemphill]{fan2023bibliometric}
Lizhou Fan, Lingyao Li, Zihui Ma, Sanggyu Lee, Huizi Yu, and Libby Hemphill.
\newblock A bibliometric review of large language models research from 2017 to
  2023.
\newblock \emph{arXiv preprint arXiv:2304.02020}, 2023.

\bibitem[Fang et~al.(2021)Fang, Xiong, Xu, and Chen]{fang2021clip2video}
Han Fang, Pengfei Xiong, Luhui Xu, and Yu~Chen.
\newblock Clip2video: Mastering video-text retrieval via image clip.
\newblock \emph{arXiv preprint arXiv: 2106.11097}, 2021.

\bibitem[Fang et~al.(2023{\natexlab{a}})Fang, Sun, Wang, Huang, Wang, and
  Cao]{fang2023eva2}
Yuxin Fang, Quan Sun, Xinggang Wang, Tiejun Huang, Xinlong Wang, and Yue Cao.
\newblock Eva-02: A visual representation for neon genesis.
\newblock \emph{arXiv preprint arXiv:2303.11331}, 2023{\natexlab{a}}.

\bibitem[Fang et~al.(2023{\natexlab{b}})Fang, Wang, Xie, Sun, Wu, Wang, Huang,
  Wang, and Cao]{fang2023eva}
Yuxin Fang, Wen Wang, Binhui Xie, Quan Sun, Ledell Wu, Xinggang Wang, Tiejun
  Huang, Xinlong Wang, and Yue Cao.
\newblock Eva: Exploring the limits of masked visual representation learning at
  scale.
\newblock In \emph{Proceedings of the IEEE/CVF Conference on Computer Vision
  and Pattern Recognition}, pages 19358--19369, 2023{\natexlab{b}}.

\bibitem[Fedorov et~al.(2012)Fedorov, Beichel, Kalpathy-Cramer, Finet,
  Fillion-Robin, Pujol, Bauer, Jennings, Fennessy, Sonka,
  et~al.]{fedorov20123d}
Andriy Fedorov, Reinhard Beichel, Jayashree Kalpathy-Cramer, Julien Finet,
  Jean-Christophe Fillion-Robin, Sonia Pujol, Christian Bauer, Dominique
  Jennings, Fiona Fennessy, Milan Sonka, et~al.
\newblock 3d slicer as an image computing platform for the quantitative imaging
  network.
\newblock \emph{Magnetic resonance imaging}, 30\penalty0 (9):\penalty0
  1323--1341, 2012.

\bibitem[Fei-Fei et~al.(2009)Fei-Fei, Deng, and Li]{fei2009imagenet}
Li~Fei-Fei, Jia Deng, and Kai Li.
\newblock Imagenet: Constructing a large-scale image database.
\newblock \emph{Journal of vision}, 9\penalty0 (8):\penalty0 1037--1037, 2009.

\bibitem[Gadre et~al.(2023)Gadre, Ilharco, Fang, Hayase, Smyrnis, Nguyen,
  Marten, Wortsman, Ghosh, Zhang, et~al.]{gadre2023datacomp}
Samir~Yitzhak Gadre, Gabriel Ilharco, Alex Fang, Jonathan Hayase, Georgios
  Smyrnis, Thao Nguyen, Ryan Marten, Mitchell Wortsman, Dhruba Ghosh, Jieyu
  Zhang, et~al.
\newblock Datacomp: In search of the next generation of multimodal datasets.
\newblock \emph{arXiv preprint arXiv:2304.14108}, 2023.

\bibitem[Gan et~al.(2022)Gan, Li, Li, Wang, Liu, Gao, et~al.]{gan2022vision}
Zhe Gan, Linjie Li, Chunyuan Li, Lijuan Wang, Zicheng Liu, Jianfeng Gao, et~al.
\newblock Vision-language pre-training: Basics, recent advances, and future
  trends.
\newblock \emph{Foundations and Trends{\textregistered} in Computer Graphics
  and Vision}, 14\penalty0 (3--4):\penalty0 163--352, 2022.

\bibitem[Gao et~al.(2023{\natexlab{a}})Gao, Han, Zhang, Lin, Geng, Zhou, Zhang,
  Lu, He, Yue, Li, and Qiao]{gao2023llamaadapterv2}
Peng Gao, Jiaming Han, Renrui Zhang, Ziyi Lin, Shijie Geng, Aojun Zhou, Wei
  Zhang, Pan Lu, Conghui He, Xiangyu Yue, Hongsheng Li, and Yu~Qiao.
\newblock Llama-adapter v2: Parameter-efficient visual instruction model.
\newblock \emph{arXiv preprint arXiv:2304.15010}, 2023{\natexlab{a}}.

\bibitem[Gao et~al.(2021)Gao, Fisch, and Chen]{gao2021making}
Tianyu Gao, Adam Fisch, and Danqi Chen.
\newblock Making pre-trained language models better few-shot learners, 2021.

\bibitem[Gao et~al.(2023{\natexlab{b}})Gao, Xia, Hu, and Gao]{gao2023desam}
Yifan Gao, Wei Xia, Dingdu Hu, and Xin Gao.
\newblock Desam: Decoupling segment anything model for generalizable medical
  image segmentation.
\newblock \emph{arXiv preprint arXiv:2306.00499}, 2023{\natexlab{b}}.

\bibitem[Ge et~al.(2023)Ge, Hua, Mei, Ji, Tan, Xu, Li, and
  Zhang]{ge2023openagi}
Yingqiang Ge, Wenyue Hua, Kai Mei, Jianchao Ji, Juntao Tan, Shuyuan Xu, Zelong
  Li, and Yongfeng Zhang.
\newblock Openagi: When llm meets domain experts.
\newblock \emph{arXiv preprint arXiv: 2304.04370}, 2023.

\bibitem[Gemmeke et~al.(2017)Gemmeke, Ellis, Freedman, Jansen, Lawrence, Moore,
  Plakal, and Ritter]{gemmeke2017audio}
Jort~F Gemmeke, Daniel~PW Ellis, Dylan Freedman, Aren Jansen, Wade Lawrence,
  R~Channing Moore, Manoj Plakal, and Marvin Ritter.
\newblock Audio set: An ontology and human-labeled dataset for audio events.
\newblock In \emph{2017 IEEE international conference on acoustics, speech and
  signal processing (ICASSP)}, pages 776--780. IEEE, 2017.

\bibitem[Ghiasi et~al.(2021{\natexlab{a}})Ghiasi, Gu, Cui, and
  Lin]{ghiasi2021scaling}
Golnaz Ghiasi, Xiuye Gu, Yin Cui, and Tsung-Yi Lin.
\newblock Scaling open-vocabulary image segmentation with image-level labels.
\newblock \emph{arXiv preprint arXiv: 2112.12143}, 2021{\natexlab{a}}.

\bibitem[Ghiasi et~al.(2021{\natexlab{b}})Ghiasi, Zoph, Cubuk, Le, and
  Lin]{ghiasi2021multi}
Golnaz Ghiasi, Barret Zoph, Ekin~D Cubuk, Quoc~V Le, and Tsung-Yi Lin.
\newblock Multi-task self-training for learning general representations.
\newblock In \emph{Proceedings of the IEEE/CVF International Conference on
  Computer Vision}, pages 8856--8865, 2021{\natexlab{b}}.

\bibitem[Girdhar et~al.(2023)Girdhar, El-Nouby, Liu, Singh, Alwala, Joulin, and
  Misra]{girdhar2023imagebind}
Rohit Girdhar, Alaaeldin El-Nouby, Zhuang Liu, Mannat Singh, Kalyan~Vasudev
  Alwala, Armand Joulin, and Ishan Misra.
\newblock Imagebind: One embedding space to bind them all.
\newblock \emph{CVPR}, 2023.

\bibitem[Gong et~al.(2023)Gong, Zhong, Ma, Li, Wang, Zhang, Heng, and
  Dou]{Gong20233DSAMadapterHA}
Shizhan Gong, Yuan Zhong, Wenao Ma, Jinpeng Li, Zhao Wang, Jingyang Zhang,
  Pheng-Ann Heng, and Qi~Dou.
\newblock 3dsam-adapter: Holistic adaptation of sam from 2d to 3d for
  promptable medical image segmentation.
\newblock \emph{ArXiv}, abs/2306.13465, 2023.

\bibitem[Goyal et~al.(2016)Goyal, Khot, Summers-Stay, Batra, and
  Parikh]{goyal2016making}
Yash Goyal, Tejas Khot, Douglas Summers-Stay, Dhruv Batra, and Devi Parikh.
\newblock Making the v in vqa matter: Elevating the role of image understanding
  in visual question answering.
\newblock \emph{Computer Vision And Pattern Recognition}, 2016.
\newblock \doi{10.1007/s11263-018-1116-0}.

\bibitem[Grauman et~al.(2022)Grauman, Westbury, Byrne, Chavis, Furnari,
  Girdhar, Hamburger, Jiang, Liu, Liu, et~al.]{grauman2022ego4d}
Kristen Grauman, Andrew Westbury, Eugene Byrne, Zachary Chavis, Antonino
  Furnari, Rohit Girdhar, Jackson Hamburger, Hao Jiang, Miao Liu, Xingyu Liu,
  et~al.
\newblock Ego4d: Around the world in 3,000 hours of egocentric video.
\newblock In \emph{Proceedings of the IEEE/CVF Conference on Computer Vision
  and Pattern Recognition}, pages 18995--19012, 2022.

\bibitem[Greshake et~al.(2023)Greshake, Abdelnabi, Mishra, Endres, Holz, and
  Fritz]{greshake2023more}
Kai Greshake, Sahar Abdelnabi, Shailesh Mishra, Christoph Endres, Thorsten
  Holz, and Mario Fritz.
\newblock More than you've asked for: A comprehensive analysis of novel prompt
  injection threats to application-integrated large language models.
\newblock \emph{arXiv preprint arXiv:2302.12173}, 2023.

\bibitem[Gu et~al.(2021)Gu, Lin, Kuo, and Cui]{gu2021open}
Xiuye Gu, Tsung-Yi Lin, Weicheng Kuo, and Yin Cui.
\newblock Open-vocabulary object detection via vision and language knowledge
  distillation.
\newblock \emph{arXiv preprint arXiv:2104.13921}, 2021.

\bibitem[Guo et~al.(2023)Guo, Tang, Zhang, Wang, Wang, Zhao, and
  Li]{guo2023viewrefer}
Ziyu Guo, Yiwen Tang, Renrui Zhang, Dong Wang, Zhigang Wang, Bin Zhao, and
  Xuelong Li.
\newblock Viewrefer: Grasp the multi-view knowledge for 3d visual grounding
  with gpt and prototype guidance.
\newblock \emph{arXiv preprint arXiv: 2303.16894}, 2023.

\bibitem[Gupta et~al.(2019)Gupta, Dollar, and Girshick]{gupta2019lvis}
Agrim Gupta, Piotr Dollar, and Ross Girshick.
\newblock Lvis: A dataset for large vocabulary instance segmentation.
\newblock In \emph{Proceedings of the IEEE/CVF conference on computer vision
  and pattern recognition}, pages 5356--5364, 2019.

\bibitem[Gupta and Kembhavi(2023)]{gupta2023visual}
Tanmay Gupta and Aniruddha Kembhavi.
\newblock Visual programming: Compositional visual reasoning without training.
\newblock In \emph{Proceedings of the IEEE/CVF Conference on Computer Vision
  and Pattern Recognition}, pages 14953--14962, 2023.

\bibitem[Gurari et~al.(2018)Gurari, Li, Stangl, Guo, Lin, Grauman, Luo, and
  Bigham]{gurari2018vizwiz}
Danna Gurari, Qing Li, Abigale~J Stangl, Anhong Guo, Chi Lin, Kristen Grauman,
  Jiebo Luo, and Jeffrey~P Bigham.
\newblock Vizwiz grand challenge: Answering visual questions from blind people.
\newblock In \emph{Proceedings of the IEEE conference on computer vision and
  pattern recognition}, pages 3608--3617, 2018.

\bibitem[Guzhov et~al.(2021)Guzhov, Raue, Hees, and
  Dengel]{guzhov2021audioclip}
Andrey Guzhov, Federico Raue, Jörn Hees, and Andreas Dengel.
\newblock Audioclip: Extending clip to image, text and audio.
\newblock \emph{arXiv preprint arXiv: 2106.13043}, 2021.

\bibitem[Hall et~al.(2023)Hall, Abrantes, Zhu, Sodunke, Shtedritski, and
  Kirk]{hall2023visogender}
Siobhan~Mackenzie Hall, Fernanda~Gon{\c{c}}alves Abrantes, Hanwen Zhu, Grace
  Sodunke, Aleksandar Shtedritski, and Hannah~Rose Kirk.
\newblock Visogender: A dataset for benchmarking gender bias in image-text
  pronoun resolution.
\newblock \emph{arXiv preprint arXiv:2306.12424}, 2023.

\bibitem[Hao et~al.(2022)Hao, Song, Dong, Huang, Chi, Wang, Ma, and
  Wei]{hao2022language}
Yaru Hao, Haoyu Song, Li~Dong, Shaohan Huang, Zewen Chi, Wenhui Wang, Shuming
  Ma, and Furu Wei.
\newblock Language models are general-purpose interfaces.
\newblock \emph{arXiv preprint arXiv: 2206.06336}, 2022.

\bibitem[Harley et~al.(2022)Harley, Fang, and Fragkiadaki]{harley2022particle}
Adam~W Harley, Zhaoyuan Fang, and Katerina Fragkiadaki.
\newblock Particle video revisited: Tracking through occlusions using point
  trajectories.
\newblock In \emph{European Conference on Computer Vision}, pages 59--75.
  Springer, 2022.

\bibitem[He et~al.(2016)He, Zhang, Ren, and Sun]{he2016deep}
Kaiming He, Xiangyu Zhang, Shaoqing Ren, and Jian Sun.
\newblock Deep residual learning for image recognition.
\newblock In \emph{Proceedings of the IEEE conference on computer vision and
  pattern recognition}, pages 770--778, 2016.

\bibitem[He et~al.(2017)He, Gkioxari, Doll{\'a}r, and Girshick]{he2017mask}
Kaiming He, Georgia Gkioxari, Piotr Doll{\'a}r, and Ross Girshick.
\newblock Mask r-cnn.
\newblock In \emph{Proceedings of the IEEE international conference on computer
  vision}, pages 2961--2969, 2017.

\bibitem[He et~al.(2020)He, Fan, Wu, Xie, and Girshick]{he2020momentum}
Kaiming He, Haoqi Fan, Yuxin Wu, Saining Xie, and Ross Girshick.
\newblock Momentum contrast for unsupervised visual representation learning.
\newblock In \emph{Proceedings of the IEEE/CVF conference on computer vision
  and pattern recognition}, pages 9729--9738, 2020.

\bibitem[He et~al.(2022)He, Chen, Xie, Li, Doll{\'a}r, and
  Girshick]{he2022masked}
Kaiming He, Xinlei Chen, Saining Xie, Yanghao Li, Piotr Doll{\'a}r, and Ross
  Girshick.
\newblock Masked autoencoders are scalable vision learners.
\newblock In \emph{Proceedings of the IEEE/CVF Conference on Computer Vision
  and Pattern Recognition}, pages 16000--16009, 2022.

\bibitem[Heller et~al.(2021)Heller, Isensee, Maier-Hein, Hou, Xie, Li, Nan, Mu,
  Lin, Han, et~al.]{heller2021state}
Nicholas Heller, Fabian Isensee, Klaus~H Maier-Hein, Xiaoshuai Hou, Chunmei
  Xie, Fengyi Li, Yang Nan, Guangrui Mu, Zhiyong Lin, Miofei Han, et~al.
\newblock The state of the art in kidney and kidney tumor segmentation in
  contrast-enhanced ct imaging: Results of the kits19 challenge.
\newblock \emph{Medical image analysis}, 67:\penalty0 101821, 2021.

\bibitem[Hendrycks et~al.(2020)Hendrycks, Burns, Basart, Zou, Mazeika, Song,
  and Steinhardt]{hendrycks2020measuring}
Dan Hendrycks, Collin Burns, Steven Basart, Andy Zou, Mantas Mazeika, Dawn
  Song, and Jacob Steinhardt.
\newblock Measuring massive multitask language understanding.
\newblock \emph{arXiv preprint arXiv:2009.03300}, 2020.

\bibitem[Hendrycks et~al.(2021)Hendrycks, Basart, Mu, Kadavath, Wang, Dorundo,
  Desai, Zhu, Parajuli, Guo, et~al.]{hendrycks2021many}
Dan Hendrycks, Steven Basart, Norman Mu, Saurav Kadavath, Frank Wang, Evan
  Dorundo, Rahul Desai, Tyler Zhu, Samyak Parajuli, Mike Guo, et~al.
\newblock The many faces of robustness: A critical analysis of
  out-of-distribution generalization.
\newblock In \emph{Proceedings of the IEEE/CVF International Conference on
  Computer Vision}, pages 8340--8349, 2021.

\bibitem[Hoffmann et~al.(2022)Hoffmann, Borgeaud, Mensch, Buchatskaya, Cai,
  Rutherford, Casas, Hendricks, Welbl, Clark, et~al.]{hoffmann2022training}
Jordan Hoffmann, Sebastian Borgeaud, Arthur Mensch, Elena Buchatskaya, Trevor
  Cai, Eliza Rutherford, Diego de~Las Casas, Lisa~Anne Hendricks, Johannes
  Welbl, Aidan Clark, et~al.
\newblock Training compute-optimal large language models.
\newblock \emph{arXiv preprint arXiv:2203.15556}, 2022.

\bibitem[Hossain et~al.(2019)Hossain, Sohel, Shiratuddin, and
  Laga]{hossain2019comprehensive}
MD~Zakir Hossain, Ferdous Sohel, Mohd~Fairuz Shiratuddin, and Hamid Laga.
\newblock A comprehensive survey of deep learning for image captioning.
\newblock \emph{ACM Computing Surveys (CsUR)}, 51\penalty0 (6):\penalty0 1--36,
  2019.

\bibitem[Hu(2023)]{hu2023neural}
Anthony Hu.
\newblock Neural world models for computer vision.
\newblock \emph{arXiv preprint arXiv:2306.09179}, 2023.

\bibitem[Hu et~al.(2021)Hu, Shen, Wallis, Allen-Zhu, Li, Wang, and
  Chen]{hu2021lora}
Edward~J. Hu, Yelong Shen, Phillip Wallis, Zeyuan Allen-Zhu, Yuanzhi Li, Shean
  Wang, and Weizhu Chen.
\newblock Lora: Low-rank adaptation of large language models.
\newblock \emph{International Conference On Learning Representations}, 2021.

\bibitem[Hu et~al.(2023)Hu, Pan, Li, and Yang]{hu2023advancing}
Mingzhe Hu, Shaoyan Pan, Yuheng Li, and Xiaofeng Yang.
\newblock Advancing medical imaging with language models: A journey from
  n-grams to chatgpt.
\newblock \emph{arXiv preprint arXiv: 2304.04920}, 2023.

\bibitem[Hu et~al.(2016)Hu, Rohrbach, and Darrell]{hu2016segmentation}
Ronghang Hu, Marcus Rohrbach, and Trevor Darrell.
\newblock Segmentation from natural language expressions.
\newblock In \emph{Computer Vision--ECCV 2016: 14th European Conference,
  Amsterdam, The Netherlands, October 11--14, 2016, Proceedings, Part I 14},
  pages 108--124. Springer, 2016.

\bibitem[Huang and Chang(2023)]{Huang2023ACL}
Jie Huang and Kevin Chen-Chuan Chang.
\newblock Towards reasoning in large language models: A survey.
\newblock \emph{ACL Findings}, 2023.

\bibitem[Huang et~al.(2023{\natexlab{a}})Huang, Dong, Wang, Hao, Singhal, Ma,
  Lv, Cui, Mohammed, Patra, Liu, Aggarwal, Chi, Bjorck, Chaudhary, Som, Song,
  and Wei]{huang2023language}
Shaohan Huang, Li~Dong, Wenhui Wang, Yaru Hao, Saksham Singhal, Shuming Ma,
  Tengchao Lv, Lei Cui, Owais~Khan Mohammed, Barun Patra, Qiang Liu, Kriti
  Aggarwal, Zewen Chi, Johan Bjorck, Vishrav Chaudhary, Subhojit Som, Xia Song,
  and Furu Wei.
\newblock Language is not all you need: Aligning perception with language
  models.
\newblock \emph{arXiv preprint arXiv: 2302.14045}, 2023{\natexlab{a}}.

\bibitem[Huang et~al.(2023{\natexlab{b}})Huang, Cao, Li, Juefei-Xu, Lin, Tsang,
  Liu, and Guo]{huang2023robustness}
Yihao Huang, Yue Cao, Tianlin Li, Felix Juefei-Xu, Di~Lin, Ivor~W Tsang, Yang
  Liu, and Qing Guo.
\newblock On the robustness of segment anything.
\newblock \emph{arXiv preprint arXiv:2305.16220}, 2023{\natexlab{b}}.

\bibitem[Hudson and Manning(2019)]{hudson2019gqa}
Drew~A Hudson and Christopher~D Manning.
\newblock Gqa: A new dataset for real-world visual reasoning and compositional
  question answering.
\newblock In \emph{Proceedings of the IEEE/CVF conference on computer vision
  and pattern recognition}, pages 6700--6709, 2019.

\bibitem[Huo et~al.(2021)Huo, Zhang, Liu, Lu, Gao, Yang, Wen, Zhang, Xu, Zheng,
  et~al.]{huo2021wenlan}
Yuqi Huo, Manli Zhang, Guangzhen Liu, Haoyu Lu, Yizhao Gao, Guoxing Yang,
  Jingyuan Wen, Heng Zhang, Baogui Xu, Weihao Zheng, et~al.
\newblock Wenlan: Bridging vision and language by large-scale multi-modal
  pre-training.
\newblock \emph{arXiv preprint arXiv:2103.06561}, 2021.

\bibitem[Ilharco et~al.(2021)Ilharco, Wortsman, Wightman, Gordon, Carlini,
  Taori, Dave, Shankar, Namkoong, Miller, Hajishirzi, Farhadi, and
  Schmidt]{ilharco_gabriel_2021_5143773}
Gabriel Ilharco, Mitchell Wortsman, Ross Wightman, Cade Gordon, Nicholas
  Carlini, Rohan Taori, Achal Dave, Vaishaal Shankar, Hongseok Namkoong, John
  Miller, Hannaneh Hajishirzi, Ali Farhadi, and Ludwig Schmidt.
\newblock Openclip, July 2021.
\newblock URL \url{https://doi.org/10.5281/zenodo.5143773}.
\newblock If you use this software, please cite it as below.

\bibitem[Jain et~al.(2023)Jain, Saifullah, Wen, Kirchenbauer, Shu, Saha,
  Goldblum, Geiping, and Goldstein]{jain2023bring}
Neel Jain, Khalid Saifullah, Yuxin Wen, John Kirchenbauer, Manli Shu, Aniruddha
  Saha, Micah Goldblum, Jonas Geiping, and Tom Goldstein.
\newblock Bring your own data! self-supervised evaluation for large language
  models, 2023.

\bibitem[Jha et~al.(2020)Jha, Smedsrud, Riegler, Halvorsen, de~Lange, Johansen,
  and Johansen]{jha2020kvasir}
Debesh Jha, Pia~H Smedsrud, Michael~A Riegler, P{\aa}l Halvorsen, Thomas
  de~Lange, Dag Johansen, and H{\aa}vard~D Johansen.
\newblock Kvasir-seg: A segmented polyp dataset.
\newblock In \emph{MultiMedia Modeling: 26th International Conference, MMM
  2020, Daejeon, South Korea, January 5--8, 2020, Proceedings, Part II 26},
  pages 451--462. Springer, 2020.

\bibitem[Ji et~al.(2022)Ji, Xiao, Chou, Fan, Zhao, Chen, and
  Van~Gool]{ji2022video}
Ge-Peng Ji, Guobao Xiao, Yu-Cheng Chou, Deng-Ping Fan, Kai Zhao, Geng Chen, and
  Luc Van~Gool.
\newblock Video polyp segmentation: A deep learning perspective.
\newblock \emph{Machine Intelligence Research}, 19\penalty0 (6):\penalty0
  531--549, 2022.

\bibitem[Ji et~al.(2018)Ji, Wei, and Lu]{ji2018fully}
Shunping Ji, Shiqing Wei, and Meng Lu.
\newblock Fully convolutional networks for multisource building extraction from
  an open aerial and satellite imagery data set.
\newblock \emph{IEEE Transactions on geoscience and remote sensing},
  57\penalty0 (1):\penalty0 574--586, 2018.

\bibitem[Jia et~al.(2021{\natexlab{a}})Jia, Yang, Xia, Chen, Parekh, Pham, Le,
  Sung, Li, and Duerig]{jia2021scaling}
Chao Jia, Yinfei Yang, Ye~Xia, Yi-Ting Chen, Zarana Parekh, Hieu Pham, Quoc Le,
  Yun-Hsuan Sung, Zhen Li, and Tom Duerig.
\newblock Scaling up visual and vision-language representation learning with
  noisy text supervision.
\newblock In \emph{International Conference on Machine Learning}, pages
  4904--4916. PMLR, 2021{\natexlab{a}}.

\bibitem[Jia et~al.(2021{\natexlab{b}})Jia, Zhu, Li, Tang, and
  Zhou]{jia2021llvip}
Xinyu Jia, Chuang Zhu, Minzhen Li, Wenqi Tang, and Wenli Zhou.
\newblock Llvip: A visible-infrared paired dataset for low-light vision.
\newblock In \emph{Proceedings of the IEEE/CVF international conference on
  computer vision}, pages 3496--3504, 2021{\natexlab{b}}.

\bibitem[Jiang et~al.(2022)Jiang, Gupta, Zhang, Wang, Dou, Chen, Fei-Fei,
  Anandkumar, Zhu, and Fan]{jiang2022vima}
Yunfan Jiang, Agrim Gupta, Zichen Zhang, Guanzhi Wang, Yongqiang Dou, Yanjun
  Chen, Li~Fei-Fei, Anima Anandkumar, Yuke Zhu, and Linxi Fan.
\newblock Vima: General robot manipulation with multimodal prompts.
\newblock \emph{arXiv preprint arXiv: 2210.03094}, 2022.

\bibitem[Jingyi et~al.(2023)Jingyi, Jiaxing, Sheng, and Shijian]{Zhang2023VLMs}
Zhang Jingyi, Huang Jiaxing, Jin Sheng, and Lu~Shijian.
\newblock Vision-language models for vision tasks: A survey.
\newblock \emph{arXiv preprint arXiv:2304.00685}, 2023.

\bibitem[Jocher et~al.(2023)Jocher, Chaurasia, and Qiu]{Jocher2023yolo}
Glenn Jocher, Ayush Chaurasia, and Jing Qiu.
\newblock Yolo by ultralytics.
\newblock \url{https://github.com/ultralytics/ultralytics}, 2023.

\bibitem[Kamath et~al.(2021)Kamath, Singh, LeCun, Synnaeve, Misra, and
  Carion]{kamath2021mdetr}
Aishwarya Kamath, Mannat Singh, Yann LeCun, Gabriel Synnaeve, Ishan Misra, and
  Nicolas Carion.
\newblock Mdetr - modulated detection for end-to-end multi-modal understanding.
\newblock \emph{arXiv preprint arXiv: 2104.12763}, 2021.

\bibitem[Kazemzadeh et~al.(2014)Kazemzadeh, Ordonez, Matten, and
  Berg]{kazemzadeh2014referitgame}
Sahar Kazemzadeh, Vicente Ordonez, Mark Matten, and Tamara Berg.
\newblock Referitgame: Referring to objects in photographs of natural scenes.
\newblock In \emph{Proceedings of the 2014 conference on empirical methods in
  natural language processing (EMNLP)}, pages 787--798, 2014.

\bibitem[Khan et~al.(2022)Khan, Naseer, Hayat, Zamir, Khan, and
  Shah]{khan2022transformers}
Salman Khan, Muzammal Naseer, Munawar Hayat, Syed~Waqas Zamir, Fahad~Shahbaz
  Khan, and Mubarak Shah.
\newblock Transformers in vision: A survey.
\newblock \emph{ACM computing surveys (CSUR)}, 54\penalty0 (10s):\penalty0
  1--41, 2022.

\bibitem[Khoreva et~al.(2019)Khoreva, Rohrbach, and Schiele]{khoreva2019video}
Anna Khoreva, Anna Rohrbach, and Bernt Schiele.
\newblock Video object segmentation with language referring expressions.
\newblock In \emph{Computer Vision--ACCV 2018: 14th Asian Conference on
  Computer Vision, Perth, Australia, December 2--6, 2018, Revised Selected
  Papers, Part IV 14}, pages 123--141. Springer, 2019.

\bibitem[Kiela et~al.(2020)Kiela, Firooz, Mohan, Goswami, Singh, Ringshia, and
  Testuggine]{kiela2020hateful}
Douwe Kiela, Hamed Firooz, Aravind Mohan, Vedanuj Goswami, Amanpreet Singh,
  Pratik Ringshia, and Davide Testuggine.
\newblock The hateful memes challenge: Detecting hate speech in multimodal
  memes.
\newblock \emph{Advances in neural information processing systems},
  33:\penalty0 2611--2624, 2020.

\bibitem[Kim et~al.(2023)Kim, Lee, Kim, Jung, Park, Kim, Yun, Kil, Lee, and
  Park]{kim2023cream}
Geewook Kim, Hodong Lee, Daehee Kim, Haeji Jung, Sanghee Park, Yoonsik Kim,
  Sangdoo Yun, Taeho Kil, Bado Lee, and Seunghyun Park.
\newblock Cream: Visually-situated natural language understanding with
  contrastive reading model and frozen large language models.
\newblock \emph{arXiv preprint arXiv:2305.15080}, 2023.

\bibitem[Kirillov et~al.(2023)Kirillov, Mintun, Ravi, Mao, Rolland, Gustafson,
  Xiao, Whitehead, Berg, Lo, et~al.]{kirillov2023segment}
Alexander Kirillov, Eric Mintun, Nikhila Ravi, Hanzi Mao, Chloe Rolland, Laura
  Gustafson, Tete Xiao, Spencer Whitehead, Alexander~C Berg, Wan-Yen Lo, et~al.
\newblock Segment anything.
\newblock \emph{arXiv preprint arXiv:2304.02643}, 2023.

\bibitem[Koh et~al.(2023)Koh, Salakhutdinov, and Fried]{koh2023grounding}
Jing~Yu Koh, Ruslan Salakhutdinov, and Daniel Fried.
\newblock Grounding language models to images for multimodal inputs and
  outputs.
\newblock 2023.

\bibitem[Krasin et~al.(2017)Krasin, Duerig, Alldrin, Ferrari, Abu-El-Haija,
  Kuznetsova, Rom, Uijlings, Popov, Veit, et~al.]{krasin2017openimages}
Ivan Krasin, Tom Duerig, Neil Alldrin, Vittorio Ferrari, Sami Abu-El-Haija,
  Alina Kuznetsova, Hassan Rom, Jasper Uijlings, Stefan Popov, Andreas Veit,
  et~al.
\newblock Openimages: A public dataset for large-scale multi-label and
  multi-class image classification.
\newblock \emph{Dataset available from https://github. com/openimages},
  2\penalty0 (3):\penalty0 18, 2017.

\bibitem[Krishna et~al.(2017)Krishna, Zhu, Groth, Johnson, Hata, Kravitz, Chen,
  Kalantidis, Li, Shamma, et~al.]{krishna2017visual}
Ranjay Krishna, Yuke Zhu, Oliver Groth, Justin Johnson, Kenji Hata, Joshua
  Kravitz, Stephanie Chen, Yannis Kalantidis, Li-Jia Li, David~A Shamma, et~al.
\newblock Visual genome: Connecting language and vision using crowdsourced
  dense image annotations.
\newblock \emph{International journal of computer vision}, 123:\penalty0
  32--73, 2017.

\bibitem[Krizhevsky et~al.(2012)Krizhevsky, Sutskever, and
  Hinton]{krizhevsky2012imagenet}
Alex Krizhevsky, Ilya Sutskever, and Geoffrey~E Hinton.
\newblock Imagenet classification with deep convolutional neural networks.
\newblock \emph{Advances in neural information processing systems}, 25, 2012.

\bibitem[Kumar et~al.(2019)Kumar, Verma, Anand, Zhou, Onder, Tsougenis, Chen,
  Heng, Li, Hu, et~al.]{kumar2019multi}
Neeraj Kumar, Ruchika Verma, Deepak Anand, Yanning Zhou, Omer~Fahri Onder,
  Efstratios Tsougenis, Hao Chen, Pheng-Ann Heng, Jiahui Li, Zhiqiang Hu,
  et~al.
\newblock A multi-organ nucleus segmentation challenge.
\newblock \emph{IEEE transactions on medical imaging}, 39\penalty0
  (5):\penalty0 1380--1391, 2019.

\bibitem[Kwon et~al.(2022)Kwon, Cai, Ravichandran, Bas, Bhotika, and
  Soatto]{kwon2022masked}
Gukyeong Kwon, Zhaowei Cai, Avinash Ravichandran, Erhan Bas, Rahul Bhotika, and
  Stefano Soatto.
\newblock Masked vision and language modeling for multi-modal representation
  learning.
\newblock \emph{arXiv preprint arXiv:2208.02131}, 2022.

\bibitem[Kwon et~al.(2023)Kwon, Cai, Ravichandran, Bas, Bhotika, and
  Soatto]{kuwon2022masked}
Gukyeong Kwon, Zhaowei Cai, Avinash Ravichandran, Erhan Bas, Rahul Bhotika, and
  Stefano Soatto.
\newblock Masked vision and language modeling for multi-modal representation
  learning.
\newblock In \emph{The Eleventh International Conference on Learning
  Representations, {ICLR} 2023, Kigali, Rwanda, May 1-5, 2023}. OpenReview.net,
  2023.
\newblock URL \url{https://openreview.net/pdf?id=ZhuXksSJYWn}.

\bibitem[Lei et~al.(2021)Lei, Xu, Gu, Fu, Zhang, Zhang, and
  Wang]{lei2021contrastive}
Wenhui Lei, Wei Xu, Ran Gu, Hao Fu, Shaoting Zhang, Shichuan Zhang, and Guotai
  Wang.
\newblock Contrastive learning of relative position regression for one-shot
  object localization in 3d medical images.
\newblock In \emph{Medical Image Computing and Computer Assisted
  Intervention--MICCAI 2021: 24th International Conference, Strasbourg, France,
  September 27--October 1, 2021, Proceedings, Part II 24}, pages 155--165.
  Springer, 2021.

\bibitem[Lester et~al.(2021)Lester, Al-Rfou, and Constant]{lester2021power}
Brian Lester, Rami Al-Rfou, and Noah Constant.
\newblock The power of scale for parameter-efficient prompt tuning.
\newblock \emph{arXiv preprint arXiv:2104.08691}, 2021.

\bibitem[Lewis et~al.(2019)Lewis, Liu, Goyal, Ghazvininejad, Mohamed, Levy,
  Stoyanov, and Zettlemoyer]{lewis2019bart}
Mike Lewis, Yinhan Liu, Naman Goyal, Marjan Ghazvininejad, Abdelrahman Mohamed,
  Omer Levy, Ves Stoyanov, and Luke Zettlemoyer.
\newblock Bart: Denoising sequence-to-sequence pre-training for natural
  language generation, translation, and comprehension.
\newblock \emph{arXiv preprint arXiv:1910.13461}, 2019.

\bibitem[Li et~al.(2023{\natexlab{a}})Li, Fang, Yang, Wang, Ye, Zhao, and
  Zhang]{li2023evaluating}
Bo~Li, Gexiang Fang, Yang Yang, Quansen Wang, Wei Ye, Wen Zhao, and Shikun
  Zhang.
\newblock Evaluating chatgpt's information extraction capabilities: An
  assessment of performance, explainability, calibration, and faithfulness.
\newblock \emph{arXiv preprint arXiv:2304.11633}, 2023{\natexlab{a}}.

\bibitem[Li et~al.(2023{\natexlab{b}})Li, Zhang, Chen, Wang, Pu, Yang, Li, and
  Liu]{li2023mimic}
Bo~Li, Yuanhan Zhang, Liangyu Chen, Jinghao Wang, Fanyi Pu, Jingkang Yang,
  Chunyuan Li, and Ziwei Liu.
\newblock Mimic-it: Multi-modal in-context instruction tuning.
\newblock \emph{arXiv preprint arXiv:2306.05425}, 2023{\natexlab{b}}.

\bibitem[Li et~al.(2021)Li, Selvaraju, Gotmare, Joty, Xiong, and
  Hoi]{li2021align}
Junnan Li, Ramprasaath Selvaraju, Akhilesh Gotmare, Shafiq Joty, Caiming Xiong,
  and Steven Chu~Hong Hoi.
\newblock Align before fuse: Vision and language representation learning with
  momentum distillation.
\newblock \emph{Advances in neural information processing systems},
  34:\penalty0 9694--9705, 2021.

\bibitem[Li et~al.(2022{\natexlab{a}})Li, Li, Xiong, and Hoi]{li2022blip}
Junnan Li, Dongxu Li, Caiming Xiong, and Steven Hoi.
\newblock Blip: Bootstrapping language-image pre-training for unified
  vision-language understanding and generation.
\newblock In \emph{International Conference on Machine Learning}, pages
  12888--12900. PMLR, 2022{\natexlab{a}}.

\bibitem[Li et~al.(2023{\natexlab{c}})Li, Li, Savarese, and Hoi]{li2023blip}
Junnan Li, Dongxu Li, Silvio Savarese, and Steven Hoi.
\newblock Blip-2: Bootstrapping language-image pre-training with frozen image
  encoders and large language models.
\newblock \emph{arXiv preprint arXiv:2301.12597}, 2023{\natexlab{c}}.

\bibitem[Li et~al.(2022{\natexlab{b}})Li, Zhang, Zhang, Yang, Li, Zhong, Wang,
  Yuan, Zhang, Hwang, et~al.]{li2022grounded}
Liunian~Harold Li, Pengchuan Zhang, Haotian Zhang, Jianwei Yang, Chunyuan Li,
  Yiwu Zhong, Lijuan Wang, Lu~Yuan, Lei Zhang, Jenq-Neng Hwang, et~al.
\newblock Grounded language-image pre-training.
\newblock In \emph{Proceedings of the IEEE/CVF Conference on Computer Vision
  and Pattern Recognition}, pages 10965--10975, 2022{\natexlab{b}}.

\bibitem[Li and Liang(2021)]{li2021prefix}
Xiang~Lisa Li and Percy Liang.
\newblock Prefix-tuning: Optimizing continuous prompts for generation.
\newblock \emph{arXiv preprint arXiv:2101.00190}, 2021.

\bibitem[Li et~al.(2023{\natexlab{d}})Li, Wang, and Xie]{li2023clipav2}
Xianhang Li, Zeyu Wang, and Cihang Xie.
\newblock Clipa-v2: Scaling clip training with 81.1
  accuracy within a \$10,000 budget; an extra \$4,000 unlocks 81.8
\newblock \emph{arXiv preprint arXiv: 2306.15658}, 2023{\natexlab{d}}.

\bibitem[Li et~al.(2023{\natexlab{e}})Li, Wang, and Xie]{li2023inverse}
Xianhang Li, Zeyu Wang, and Cihang Xie.
\newblock An inverse scaling law for clip training.
\newblock \emph{arXiv preprint arXiv: 2305.07017}, 2023{\natexlab{e}}.

\bibitem[Li et~al.(2023{\natexlab{f}})Li, Fan, Hu, Feichtenhofer, and
  He]{li2023scaling}
Yanghao Li, Haoqi Fan, Ronghang Hu, Christoph Feichtenhofer, and Kaiming He.
\newblock Scaling language-image pre-training via masking.
\newblock In \emph{Proceedings of the IEEE/CVF Conference on Computer Vision
  and Pattern Recognition}, pages 23390--23400, 2023{\natexlab{f}}.

\bibitem[Li et~al.(2023{\natexlab{g}})Li, Du, Zhou, Wang, Zhao, and
  Wen]{li2023evaluatingb}
Yifan Li, Yifan Du, Kun Zhou, Jinpeng Wang, Wayne~Xin Zhao, and Ji-Rong Wen.
\newblock Evaluating object hallucination in large vision-language models.
\newblock \emph{arXiv preprint arXiv:2305.10355}, 2023{\natexlab{g}}.

\bibitem[Liangliang et~al.(2023)Liangliang, Haobo, Guangze, Changhong, and
  Jia]{sam_da}
Yao Liangliang, Zuo Haobo, Zheng Guangze, Fu~Changhong, and Pan Jia.
\newblock Sam-da: Uav tracks anything at night with sam-powered domain
  adaptation.
\newblock \emph{arXiv:2307.01024}, 2023.

\bibitem[Lin et~al.(2014)Lin, Maire, Belongie, Hays, Perona, Ramanan,
  Doll{\'a}r, and Zitnick]{lin2014microsoft}
Tsung-Yi Lin, Michael Maire, Serge Belongie, James Hays, Pietro Perona, Deva
  Ramanan, Piotr Doll{\'a}r, and C~Lawrence Zitnick.
\newblock Microsoft coco: Common objects in context.
\newblock In \emph{Computer Vision--ECCV 2014: 13th European Conference,
  Zurich, Switzerland, September 6-12, 2014, Proceedings, Part V 13}, pages
  740--755. Springer, 2014.

\bibitem[Lin et~al.(2017{\natexlab{a}})Lin, Doll{\'a}r, Girshick, He,
  Hariharan, and Belongie]{lin2017feature}
Tsung-Yi Lin, Piotr Doll{\'a}r, Ross Girshick, Kaiming He, Bharath Hariharan,
  and Serge Belongie.
\newblock Feature pyramid networks for object detection.
\newblock In \emph{Proceedings of the IEEE conference on computer vision and
  pattern recognition}, pages 2117--2125, 2017{\natexlab{a}}.

\bibitem[Lin et~al.(2017{\natexlab{b}})Lin, Goyal, Girshick, He, and
  Doll{\'a}r]{lin2017focal}
Tsung-Yi Lin, Priya Goyal, Ross Girshick, Kaiming He, and Piotr Doll{\'a}r.
\newblock Focal loss for dense object detection.
\newblock In \emph{Proceedings of the IEEE international conference on computer
  vision}, pages 2980--2988, 2017{\natexlab{b}}.

\bibitem[Lin et~al.(2023)Lin, Karlinsky, Shvetsova, Possegger, Kozinski, Panda,
  Feris, Kuehne, and Bischof]{lin2023match}
Wei Lin, Leonid Karlinsky, Nina Shvetsova, Horst Possegger, Mateusz Kozinski,
  Rameswar Panda, Rogerio Feris, Hilde Kuehne, and Horst Bischof.
\newblock Match, expand and improve: Unsupervised finetuning for zero-shot
  action recognition with language knowledge.
\newblock \emph{arXiv preprint arXiv:2303.08914}, 2023.

\bibitem[Lin and Chen(2023)]{lin2023llm}
Yen-Ting Lin and Yun-Nung Chen.
\newblock Llm-eval: Unified multi-dimensional automatic evaluation for
  open-domain conversations with large language models.
\newblock \emph{arXiv preprint arXiv:2305.13711}, 2023.

\bibitem[Liu et~al.(2023{\natexlab{a}})Liu, Chen, Guan, Zhou, Zhu, and
  Zhou]{liu2023remoteclip}
Fan Liu, Delong Chen, Zhangqingyun Guan, Xiaocong Zhou, Jiale Zhu, and Jun
  Zhou.
\newblock Remoteclip: A vision language foundation model for remote sensing.
\newblock \emph{arXiv preprint arXiv: 2306.11029}, 2023{\natexlab{a}}.

\bibitem[Liu et~al.(2023{\natexlab{b}})Liu, Li, Wu, and Lee]{liu2023llava}
Haotian Liu, Chunyuan Li, Qingyang Wu, and Yong~Jae Lee.
\newblock Visual instruction tuning.
\newblock 2023{\natexlab{b}}.

\bibitem[Liu et~al.(2023{\natexlab{c}})Liu, Li, Wu, and Lee]{liu2023visual}
Haotian Liu, Chunyuan Li, Qingyang Wu, and Yong~Jae Lee.
\newblock Visual instruction tuning.
\newblock \emph{arXiv preprint arXiv:2304.08485}, 2023{\natexlab{c}}.

\bibitem[Liu et~al.(2023{\natexlab{d}})Liu, Jin, Wang, Cheng, Dou, and
  Wen]{liu2023reta}
Jiongnan Liu, Jiajie Jin, Zihan Wang, Jiehan Cheng, Zhicheng Dou, and Ji-Rong
  Wen.
\newblock Reta-llm: A retrieval-augmented large language model toolkit.
\newblock \emph{arXiv preprint arXiv:2306.05212}, 2023{\natexlab{d}}.

\bibitem[Liu et~al.(2023{\natexlab{e}})Liu, Li, Calinon, and
  Chen]{liu2023softgpt}
Junjia Liu, Zhihao Li, Sylvain Calinon, and Fei Chen.
\newblock Softgpt: Learn goal-oriented soft object manipulation skills by
  generative pre-trained heterogeneous graph transformer.
\newblock \emph{arXiv preprint arXiv:2306.12677}, 2023{\natexlab{e}}.

\bibitem[Liu et~al.(2023{\natexlab{f}})Liu, Fan, Johns, Yu, Xiao, and
  Anandkumar]{liu2023prismer}
Shikun Liu, Linxi Fan, Edward Johns, Zhiding Yu, Chaowei Xiao, and Anima
  Anandkumar.
\newblock Prismer: A vision-language model with an ensemble of experts.
\newblock \emph{arXiv preprint arXiv: 2303.02506}, 2023{\natexlab{f}}.

\bibitem[Liu et~al.()Liu, Zeng, Ren, Li, Zhang, Yang, Li, Yang, Su, Zhu,
  et~al.]{liu2023grounding}
Shilong Liu, Zhaoyang Zeng, Tianhe Ren, Feng Li, Hao Zhang, Jie Yang, Chunyuan
  Li, Jianwei Yang, Hang Su, Jun Zhu, et~al.
\newblock Grounding dino: Marrying dino with grounded pre-training for open-set
  object detection.
\newblock \emph{arXiv preprint arXiv:2303.05499}.

\bibitem[Liu and Zuo(2023)]{liu2023stone}
Weihua Liu and Yong Zuo.
\newblock Stone needle: A general multimodal large-scale model framework
  towards healthcare.
\newblock \emph{arXiv preprint arXiv: 2306.16034}, 2023.

\bibitem[Liu et~al.(2021{\natexlab{a}})Liu, Ji, Fu, Tam, Du, Yang, and
  Tang]{liu2021p}
Xiao Liu, Kaixuan Ji, Yicheng Fu, Weng~Lam Tam, Zhengxiao Du, Zhilin Yang, and
  Jie Tang.
\newblock P-tuning v2: Prompt tuning can be comparable to fine-tuning
  universally across scales and tasks.
\newblock \emph{arXiv preprint arXiv:2110.07602}, 2021{\natexlab{a}}.

\bibitem[Liu et~al.(2023{\natexlab{g}})Liu, Zhang, She, Kheradmand, and
  Armand]{liu2023samm}
Yihao Liu, Jiaming Zhang, Zhangcong She, Amir Kheradmand, and Mehran Armand.
\newblock Samm (segment any medical model): A 3d slicer integration to sam.
\newblock \emph{arXiv preprint arXiv:2304.05622}, 2023{\natexlab{g}}.

\bibitem[Liu et~al.(2019)Liu, Ott, Goyal, Du, Joshi, Chen, Levy, Lewis,
  Zettlemoyer, and Stoyanov]{liu2019roberta}
Yinhan Liu, Myle Ott, Naman Goyal, Jingfei Du, Mandar Joshi, Danqi Chen, Omer
  Levy, Mike Lewis, Luke Zettlemoyer, and Veselin Stoyanov.
\newblock Roberta: A robustly optimized bert pretraining approach, 2019.

\bibitem[Liu et~al.(2021{\natexlab{b}})Liu, Lin, Cao, Hu, Wei, Zhang, Lin, and
  Guo]{liu2021swin}
Ze~Liu, Yutong Lin, Yue Cao, Han Hu, Yixuan Wei, Zheng Zhang, Stephen Lin, and
  Baining Guo.
\newblock Swin transformer: Hierarchical vision transformer using shifted
  windows.
\newblock In \emph{Proceedings of the IEEE/CVF international conference on
  computer vision}, pages 10012--10022, 2021{\natexlab{b}}.

\bibitem[Long et~al.(2022)Long, Cao, Han, and Yang]{Long2022VLM}
Siqu Long, Feiqi Cao, Soyeon~Caren Han, and Haiqin Yang.
\newblock Vision-and-language pretrained models: A survey.
\newblock In \emph{IJCAI}, 2022.

\bibitem[Lu et~al.(2019)Lu, Batra, Parikh, and Lee]{lu2019vilbert}
Jiasen Lu, Dhruv Batra, Devi Parikh, and Stefan Lee.
\newblock Vilbert: Pretraining task-agnostic visiolinguistic representations
  for vision-and-language tasks.
\newblock \emph{Advances in neural information processing systems}, 32, 2019.

\bibitem[Lu et~al.(2021)Lu, Qiu, Chen, Xia, Zhao, Zhang, Yu, Liang, and
  Zhu]{lu2021iconqa}
Pan Lu, Liang Qiu, Jiaqi Chen, Tony Xia, Yizhou Zhao, Wei Zhang, Zhou Yu,
  Xiaodan Liang, and Song-Chun Zhu.
\newblock Iconqa: A new benchmark for abstract diagram understanding and visual
  language reasoning.
\newblock \emph{arXiv preprint arXiv:2110.13214}, 2021.

\bibitem[Lu et~al.(2022)Lu, Mishra, Xia, Qiu, Chang, Zhu, Tafjord, Clark, and
  Kalyan]{lu2022learn}
Pan Lu, Swaroop Mishra, Tony Xia, Liang Qiu, Kai-Wei Chang, Song-Chun Zhu,
  Oyvind Tafjord, Peter Clark, and Ashwin Kalyan.
\newblock Learn to explain: Multimodal reasoning via thought chains for science
  question answering.
\newblock In \emph{The 36th Conference on Neural Information Processing Systems
  (NeurIPS)}, 2022.

\bibitem[Lu et~al.(2023)Lu, Peng, Cheng, Galley, Chang, Wu, Zhu, and
  Gao]{lu2023chameleon}
Pan Lu, Baolin Peng, Hao Cheng, Michel Galley, Kai-Wei Chang, Ying~Nian Wu,
  Song-Chun Zhu, and Jianfeng Gao.
\newblock Chameleon: Plug-and-play compositional reasoning with large language
  models.
\newblock \emph{arXiv preprint arXiv:2304.09842}, 2023.

\bibitem[L{\"u}ddecke and Ecker(2022)]{luddecke2022image}
Timo L{\"u}ddecke and Alexander Ecker.
\newblock Image segmentation using text and image prompts.
\newblock In \emph{Proceedings of the IEEE/CVF Conference on Computer Vision
  and Pattern Recognition}, pages 7086--7096, 2022.

\bibitem[Luo et~al.(2023)Luo, Zhao, Yang, Dong, Qiu, Lu, Wang, and
  Wei]{luo2023valley}
Ruipu Luo, Ziwang Zhao, Min Yang, Junwei Dong, Minghui Qiu, Pengcheng Lu, Tao
  Wang, and Zhongyu Wei.
\newblock Valley: Video assistant with large language model enhanced ability.
\newblock \emph{arXiv preprint arXiv: 2306.07207}, 2023.

\bibitem[Lynch et~al.(2022)Lynch, Wahid, Tompson, Ding, Betker, Baruch,
  Armstrong, and Florence]{lynch2022interactive}
Corey Lynch, Ayzaan Wahid, Jonathan Tompson, Tianli Ding, James Betker, Robert
  Baruch, Travis Armstrong, and Pete Florence.
\newblock Interactive language: Talking to robots in real time.
\newblock \emph{arXiv preprint arXiv:2210.06407}, 2022.

\bibitem[Lyu et~al.(2023)Lyu, Liu, Wu, Du, Huang, Tu, Shi, and Wang]{Macaw-LLM}
Chenyang Lyu, Bingshuai Liu, Minghao Wu, Zefeng Du, Xinting Huang, Zhaopeng Tu,
  Shuming Shi, and Longyue Wang.
\newblock Macaw-llm: Multi-modal language modeling with image, video, audio,
  and text integration.
\newblock \url{https://github.com/lyuchenyang/Macaw-LLM}, 2023.

\bibitem[Ma and Wang(2023)]{ma2023segment}
Jun Ma and Bo~Wang.
\newblock Segment anything in medical images.
\newblock \emph{arXiv preprint arXiv: 2304.12306}, 2023.

\bibitem[Maaz et~al.(2023)Maaz, Rasheed, Khan, and Khan]{maaz2023videochatgpt}
Muhammad Maaz, Hanoona Rasheed, Salman Khan, and Fahad~Shahbaz Khan.
\newblock Video-chatgpt: Towards detailed video understanding via large vision
  and language models.
\newblock \emph{arXiv preprint arXiv: 2306.05424}, 2023.

\bibitem[Mahajan et~al.(2018)Mahajan, Girshick, Ramanathan, He, Paluri, Li,
  Bharambe, and Van Der~Maaten]{mahajan2018exploring}
Dhruv Mahajan, Ross Girshick, Vignesh Ramanathan, Kaiming He, Manohar Paluri,
  Yixuan Li, Ashwin Bharambe, and Laurens Van Der~Maaten.
\newblock Exploring the limits of weakly supervised pretraining.
\newblock In \emph{Proceedings of the European conference on computer vision
  (ECCV)}, pages 181--196, 2018.

\bibitem[Mai et~al.(2023)Mai, Chen, Li, Qian, Elhoseiny, and
  Ghanem]{mai2023llm}
Jinjie Mai, Jun Chen, Bing Li, Guocheng Qian, Mohamed Elhoseiny, and Bernard
  Ghanem.
\newblock Llm as a robotic brain: Unifying egocentric memory and control.
\newblock \emph{arXiv preprint arXiv:2304.09349}, 2023.

\bibitem[Mao et~al.(2016)Mao, Huang, Toshev, Camburu, Yuille, and
  Murphy]{mao2016generation}
Junhua Mao, Jonathan Huang, Alexander Toshev, Oana Camburu, Alan~L Yuille, and
  Kevin Murphy.
\newblock Generation and comprehension of unambiguous object descriptions.
\newblock In \emph{Proceedings of the IEEE conference on computer vision and
  pattern recognition}, pages 11--20, 2016.

\bibitem[Marino et~al.(2019)Marino, Rastegari, Farhadi, and
  Mottaghi]{marino2019ok}
Kenneth Marino, Mohammad Rastegari, Ali Farhadi, and Roozbeh Mottaghi.
\newblock Ok-vqa: A visual question answering benchmark requiring external
  knowledge.
\newblock In \emph{Proceedings of the IEEE/cvf conference on computer vision
  and pattern recognition}, pages 3195--3204, 2019.

\bibitem[Maus et~al.(2023)Maus, Chao, Wong, and Gardner]{maus2023adversarial}
Natalie Maus, Patrick Chao, Eric Wong, and Jacob Gardner.
\newblock Adversarial prompting for black box foundation models.
\newblock \emph{arXiv preprint arXiv:2302.04237}, 2023.

\bibitem[Minderer et~al.(2022)Minderer, Gritsenko, Stone, Neumann, Weissenborn,
  Dosovitskiy, Mahendran, Arnab, Dehghani, Shen, Wang, Zhai, Kipf, and
  Houlsby]{minderer2022simple}
Matthias Minderer, Alexey Gritsenko, Austin Stone, Maxim Neumann, Dirk
  Weissenborn, Alexey Dosovitskiy, Aravindh Mahendran, Anurag Arnab, Mostafa
  Dehghani, Zhuoran Shen, Xiao Wang, Xiaohua Zhai, Thomas Kipf, and Neil
  Houlsby.
\newblock Simple open-vocabulary object detection with vision transformers.
\newblock \emph{arXiv preprint arXiv: 2205.06230}, 2022.

\bibitem[Mishra et~al.(2019)Mishra, Shekhar, Singh, and
  Chakraborty]{mishra2019ocr}
Anand Mishra, Shashank Shekhar, Ajeet~Kumar Singh, and Anirban Chakraborty.
\newblock Ocr-vqa: Visual question answering by reading text in images.
\newblock In \emph{2019 international conference on document analysis and
  recognition (ICDAR)}, pages 947--952. IEEE, 2019.

\bibitem[Moor et~al.(2023)Moor, Banerjee, Abad, Krumholz, Leskovec, Topol, and
  Rajpurkar]{moor2023foundation}
Michael Moor, Oishi Banerjee, Zahra Shakeri~Hossein Abad, Harlan~M Krumholz,
  Jure Leskovec, Eric~J Topol, and Pranav Rajpurkar.
\newblock Foundation models for generalist medical artificial intelligence.
\newblock \emph{Nature}, 616\penalty0 (7956):\penalty0 259--265, 2023.

\bibitem[Mottaghi et~al.(2014)Mottaghi, Chen, Liu, Cho, Lee, Fidler, Urtasun,
  and Yuille]{mottaghi2014role}
Roozbeh Mottaghi, Xianjie Chen, Xiaobai Liu, Nam-Gyu Cho, Seong-Whan Lee, Sanja
  Fidler, Raquel Urtasun, and Alan Yuille.
\newblock The role of context for object detection and semantic segmentation in
  the wild.
\newblock In \emph{Proceedings of the IEEE conference on computer vision and
  pattern recognition}, pages 891--898, 2014.

\bibitem[Mu et~al.(2022)Mu, Kirillov, Wagner, and Xie]{mu2022slip}
Norman Mu, Alexander Kirillov, David Wagner, and Saining Xie.
\newblock Slip: Self-supervision meets language-image pre-training.
\newblock In \emph{Computer Vision--ECCV 2022: 17th European Conference, Tel
  Aviv, Israel, October 23--27, 2022, Proceedings, Part XXVI}, pages 529--544.
  Springer, 2022.

\bibitem[Mu et~al.(2023)Mu, Zhang, Hu, Wang, Ding, Jin, Wang, Dai, Qiao, and
  Luo]{mu2023embodiedgpt}
Yao Mu, Qinglong Zhang, Mengkang Hu, Wenhai Wang, Mingyu Ding, Jun Jin, Bin
  Wang, Jifeng Dai, Yu~Qiao, and Ping Luo.
\newblock Embodiedgpt: Vision-language pre-training via embodied chain of
  thought.
\newblock \emph{arXiv preprint arXiv:2305.15021}, 2023.

\bibitem[M{\"u}ndler et~al.(2023)M{\"u}ndler, He, Jenko, and
  Vechev]{mundler2023self}
Niels M{\"u}ndler, Jingxuan He, Slobodan Jenko, and Martin Vechev.
\newblock Self-contradictory hallucinations of large language models:
  Evaluation, detection and mitigation.
\newblock \emph{arXiv preprint arXiv:2305.15852}, 2023.

\bibitem[Nagaraja et~al.(2016)Nagaraja, Morariu, and
  Davis]{nagaraja2016modeling}
Varun~K Nagaraja, Vlad~I Morariu, and Larry~S Davis.
\newblock Modeling context between objects for referring expression
  understanding.
\newblock In \emph{Computer Vision--ECCV 2016: 14th European Conference,
  Amsterdam, The Netherlands, October 11--14, 2016, Proceedings, Part IV 14},
  pages 792--807. Springer, 2016.

\bibitem[OpenAI(2023)]{openai2023gpt4}
OpenAI.
\newblock Gpt-4 technical report.
\newblock \emph{PREPRINT}, 2023.

\bibitem[Ordonez et~al.(2011)Ordonez, Kulkarni, and Berg]{ordonez2011im2text}
Vicente Ordonez, Girish Kulkarni, and Tamara Berg.
\newblock Im2text: Describing images using 1 million captioned photographs.
\newblock \emph{Advances in neural information processing systems}, 24, 2011.

\bibitem[Pan et~al.(2023)Pan, Lin, Ge, Zhu, Zhang, Wang, Qiao, and
  Li]{pan2023retrieving}
Junting Pan, Ziyi Lin, Yuying Ge, Xiatian Zhu, Renrui Zhang, Yi~Wang, Yu~Qiao,
  and Hongsheng Li.
\newblock Retrieving-to-answer: Zero-shot video question answering with frozen
  large language models.
\newblock \emph{arXiv preprint arXiv:2306.11732}, 2023.

\bibitem[Parisi et~al.(2019)Parisi, Kemker, Part, Kanan, and
  Wermter]{parisi2019continual}
German~I Parisi, Ronald Kemker, Jose~L Part, Christopher Kanan, and Stefan
  Wermter.
\newblock Continual lifelong learning with neural networks: A review.
\newblock \emph{Neural networks}, 113:\penalty0 54--71, 2019.

\bibitem[Park et~al.(2023)Park, Lim, Lee, Park, Yu, and Choi]{park2023clara}
Jeongeun Park, Seungwon Lim, Joonhyung Lee, Sangbeom Park, Youngjae Yu, and
  Sungjoon Choi.
\newblock Clara: Classifying and disambiguating user commands for reliable
  interactive robotic agents.
\newblock \emph{arXiv preprint arXiv:2306.10376}, 2023.

\bibitem[Peng et~al.(2022)Peng, Dong, Bao, Ye, and Wei]{peng2022beit}
Zhiliang Peng, Li~Dong, Hangbo Bao, Qixiang Ye, and Furu Wei.
\newblock Beit v2: Masked image modeling with vector-quantized visual
  tokenizers.
\newblock \emph{arXiv preprint arXiv:2208.06366}, 2022.

\bibitem[Peng et~al.(2023)Peng, Wang, Dong, Hao, Huang, Ma, and
  Wei]{peng2023kosmos2}
Zhiliang Peng, Wenhui Wang, Li~Dong, Yaru Hao, Shaohan Huang, Shuming Ma, and
  Furu Wei.
\newblock Kosmos-2: Grounding multimodal large language models to the world.
\newblock \emph{arXiv preprint arXiv: 2306.14824}, 2023.

\bibitem[Perez and Ribeiro(2022)]{perez2022ignore}
F{\'a}bio Perez and Ian Ribeiro.
\newblock Ignore previous prompt: Attack techniques for language models.
\newblock \emph{arXiv preprint arXiv:2211.09527}, 2022.

\bibitem[Pont-Tuset et~al.(2020)Pont-Tuset, Uijlings, Changpinyo, Soricut, and
  Ferrari]{pont2020connecting}
Jordi Pont-Tuset, Jasper Uijlings, Soravit Changpinyo, Radu Soricut, and
  Vittorio Ferrari.
\newblock Connecting vision and language with localized narratives.
\newblock In \emph{Computer Vision--ECCV 2020: 16th European Conference,
  Glasgow, UK, August 23--28, 2020, Proceedings, Part V 16}, pages 647--664.
  Springer, 2020.

\bibitem[Qiu et~al.(2023)Qiu, Hu, Li, and Liu]{qiu2023learnable}
Zhongxi Qiu, Yan Hu, Heng Li, and Jiang Liu.
\newblock Learnable ophthalmology sam.
\newblock \emph{arXiv preprint arXiv:2304.13425}, 2023.

\bibitem[Radford et~al.(2019)Radford, Wu, Child, Luan, Amodei, Sutskever,
  et~al.]{radford2019language}
Alec Radford, Jeffrey Wu, Rewon Child, David Luan, Dario Amodei, Ilya
  Sutskever, et~al.
\newblock Language models are unsupervised multitask learners.
\newblock \emph{OpenAI blog}, 1\penalty0 (8):\penalty0 9, 2019.

\bibitem[Radford et~al.(2021)Radford, Kim, Hallacy, Ramesh, Goh, Agarwal,
  Sastry, Askell, Mishkin, Clark, et~al.]{radford2021learning}
Alec Radford, Jong~Wook Kim, Chris Hallacy, Aditya Ramesh, Gabriel Goh,
  Sandhini Agarwal, Girish Sastry, Amanda Askell, Pamela Mishkin, Jack Clark,
  et~al.
\newblock Learning transferable visual models from natural language
  supervision.
\newblock In \emph{International conference on machine learning}, pages
  8748--8763. PMLR, 2021.

\bibitem[Raffel et~al.(2020)Raffel, Shazeer, Roberts, Lee, Narang, Matena,
  Zhou, Li, and Liu]{raffel2020exploring}
Colin Raffel, Noam Shazeer, Adam Roberts, Katherine Lee, Sharan Narang, Michael
  Matena, Yanqi Zhou, Wei Li, and Peter~J Liu.
\newblock Exploring the limits of transfer learning with a unified text-to-text
  transformer.
\newblock \emph{The Journal of Machine Learning Research}, 21\penalty0
  (1):\penalty0 5485--5551, 2020.

\bibitem[Rajič et~al.(2023)Rajič, Ke, Tai, Tang, Danelljan, and Yu]{sam-pt}
Frano Rajič, Lei Ke, Yu-Wing Tai, Chi-Keung Tang, Martin Danelljan, and Fisher
  Yu.
\newblock Segment anything meets point tracking.
\newblock \emph{arXiv:2307.01197}, 2023.

\bibitem[Ranjit et~al.(2023)Ranjit, Ganapathy, Manuel, and
  Ganu]{ranjit2023retrieval}
Mercy Ranjit, Gopinath Ganapathy, Ranjit Manuel, and Tanuja Ganu.
\newblock Retrieval augmented chest x-ray report generation using openai gpt
  models.
\newblock \emph{arXiv preprint arXiv:2305.03660}, 2023.

\bibitem[Raven(2003)]{raven2003raven}
Jean Raven.
\newblock Raven progressive matrices.
\newblock In \emph{Handbook of nonverbal assessment}, pages 223--237. Springer,
  2003.

\bibitem[Recht et~al.(2019)Recht, Roelofs, Schmidt, and
  Shankar]{recht2019imagenet}
Benjamin Recht, Rebecca Roelofs, Ludwig Schmidt, and Vaishaal Shankar.
\newblock Do imagenet classifiers generalize to imagenet?
\newblock In \emph{International conference on machine learning}, pages
  5389--5400. PMLR, 2019.

\bibitem[Ren et~al.(2023)Ren, Dixit, Bodrova, Singh, Tu, Brown, Xu, Takayama,
  Xia, Varley, et~al.]{ren2023robots}
Allen~Z Ren, Anushri Dixit, Alexandra Bodrova, Sumeet Singh, Stephen Tu, Noah
  Brown, Peng Xu, Leila Takayama, Fei Xia, Jake Varley, et~al.
\newblock Robots that ask for help: Uncertainty alignment for large language
  model planners.
\newblock \emph{arXiv preprint arXiv:2307.01928}, 2023.

\bibitem[Ren et~al.(2015)Ren, He, Girshick, and Sun]{ren2015faster}
Shaoqing Ren, Kaiming He, Ross Girshick, and Jian Sun.
\newblock Faster r-cnn: Towards real-time object detection with region proposal
  networks.
\newblock \emph{Advances in neural information processing systems}, 28, 2015.

\bibitem[Rezatofighi et~al.(2019)Rezatofighi, Tsoi, Gwak, Sadeghian, Reid, and
  Savarese]{rezatofighi2019generalized}
Hamid Rezatofighi, Nathan Tsoi, JunYoung Gwak, Amir Sadeghian, Ian Reid, and
  Silvio Savarese.
\newblock Generalized intersection over union: A metric and a loss for bounding
  box regression.
\newblock In \emph{Proceedings of the IEEE/CVF conference on computer vision
  and pattern recognition}, pages 658--666, 2019.

\bibitem[Ronneberger et~al.(2015)Ronneberger, Fischer, and
  Brox]{ronneberger2015u}
Olaf Ronneberger, Philipp Fischer, and Thomas Brox.
\newblock U-net: Convolutional networks for biomedical image segmentation.
\newblock In \emph{Medical Image Computing and Computer-Assisted
  Intervention--MICCAI 2015: 18th International Conference, Munich, Germany,
  October 5-9, 2015, Proceedings, Part III 18}, pages 234--241. Springer, 2015.

\bibitem[Schaeffer et~al.(2023)Schaeffer, Miranda, and
  Koyejo]{schaeffer2023emergent}
Rylan Schaeffer, Brando Miranda, and Sanmi Koyejo.
\newblock Are emergent abilities of large language models a mirage?
\newblock \emph{arXiv preprint arXiv:2304.15004}, 2023.

\bibitem[Schuhmann et~al.(2021)Schuhmann, Vencu, Beaumont, Kaczmarczyk, Mullis,
  Katta, Coombes, Jitsev, and Komatsuzaki]{schuhmann2021laion}
Christoph Schuhmann, Richard Vencu, Romain Beaumont, Robert Kaczmarczyk,
  Clayton Mullis, Aarush Katta, Theo Coombes, Jenia Jitsev, and Aran
  Komatsuzaki.
\newblock Laion-400m: Open dataset of clip-filtered 400 million image-text
  pairs.
\newblock \emph{arXiv preprint arXiv:2111.02114}, 2021.

\bibitem[Schuhmann et~al.(2022)Schuhmann, Beaumont, Vencu, Gordon, Wightman,
  Cherti, Coombes, Katta, Mullis, Wortsman, Schramowski, Kundurthy, Crowson,
  Schmidt, Kaczmarczyk, and Jitsev]{schuhmann2022laion}
Christoph Schuhmann, Romain Beaumont, Richard Vencu, Cade Gordon, Ross
  Wightman, Mehdi Cherti, Theo Coombes, Aarush Katta, Clayton Mullis, Mitchell
  Wortsman, Patrick Schramowski, Srivatsa Kundurthy, Katherine Crowson, Ludwig
  Schmidt, Robert Kaczmarczyk, and Jenia Jitsev.
\newblock Laion-5b: An open large-scale dataset for training next generation
  image-text models, 2022.

\bibitem[Schwenk et~al.(2022)Schwenk, Khandelwal, Clark, Marino, and
  Mottaghi]{schwenk2022okvqa}
Dustin Schwenk, Apoorv Khandelwal, Christopher Clark, Kenneth Marino, and
  Roozbeh Mottaghi.
\newblock A-okvqa: A benchmark for visual question answering using world
  knowledge.
\newblock In \emph{European Conference on Computer Vision}, pages 146--162.
  Springer, 2022.

\bibitem[Seo et~al.(2020)Seo, Lee, and Han]{seo2020urvos}
Seonguk Seo, Joon-Young Lee, and Bohyung Han.
\newblock Urvos: Unified referring video object segmentation network with a
  large-scale benchmark.
\newblock In \emph{Computer Vision--ECCV 2020: 16th European Conference,
  Glasgow, UK, August 23--28, 2020, Proceedings, Part XV 16}, pages 208--223.
  Springer, 2020.

\bibitem[Shah et~al.(2021{\natexlab{a}})Shah, Eysenbach, Kahn, Rhinehart, and
  Levine]{shah2021ving}
Dhruv Shah, Benjamin Eysenbach, Gregory Kahn, Nicholas Rhinehart, and Sergey
  Levine.
\newblock Ving: Learning open-world navigation with visual goals.
\newblock In \emph{2021 IEEE International Conference on Robotics and
  Automation (ICRA)}, pages 13215--13222. IEEE, 2021{\natexlab{a}}.

\bibitem[Shah et~al.(2022)Shah, Osinski, Ichter, and Levine]{shah2022lmnav}
Dhruv Shah, Blazej Osinski, Brian Ichter, and Sergey Levine.
\newblock Lm-nav: Robotic navigation with large pre-trained models of language,
  vision, and action.
\newblock \emph{arXiv preprint arXiv: 2207.04429}, 2022.

\bibitem[Shah et~al.(2021{\natexlab{b}})Shah, Wild, Wang, Alex, Houghton, Guss,
  Mohanty, Kanervisto, Milani, Topin, et~al.]{shah2021minerl}
Rohin Shah, Cody Wild, Steven~H Wang, Neel Alex, Brandon Houghton, William
  Guss, Sharada Mohanty, Anssi Kanervisto, Stephanie Milani, Nicholay Topin,
  et~al.
\newblock The minerl basalt competition on learning from human feedback.
\newblock \emph{arXiv preprint arXiv:2107.01969}, 2021{\natexlab{b}}.

\bibitem[Shaharabany et~al.(2023)Shaharabany, Dahan, Giryes, and
  Wolf]{shaharabany2023autosam}
Tal Shaharabany, Aviad Dahan, Raja Giryes, and Lior Wolf.
\newblock Autosam: Adapting sam to medical images by overloading the prompt
  encoder.
\newblock \emph{arXiv preprint arXiv:2306.06370}, 2023.

\bibitem[Shao et~al.(2019)Shao, Li, Zhang, Peng, Yu, Zhang, Li, and
  Sun]{shao2019objects365}
Shuai Shao, Zeming Li, Tianyuan Zhang, Chao Peng, Gang Yu, Xiangyu Zhang, Jing
  Li, and Jian Sun.
\newblock Objects365: A large-scale, high-quality dataset for object detection.
\newblock In \emph{Proceedings of the IEEE/CVF international conference on
  computer vision}, pages 8430--8439, 2019.

\bibitem[Sharma et~al.(2018)Sharma, Ding, Goodman, and
  Soricut]{sharma2018conceptual}
Piyush Sharma, Nan Ding, Sebastian Goodman, and Radu Soricut.
\newblock Conceptual captions: A cleaned, hypernymed, image alt-text dataset
  for automatic image captioning.
\newblock In \emph{Proceedings of ACL}, 2018.

\bibitem[Shen et~al.(2023)Shen, Song, Tan, Li, Lu, and
  Zhuang]{shen2023hugginggpt}
Yongliang Shen, Kaitao Song, Xu~Tan, Dongsheng Li, Weiming Lu, and Yueting
  Zhuang.
\newblock Hugginggpt: Solving ai tasks with chatgpt and its friends in hugging
  face.
\newblock \emph{arXiv preprint arXiv: 2303.17580}, 2023.

\bibitem[Shtedritski et~al.(2023)Shtedritski, Rupprecht, and
  Vedaldi]{shtedritski2023does}
Aleksandar Shtedritski, Christian Rupprecht, and Andrea Vedaldi.
\newblock What does clip know about a red circle? visual prompt engineering for
  vlms.
\newblock \emph{arXiv preprint arXiv:2304.06712}, 2023.

\bibitem[Shukor et~al.(2023)Shukor, Dancette, and Cord]{shukor2023ep}
Mustafa Shukor, Corentin Dancette, and Matthieu Cord.
\newblock ep-alm: Efficient perceptual augmentation of language models.
\newblock \emph{arXiv preprint arXiv:2303.11403}, 2023.

\bibitem[Sidorov et~al.(2020)Sidorov, Hu, Rohrbach, and
  Singh]{sidorov2020textcaps}
Oleksii Sidorov, Ronghang Hu, Marcus Rohrbach, and Amanpreet Singh.
\newblock Textcaps: a dataset for image captioning with reading comprehension.
\newblock In \emph{Computer Vision--ECCV 2020: 16th European Conference,
  Glasgow, UK, August 23--28, 2020, Proceedings, Part II 16}, pages 742--758.
  Springer, 2020.

\bibitem[Silberman et~al.(2012)Silberman, Hoiem, Kohli, and
  Fergus]{silberman2012indoor}
Nathan Silberman, Derek Hoiem, Pushmeet Kohli, and Rob Fergus.
\newblock Indoor segmentation and support inference from rgbd images.
\newblock In \emph{Computer Vision--ECCV 2012: 12th European Conference on
  Computer Vision, Florence, Italy, October 7-13, 2012, Proceedings, Part V
  12}, pages 746--760. Springer, 2012.

\bibitem[Silva et~al.(2014)Silva, Histace, Romain, Dray, and
  Granado]{silva2014toward}
Juan Silva, Aymeric Histace, Olivier Romain, Xavier Dray, and Bertrand Granado.
\newblock Toward embedded detection of polyps in wce images for early diagnosis
  of colorectal cancer.
\newblock \emph{International journal of computer assisted radiology and
  surgery}, 9:\penalty0 283--293, 2014.

\bibitem[Singh et~al.(2019)Singh, Natarajan, Shah, Jiang, Chen, Batra, Parikh,
  and Rohrbach]{singh2019towards}
Amanpreet Singh, Vivek Natarajan, Meet Shah, Yu~Jiang, Xinlei Chen, Dhruv
  Batra, Devi Parikh, and Marcus Rohrbach.
\newblock Towards vqa models that can read.
\newblock In \emph{Proceedings of the IEEE/CVF conference on computer vision
  and pattern recognition}, pages 8317--8326, 2019.

\bibitem[Singh et~al.(2022)Singh, Hu, Goswami, Couairon, Galuba, Rohrbach, and
  Kiela]{singh2022flava}
Amanpreet Singh, Ronghang Hu, Vedanuj Goswami, Guillaume Couairon, Wojciech
  Galuba, Marcus Rohrbach, and Douwe Kiela.
\newblock Flava: A foundational language and vision alignment model.
\newblock In \emph{Proceedings of the IEEE/CVF Conference on Computer Vision
  and Pattern Recognition}, pages 15638--15650, 2022.

\bibitem[Sirinukunwattana et~al.(2017)Sirinukunwattana, Pluim, Chen, Qi, Heng,
  Guo, Wang, Matuszewski, Bruni, Sanchez, et~al.]{sirinukunwattana2017gland}
Korsuk Sirinukunwattana, Josien~PW Pluim, Hao Chen, Xiaojuan Qi, Pheng-Ann
  Heng, Yun~Bo Guo, Li~Yang Wang, Bogdan~J Matuszewski, Elia Bruni, Urko
  Sanchez, et~al.
\newblock Gland segmentation in colon histology images: The glas challenge
  contest.
\newblock \emph{Medical image analysis}, 35:\penalty0 489--502, 2017.

\bibitem[Skreta et~al.(2023)Skreta, Yoshikawa, Arellano-Rubach, Ji, Kristensen,
  Darvish, Aspuru-Guzik, Shkurti, and Garg]{skreta2023errors}
Marta Skreta, Naruki Yoshikawa, Sebastian Arellano-Rubach, Zhi Ji, Lasse~Bjørn
  Kristensen, Kourosh Darvish, Alán Aspuru-Guzik, Florian Shkurti, and Animesh
  Garg.
\newblock Errors are useful prompts: Instruction guided task programming with
  verifier-assisted iterative prompting, 2023.

\bibitem[Song et~al.(2015)Song, Lichtenberg, and Xiao]{song2015sun}
Shuran Song, Samuel~P Lichtenberg, and Jianxiong Xiao.
\newblock Sun rgb-d: A rgb-d scene understanding benchmark suite.
\newblock In \emph{Proceedings of the IEEE conference on computer vision and
  pattern recognition}, pages 567--576, 2015.

\bibitem[Srinivasan et~al.(2021)Srinivasan, Raman, Chen, Bendersky, and
  Najork]{srinivasan2021wit}
Krishna Srinivasan, Karthik Raman, Jiecao Chen, Michael Bendersky, and Marc
  Najork.
\newblock Wit: Wikipedia-based image text dataset for multimodal multilingual
  machine learning.
\newblock \emph{arXiv preprint arXiv: 2103.01913}, 2021.

\bibitem[Stella et~al.(2023)Stella, Della~Santina, and Hughes]{stella2023can}
Francesco Stella, Cosimo Della~Santina, and Josie Hughes.
\newblock Can large language models design a robot?
\newblock \emph{arXiv preprint arXiv:2303.15324}, 2023.

\bibitem[Sun et~al.(2023{\natexlab{a}})Sun, Fang, Wu, Wang, and
  Cao]{sun2023eva}
Quan Sun, Yuxin Fang, Ledell Wu, Xinlong Wang, and Yue Cao.
\newblock Eva-clip: Improved training techniques for clip at scale.
\newblock \emph{arXiv preprint arXiv:2303.15389}, 2023{\natexlab{a}}.

\bibitem[Sun et~al.(2022)Sun, Dong, Patra, Ma, Huang, Benhaim, Chaudhary, Song,
  and Wei]{sun2022length}
Yutao Sun, Li~Dong, Barun Patra, Shuming Ma, Shaohan Huang, Alon Benhaim,
  Vishrav Chaudhary, Xia Song, and Furu Wei.
\newblock A length-extrapolatable transformer.
\newblock \emph{arXiv preprint arXiv:2212.10554}, 2022.

\bibitem[Sun et~al.(2023{\natexlab{b}})Sun, Dong, Huang, Ma, Xia, Xue, Wang,
  and Wei]{sun2023retentive}
Yutao Sun, Li~Dong, Shaohan Huang, Shuming Ma, Yuqing Xia, Jilong Xue, Jianyong
  Wang, and Furu Wei.
\newblock Retentive network: A successor to transformer for large language
  models.
\newblock \emph{arXiv preprint arXiv:2307.08621}, 2023{\natexlab{b}}.

\bibitem[Surís et~al.(2023)Surís, Menon, and Vondrick]{surís2023vipergpt}
Dídac Surís, Sachit Menon, and Carl Vondrick.
\newblock Vipergpt: Visual inference via python execution for reasoning.
\newblock \emph{arXiv preprint arXiv: 2303.08128}, 2023.

\bibitem[Tajbakhsh et~al.(2015)Tajbakhsh, Gurudu, and
  Liang]{tajbakhsh2015automated}
Nima Tajbakhsh, Suryakanth~R Gurudu, and Jianming Liang.
\newblock Automated polyp detection in colonoscopy videos using shape and
  context information.
\newblock \emph{IEEE transactions on medical imaging}, 35\penalty0
  (2):\penalty0 630--644, 2015.

\bibitem[Tan and Bansal(2019)]{tan2019lxmert}
Hao Tan and Mohit Bansal.
\newblock Lxmert: Learning cross-modality encoder representations from
  transformers.
\newblock \emph{arXiv preprint arXiv:1908.07490}, 2019.

\bibitem[Tan and Le(2019)]{tan2019efficientnet}
Mingxing Tan and Quoc Le.
\newblock Efficientnet: Rethinking model scaling for convolutional neural
  networks.
\newblock In \emph{International conference on machine learning}, pages
  6105--6114. PMLR, 2019.

\bibitem[Tarvainen and Valpola(2017)]{tarvainen2017mean}
Antti Tarvainen and Harri Valpola.
\newblock Mean teachers are better role models: Weight-averaged consistency
  targets improve semi-supervised deep learning results.
\newblock \emph{Advances in neural information processing systems}, 30, 2017.

\bibitem[Thawkar et~al.(2023)Thawkar, Shaker, Mullappilly, Cholakkal, Anwer,
  Khan, Laaksonen, and Khan]{thawkar2023xraygpt}
Omkar Thawkar, Abdelrahman Shaker, Sahal~Shaji Mullappilly, Hisham Cholakkal,
  Rao~Muhammad Anwer, Salman Khan, Jorma Laaksonen, and Fahad~Shahbaz Khan.
\newblock Xraygpt: Chest radiographs summarization using medical
  vision-language models.
\newblock \emph{arXiv preprint arXiv:2306.07971}, 2023.

\bibitem[Thomee et~al.(2016)Thomee, Shamma, Friedland, Elizalde, Ni, Poland,
  Borth, and Li]{thomee2016yfcc100m}
Bart Thomee, David~A Shamma, Gerald Friedland, Benjamin Elizalde, Karl Ni,
  Douglas Poland, Damian Borth, and Li-Jia Li.
\newblock Yfcc100m: The new data in multimedia research.
\newblock \emph{Communications of the ACM}, 59\penalty0 (2):\penalty0 64--73,
  2016.

\bibitem[Tong et~al.(2023)Tong, Jones, and Steinhardt]{tong2023mass}
Shengbang Tong, Erik Jones, and Jacob Steinhardt.
\newblock Mass-producing failures of multimodal systems with language models.
\newblock \emph{arXiv preprint arXiv:2306.12105}, 2023.

\bibitem[Touvron et~al.(2021)Touvron, Cord, Douze, Massa, Sablayrolles, and
  J{\'e}gou]{touvron2021training}
Hugo Touvron, Matthieu Cord, Matthijs Douze, Francisco Massa, Alexandre
  Sablayrolles, and Herv{\'e} J{\'e}gou.
\newblock Training data-efficient image transformers \& distillation through
  attention.
\newblock In \emph{International conference on machine learning}, pages
  10347--10357. PMLR, 2021.

\bibitem[Touvron et~al.(2023)Touvron, Lavril, Izacard, Martinet, Lachaux,
  Lacroix, Rozi{\`e}re, Goyal, Hambro, Azhar, et~al.]{touvron2023llama}
Hugo Touvron, Thibaut Lavril, Gautier Izacard, Xavier Martinet, Marie-Anne
  Lachaux, Timoth{\'e}e Lacroix, Baptiste Rozi{\`e}re, Naman Goyal, Eric
  Hambro, Faisal Azhar, et~al.
\newblock Llama: Open and efficient foundation language models.
\newblock \emph{arXiv preprint arXiv:2302.13971}, 2023.

\bibitem[Tschannen et~al.(2023)Tschannen, Kumar, Steiner, Zhai, Houlsby, and
  Beyer]{tschannen2023image}
Michael Tschannen, Manoj Kumar, Andreas Steiner, Xiaohua Zhai, Neil Houlsby,
  and Lucas Beyer.
\newblock Image captioners are scalable vision learners too.
\newblock \emph{arXiv preprint arXiv:2306.07915}, 2023.

\bibitem[Tsimpoukelli et~al.(2021)Tsimpoukelli, Menick, Cabi, Eslami, Vinyals,
  and Hill]{tsimpoukelli2021multimodal}
Maria Tsimpoukelli, Jacob~L Menick, Serkan Cabi, SM~Eslami, Oriol Vinyals, and
  Felix Hill.
\newblock Multimodal few-shot learning with frozen language models.
\newblock \emph{Advances in Neural Information Processing Systems},
  34:\penalty0 200--212, 2021.

\bibitem[Tu et~al.(2005)Tu, Chen, Yuille, and Zhu]{tu2005image}
Zhuowen Tu, Xiangrong Chen, Alan~L Yuille, and Song-Chun Zhu.
\newblock Image parsing: Unifying segmentation, detection, and recognition.
\newblock \emph{International Journal of computer vision}, 63:\penalty0
  113--140, 2005.

\bibitem[Vaswani et~al.(2017)Vaswani, Shazeer, Parmar, Uszkoreit, Jones, Gomez,
  Kaiser, and Polosukhin]{vaswani2017attention}
Ashish Vaswani, Noam Shazeer, Niki Parmar, Jakob Uszkoreit, Llion Jones,
  Aidan~N Gomez, {\L}ukasz Kaiser, and Illia Polosukhin.
\newblock Attention is all you need.
\newblock \emph{Advances in neural information processing systems}, 30, 2017.

\bibitem[Wake et~al.(2023)Wake, Kanehira, Sasabuchi, Takamatsu, and
  Ikeuchi]{wake2023chatgpt}
Naoki Wake, Atsushi Kanehira, Kazuhiro Sasabuchi, Jun Takamatsu, and Katsushi
  Ikeuchi.
\newblock Chatgpt empowered long-step robot control in various environments: A
  case application.
\newblock \emph{arXiv preprint arXiv: 2304.03893}, 2023.

\bibitem[Wang et~al.(2023{\natexlab{a}})Wang, Xie, Jiang, Mandlekar, Xiao, Zhu,
  Fan, and Anandkumar]{wang2023voyager}
Guanzhi Wang, Yuqi Xie, Yunfan Jiang, Ajay Mandlekar, Chaowei Xiao, Yuke Zhu,
  Linxi Fan, and Anima Anandkumar.
\newblock Voyager: An open-ended embodied agent with large language models.
\newblock \emph{arXiv preprint arXiv:2305.16291}, 2023{\natexlab{a}}.

\bibitem[Wang et~al.(2022{\natexlab{a}})Wang, Ma, Huang, Dong, Wang, Peng, Wu,
  Bajaj, Singhal, Benhaim, Patra, Liu, Chaudhary, Song, and
  Wei]{wang2022foundation}
Hongyu Wang, Shuming Ma, Shaohan Huang, Li~Dong, Wenhui Wang, Zhiliang Peng,
  Yu~Wu, Payal Bajaj, Saksham Singhal, Alon Benhaim, Barun Patra, Zhun Liu,
  Vishrav Chaudhary, Xia Song, and Furu Wei.
\newblock Foundation transformers, 2022{\natexlab{a}}.

\bibitem[Wang et~al.(2022{\natexlab{b}})Wang, Yang, Hu, Li, Lin, Gan, Liu, Liu,
  and Wang]{wang2022git}
Jianfeng Wang, Zhengyuan Yang, Xiaowei Hu, Linjie Li, Kevin Lin, Zhe Gan,
  Zicheng Liu, Ce~Liu, and Lijuan Wang.
\newblock Git: A generative image-to-text transformer for vision and language.
\newblock \emph{arXiv preprint arXiv:2205.14100}, 2022{\natexlab{b}}.

\bibitem[Wang et~al.(2022{\natexlab{c}})Wang, Ge, Cai, Yan, Lin, Shan, Qie, and
  Shou]{wang2022object}
Jinpeng Wang, Yixiao Ge, Guanyu Cai, Rui Yan, Xudong Lin, Ying Shan, Xiaohu
  Qie, and Mike~Zheng Shou.
\newblock Object-aware video-language pre-training for retrieval.
\newblock In \emph{Proceedings of the IEEE/CVF conference on computer vision
  and pattern recognition}, pages 3313--3322, 2022{\natexlab{c}}.

\bibitem[Wang et~al.(2023{\natexlab{b}})Wang, Liu, Park, Chen, and
  Xiao]{wang2023adversarial}
Jiongxiao Wang, Zichen Liu, Keun~Hee Park, Muhao Chen, and Chaowei Xiao.
\newblock Adversarial demonstration attacks on large language models.
\newblock \emph{arXiv preprint arXiv:2305.14950}, 2023{\natexlab{b}}.

\bibitem[Wang et~al.(2023{\natexlab{c}})Wang, Chen, Luo, Dai, Yuan, Wu, and
  Jiang]{wang2023chatvideo}
Junke Wang, Dongdong Chen, Chong Luo, Xiyang Dai, Lu~Yuan, Zuxuan Wu, and
  Yu-Gang Jiang.
\newblock Chatvideo: A tracklet-centric multimodal and versatile video
  understanding system.
\newblock \emph{arXiv preprint arXiv:2304.14407}, 2023{\natexlab{c}}.

\bibitem[Wang et~al.(2023{\natexlab{d}})Wang, Zhang, Su, and
  Zhu]{wang2023comprehensive}
Liyuan Wang, Xingxing Zhang, Hang Su, and Jun Zhu.
\newblock A comprehensive survey of continual learning: Theory, method and
  application.
\newblock \emph{arXiv preprint arXiv:2302.00487}, 2023{\natexlab{d}}.

\bibitem[Wang et~al.(2023{\natexlab{e}})Wang, Zhang, Fei, Zheng, Tang, Li, Gao,
  and Zhao]{wang2023caption}
Teng Wang, Jinrui Zhang, Junjie Fei, Hao Zheng, Yunlong Tang, Zhe Li, Mingqi
  Gao, and Shanshan Zhao.
\newblock Caption anything: Interactive image description with diverse
  multimodal controls.
\newblock \emph{arXiv preprint arXiv: 2305.02677}, 2023{\natexlab{e}}.

\bibitem[Wang et~al.(2021{\natexlab{a}})Wang, Feiszli, Wang, and
  Tran]{wang2021unidentified}
Weiyao Wang, Matt Feiszli, Heng Wang, and Du~Tran.
\newblock Unidentified video objects: A benchmark for dense, open-world
  segmentation.
\newblock In \emph{Proceedings of the IEEE/CVF International Conference on
  Computer Vision}, pages 10776--10785, 2021{\natexlab{a}}.

\bibitem[Wang et~al.(2023{\natexlab{f}})Wang, Chen, Chen, Wu, Zhu, Zeng, Luo,
  Lu, Zhou, Qiao, and Dai]{wang2023visionllm}
Wenhai Wang, Zhe Chen, Xiaokang Chen, Jiannan Wu, Xizhou Zhu, Gang Zeng, Ping
  Luo, Tong Lu, Jie Zhou, Yu~Qiao, and Jifeng Dai.
\newblock Visionllm: Large language model is also an open-ended decoder for
  vision-centric tasks.
\newblock \emph{arXiv preprint arXiv: 2305.11175}, 2023{\natexlab{f}}.

\bibitem[Wang et~al.(2022{\natexlab{d}})Wang, Bao, Dong, Bjorck, Peng, Liu,
  Aggarwal, Mohammed, Singhal, Som, et~al.]{wang2022image}
Wenhui Wang, Hangbo Bao, Li~Dong, Johan Bjorck, Zhiliang Peng, Qiang Liu, Kriti
  Aggarwal, Owais~Khan Mohammed, Saksham Singhal, Subhojit Som, et~al.
\newblock Image as a foreign language: Beit pretraining for all vision and
  vision-language tasks.
\newblock \emph{arXiv preprint arXiv:2208.10442}, 2022{\natexlab{d}}.

\bibitem[Wang et~al.(2019)Wang, Wu, Chen, Li, Wang, and Wang]{wang2019vatex}
Xin Wang, Jiawei Wu, Junkun Chen, Lei Li, Yuan-Fang Wang, and William~Yang
  Wang.
\newblock Vatex: A large-scale, high-quality multilingual dataset for
  video-and-language research.
\newblock In \emph{Proceedings of the IEEE/CVF International Conference on
  Computer Vision}, pages 4581--4591, 2019.

\bibitem[Wang et~al.(2023{\natexlab{g}})Wang, Wang, Cao, Shen, and
  Huang]{wang2023images}
Xinlong Wang, Wen Wang, Yue Cao, Chunhua Shen, and Tiejun Huang.
\newblock Images speak in images: A generalist painter for in-context visual
  learning.
\newblock In \emph{Proceedings of the IEEE/CVF Conference on Computer Vision
  and Pattern Recognition}, pages 6830--6839, 2023{\natexlab{g}}.

\bibitem[Wang et~al.(2023{\natexlab{h}})Wang, Zhang, Cao, Wang, Shen, and
  Huang]{wang2023seggpt}
Xinlong Wang, Xiaosong Zhang, Yue Cao, Wen Wang, Chunhua Shen, and Tiejun
  Huang.
\newblock Seggpt: Segmenting everything in context.
\newblock \emph{arXiv preprint arXiv:2304.03284}, 2023{\natexlab{h}}.

\bibitem[Wang et~al.(2022{\natexlab{e}})Wang, Lu, Li, Tao, Guo, Gong, and
  Liu]{wang2022cris}
Zhaoqing Wang, Yu~Lu, Qiang Li, Xunqiang Tao, Yandong Guo, Mingming Gong, and
  Tongliang Liu.
\newblock Cris: Clip-driven referring image segmentation.
\newblock In \emph{Proceedings of the IEEE/CVF conference on computer vision
  and pattern recognition}, pages 11686--11695, 2022{\natexlab{e}}.

\bibitem[Wang et~al.(2023{\natexlab{i}})Wang, Li, Chen, Lim, Torralba, Zhao,
  and Wang]{wang2023detecting}
Zhenyu Wang, Yali Li, Xi~Chen, Ser-Nam Lim, Antonio Torralba, Hengshuang Zhao,
  and Shengjin Wang.
\newblock Detecting everything in the open world: Towards universal object
  detection.
\newblock In \emph{Proceedings of the IEEE/CVF Conference on Computer Vision
  and Pattern Recognition}, pages 11433--11443, 2023{\natexlab{i}}.

\bibitem[Wang et~al.(2021{\natexlab{b}})Wang, Yu, Yu, Dai, Tsvetkov, and
  Cao]{wang2021simvlm}
Zirui Wang, Jiahui Yu, Adams~Wei Yu, Zihang Dai, Yulia Tsvetkov, and Yuan Cao.
\newblock Simvlm: Simple visual language model pretraining with weak
  supervision.
\newblock \emph{arXiv preprint arXiv:2108.10904}, 2021{\natexlab{b}}.

\bibitem[Wanna et~al.(2023)Wanna, Parra, Valner, Kruusam{\"a}e, and
  Pryor]{wanna2023multimodal}
Selma Wanna, Fabian Parra, Robert Valner, Karl Kruusam{\"a}e, and Mitch Pryor.
\newblock Multimodal grounding for embodied ai via augmented reality headsets
  for natural language driven task planning.
\newblock \emph{arXiv preprint arXiv:2304.13676}, 2023.

\bibitem[Wei et~al.(2018)Wei, Wang, Yang, and Liu]{wei2018deep}
Chen Wei, Wenjing Wang, Wenhan Yang, and Jiaying Liu.
\newblock Deep retinex decomposition for low-light enhancement.
\newblock \emph{arXiv preprint arXiv:1808.04560}, 2018.

\bibitem[Wei et~al.(2022{\natexlab{a}})Wei, Tay, Bommasani, Raffel, Zoph,
  Borgeaud, Yogatama, Bosma, Zhou, Metzler, et~al.]{wei2022emergent}
Jason Wei, Yi~Tay, Rishi Bommasani, Colin Raffel, Barret Zoph, Sebastian
  Borgeaud, Dani Yogatama, Maarten Bosma, Denny Zhou, Donald Metzler, et~al.
\newblock Emergent abilities of large language models.
\newblock \emph{arXiv preprint arXiv:2206.07682}, 2022{\natexlab{a}}.

\bibitem[Wei et~al.(2022{\natexlab{b}})Wei, Wang, Schuurmans, Bosma, Xia, Chi,
  Le, Zhou, et~al.]{wei2022chain}
Jason Wei, Xuezhi Wang, Dale Schuurmans, Maarten Bosma, Fei Xia, Ed~Chi, Quoc~V
  Le, Denny Zhou, et~al.
\newblock Chain-of-thought prompting elicits reasoning in large language
  models.
\newblock \emph{Advances in Neural Information Processing Systems},
  35:\penalty0 24824--24837, 2022{\natexlab{b}}.

\bibitem[Wenhui et~al.(2023)Wenhui, Xu, Xiaofan, Kang, and
  Zhang]{Lei2023medlam}
Lei Wenhui, Wei Xu, Zhang Xiaofan, Li~Kang, and Shaoting Zhang.
\newblock Medlsam: Localize and segment anything model for 3d medical images.
\newblock \emph{arXiv preprint arXiv:}, 2023.

\bibitem[Wu et~al.(2023{\natexlab{a}})Wu, Yin, Qi, Wang, Tang, and
  Duan]{wu2023visual}
Chenfei Wu, Shengming Yin, Weizhen Qi, Xiaodong Wang, Zecheng Tang, and Nan
  Duan.
\newblock Visual chatgpt: Talking, drawing and editing with visual foundation
  models.
\newblock \emph{arXiv preprint arXiv: 2303.04671}, 2023{\natexlab{a}}.

\bibitem[Wu et~al.(2020)Wu, Lin, Cohen, Bui, and Maji]{wu2020phrasecut}
Chenyun Wu, Zhe Lin, Scott Cohen, Trung Bui, and Subhransu Maji.
\newblock Phrasecut: Language-based image segmentation in the wild.
\newblock In \emph{Proceedings of the IEEE/CVF Conference on Computer Vision
  and Pattern Recognition}, pages 10216--10225, 2020.

\bibitem[Wu et~al.(2023{\natexlab{b}})Wu, Antonova, Kan, Lepert, Zeng, Song,
  Bohg, Rusinkiewicz, and Funkhouser]{wu2023tidybot}
Jimmy Wu, Rika Antonova, Adam Kan, Marion Lepert, Andy Zeng, Shuran Song,
  Jeannette Bohg, Szymon Rusinkiewicz, and Thomas Funkhouser.
\newblock Tidybot: Personalized robot assistance with large language models.
\newblock \emph{arXiv preprint arXiv:2305.05658}, 2023{\natexlab{b}}.

\bibitem[Wu et~al.(2023{\natexlab{c}})Wu, Fu, Fang, Liu, Wang, Xu, Jin, and
  Arbel]{wu2023medical}
Junde Wu, Rao Fu, Huihui Fang, Yuanpei Liu, Zhaowei Wang, Yanwu Xu, Yueming
  Jin, and Tal Arbel.
\newblock Medical sam adapter: Adapting segment anything model for medical
  image segmentation.
\newblock \emph{arXiv preprint arXiv:2304.12620}, 2023{\natexlab{c}}.

\bibitem[Wu et~al.(2023{\natexlab{d}})Wu, Wang, Xu, Lu, and
  Yan]{wu2023embodied}
Zhenyu Wu, Ziwei Wang, Xiuwei Xu, Jiwen Lu, and Haibin Yan.
\newblock Embodied task planning with large language models.
\newblock \emph{arXiv preprint arXiv:2307.01848}, 2023{\natexlab{d}}.

\bibitem[Xu et~al.(2017)Xu, Zhao, Xiao, Wu, Zhang, He, and Zhuang]{xu2017video}
Dejing Xu, Zhou Zhao, Jun Xiao, Fei Wu, Hanwang Zhang, Xiangnan He, and Yueting
  Zhuang.
\newblock Video question answering via gradually refined attention over
  appearance and motion.
\newblock In \emph{Proceedings of the 25th ACM international conference on
  Multimedia}, pages 1645--1653, 2017.

\bibitem[Xu et~al.(2022{\natexlab{a}})Xu, Zhang, Cai, Rezatofighi, Yu, Tao, and
  Geiger]{xu2022unifying}
Haofei Xu, Jing Zhang, Jianfei Cai, Hamid Rezatofighi, Fisher Yu, Dacheng Tao,
  and Andreas Geiger.
\newblock Unifying flow, stereo and depth estimation.
\newblock \emph{arXiv preprint arXiv:2211.05783}, 2022{\natexlab{a}}.

\bibitem[Xu et~al.(2022{\natexlab{b}})Xu, Mello, Liu, Byeon, Breuel, Kautz, and
  Wang]{xu2022groupvit}
Jiarui Xu, Shalini~De Mello, Sifei Liu, Wonmin Byeon, Thomas Breuel, J.~Kautz,
  and X.~Wang.
\newblock Groupvit: Semantic segmentation emerges from text supervision.
\newblock \emph{Computer Vision And Pattern Recognition}, 2022{\natexlab{b}}.
\newblock \doi{10.1109/CVPR52688.2022.01760}.

\bibitem[Xu et~al.(2023{\natexlab{a}})Xu, Liu, Vahdat, Byeon, Wang, and
  De~Mello]{xu2023open}
Jiarui Xu, Sifei Liu, Arash Vahdat, Wonmin Byeon, Xiaolong Wang, and Shalini
  De~Mello.
\newblock Open-vocabulary panoptic segmentation with text-to-image diffusion
  models.
\newblock In \emph{Proceedings of the IEEE/CVF Conference on Computer Vision
  and Pattern Recognition}, pages 2955--2966, 2023{\natexlab{a}}.

\bibitem[Xu et~al.(2016)Xu, Mei, Yao, and Rui]{xu2016msr}
Jun Xu, Tao Mei, Ting Yao, and Yong Rui.
\newblock Msr-vtt: A large video description dataset for bridging video and
  language.
\newblock In \emph{Proceedings of the IEEE conference on computer vision and
  pattern recognition}, pages 5288--5296, 2016.

\bibitem[Xu et~al.(2023{\natexlab{b}})Xu, Zhu, and Clifton]{Peng2023MMT}
Peng Xu, Xiatian Zhu, and David Clifton.
\newblock Multimodal learning with transformers: A survey.
\newblock \emph{TPAMI}, pages 1--21, 2023{\natexlab{b}}.

\bibitem[Xu et~al.(2022{\natexlab{c}})Xu, Wu, Rosenman, Lal, and
  Duan]{xu2022bridge}
Xiao Xu, Chenfei Wu, Shachar Rosenman, Vasudev Lal, and Nan Duan.
\newblock Bridge-tower: Building bridges between encoders in vision-language
  representation learning.
\newblock \emph{arXiv preprint arXiv:2206.08657}, 2022{\natexlab{c}}.

\bibitem[Xue et~al.(2020)Xue, Constant, Roberts, Kale, Al-Rfou, Siddhant,
  Barua, and Raffel]{xue2020mt5}
Linting Xue, Noah Constant, Adam Roberts, Mihir Kale, Rami Al-Rfou, Aditya
  Siddhant, Aditya Barua, and Colin Raffel.
\newblock mt5: A massively multilingual pre-trained text-to-text transformer.
\newblock \emph{arXiv preprint arXiv:2010.11934}, 2020.

\bibitem[Yang et~al.(2021)Yang, Miech, Sivic, Laptev, and Schmid]{yang2021just}
Antoine Yang, Antoine Miech, Josef Sivic, Ivan Laptev, and Cordelia Schmid.
\newblock Just ask: Learning to answer questions from millions of narrated
  videos.
\newblock In \emph{Proceedings of the IEEE/CVF international conference on
  computer vision}, pages 1686--1697, 2021.

\bibitem[Yang et~al.(2023{\natexlab{a}})Yang, Tan, Jin, Liu, Fu, Song, and
  Wang]{yang2023pave}
Jiange Yang, Wenhui Tan, Chuhao Jin, Bei Liu, Jianlong Fu, Ruihua Song, and
  Limin Wang.
\newblock Pave the way to grasp anything: Transferring foundation models for
  universal pick-place robots.
\newblock \emph{arXiv preprint arXiv:2306.05716}, 2023{\natexlab{a}}.

\bibitem[Yang et~al.(2022{\natexlab{a}})Yang, Li, Dai, and Gao]{yang2022focal}
Jianwei Yang, Chunyuan Li, Xiyang Dai, and Jianfeng Gao.
\newblock Focal modulation networks.
\newblock \emph{Advances in Neural Information Processing Systems},
  35:\penalty0 4203--4217, 2022{\natexlab{a}}.

\bibitem[Yang et~al.(2022{\natexlab{b}})Yang, Li, Zhang, Xiao, Liu, Yuan, and
  Gao]{yang2022unified}
Jianwei Yang, Chunyuan Li, Pengchuan Zhang, Bin Xiao, Ce~Liu, Lu~Yuan, and
  Jianfeng Gao.
\newblock Unified contrastive learning in image-text-label space.
\newblock In \emph{Proceedings of the IEEE/CVF Conference on Computer Vision
  and Pattern Recognition}, pages 19163--19173, 2022{\natexlab{b}}.

\bibitem[Yang et~al.(2023{\natexlab{b}})Yang, Jin, Tang, Han, Feng, Jiang, Yin,
  and Hu]{Jingfeng2023LLM}
Jingfeng Yang, Hongye Jin, Ruixiang Tang, Xiaotian Han, Qizhang Feng, Haoming
  Jiang, Bing Yin, and Xia Hu.
\newblock Harnessing the power of llms in practice: A survey on chatgpt and
  beyond.
\newblock \emph{arXiv preprint arXiv:2304.13712}, 2023{\natexlab{b}}.

\bibitem[Yang et~al.(2023{\natexlab{c}})Yang, Gao, Li, Gao, Wang, and
  Zheng]{yang2023track}
Jinyu Yang, Mingqi Gao, Zhe Li, Shang Gao, Fangjing Wang, and Feng Zheng.
\newblock Track anything: Segment anything meets videos.
\newblock \emph{arXiv preprint arXiv: 2304.11968}, 2023{\natexlab{c}}.

\bibitem[Yang et~al.(2022{\natexlab{c}})Yang, Zhang, Song, Hong, Xu, Zhao,
  Shao, Zhang, Cui, and Yang]{yang2022diffusion}
Ling Yang, Zhilong Zhang, Yang Song, Shenda Hong, Runsheng Xu, Yue Zhao,
  Yingxia Shao, Wentao Zhang, Bin Cui, and Ming-Hsuan Yang.
\newblock Diffusion models: A comprehensive survey of methods and applications.
\newblock \emph{arXiv preprint arXiv:2209.00796}, 2022{\natexlab{c}}.

\bibitem[Yang et~al.(2023{\natexlab{d}})Yang, Li, Wang, Lin, Azarnasab, Ahmed,
  Liu, Liu, Zeng, and Wang]{yang2023mmreact}
Zhengyuan Yang, Linjie Li, Jianfeng Wang, Kevin Lin, Ehsan Azarnasab, Faisal
  Ahmed, Zicheng Liu, Ce~Liu, Michael Zeng, and Lijuan Wang.
\newblock Mm-react: Prompting chatgpt for multimodal reasoning and action.
\newblock \emph{arXiv preprint arXiv: 2303.11381}, 2023{\natexlab{d}}.

\bibitem[Yang and Yang(2022)]{yang2022decoupling}
Zongxin Yang and Yi~Yang.
\newblock Decoupling features in hierarchical propagation for video object
  segmentation.
\newblock \emph{Advances in Neural Information Processing Systems},
  35:\penalty0 36324--36336, 2022.

\bibitem[Yao et~al.(2021)Yao, Huang, Hou, Lu, Niu, Xu, Liang, Li, Jiang, and
  Xu]{yao2021filip}
Lewei Yao, Runhui Huang, Lu~Hou, Guansong Lu, Minzhe Niu, Hang Xu, Xiaodan
  Liang, Zhenguo Li, Xin Jiang, and Chunjing Xu.
\newblock Filip: fine-grained interactive language-image pre-training.
\newblock \emph{arXiv preprint arXiv:2111.07783}, 2021.

\bibitem[Ye et~al.(2022)Ye, Fu, Zheng, Paudel, and Chen]{ye2022unsupervised}
Junjie Ye, Changhong Fu, Guangze Zheng, Danda~Pani Paudel, and Guang Chen.
\newblock Unsupervised domain adaptation for nighttime aerial tracking.
\newblock In \emph{Proceedings of the IEEE/CVF Conference on Computer Vision
  and Pattern Recognition}, pages 8896--8905, 2022.

\bibitem[Ye et~al.(2019)Ye, Rochan, Liu, and Wang]{ye2019cross}
Linwei Ye, Mrigank Rochan, Zhi Liu, and Yang Wang.
\newblock Cross-modal self-attention network for referring image segmentation.
\newblock In \emph{Proceedings of the IEEE/CVF conference on computer vision
  and pattern recognition}, pages 10502--10511, 2019.

\bibitem[Ye et~al.(2023)Ye, Xu, Xu, Ye, Yan, Zhou, Wang, Hu, Shi, Shi, Jiang,
  Li, Xu, Chen, Tian, Qi, Zhang, and Huang]{ye2023mplugowl}
Qinghao Ye, Haiyang Xu, Guohai Xu, Jiabo Ye, Ming Yan, Yiyang Zhou, Junyang
  Wang, Anwen Hu, Pengcheng Shi, Yaya Shi, Chaoya Jiang, Chenliang Li, Yuanhong
  Xu, Hehong Chen, Junfeng Tian, Qian Qi, Ji~Zhang, and Fei Huang.
\newblock mplug-owl: Modularization empowers large language models with
  multimodality, 2023.

\bibitem[Yoneda et~al.(2023)Yoneda, Fang, Li, Zhang, Jiang, Lin, Picker, Yunis,
  Mei, and Walter]{yoneda2023statler}
Takuma Yoneda, Jiading Fang, Peng Li, Huanyu Zhang, Tianchong Jiang, Shengjie
  Lin, Ben Picker, David Yunis, Hongyuan Mei, and Matthew~R Walter.
\newblock Statler: State-maintaining language models for embodied reasoning.
\newblock \emph{arXiv preprint arXiv:2306.17840}, 2023.

\bibitem[Yonglin et~al.(2023)Yonglin, Jing, Xiao, and Long]{refsam}
Li~Yonglin, Zhang Jing, Teng Xiao, and Lan Long.
\newblock Refsam: Efficiently adapting segmenting anything model for referring
  video object segmentation.
\newblock \emph{arXiv:2307.00997}, 2023.

\bibitem[You et~al.(2023)You, Ye, Zhou, Zhu, and Du]{you2023robot}
Hengxu You, Yang Ye, Tianyu Zhou, Qi~Zhu, and Jing Du.
\newblock Robot-enabled construction assembly with automated sequence planning
  based on chatgpt: Robogpt.
\newblock \emph{arXiv preprint arXiv:2304.11018}, 2023.

\bibitem[Young et~al.(2014)Young, Lai, Hodosh, and Hockenmaier]{young2014image}
Peter Young, Alice Lai, Micah Hodosh, and Julia Hockenmaier.
\newblock From image descriptions to visual denotations: New similarity metrics
  for semantic inference over event descriptions.
\newblock \emph{Transactions of the Association for Computational Linguistics},
  2:\penalty0 67--78, 2014.

\bibitem[Yu et~al.(2022)Yu, Wang, Vasudevan, Yeung, Seyedhosseini, and
  Wu]{yu2022coca}
Jiahui Yu, Zirui Wang, Vijay Vasudevan, Legg Yeung, Mojtaba Seyedhosseini, and
  Yonghui Wu.
\newblock Coca: Contrastive captioners are image-text foundation models.
\newblock \emph{arXiv preprint arXiv:2205.01917}, 2022.

\bibitem[Yu et~al.(2016)Yu, Poirson, Yang, Berg, and Berg]{yu2016modeling}
Licheng Yu, Patrick Poirson, Shan Yang, Alexander~C Berg, and Tamara~L Berg.
\newblock Modeling context in referring expressions.
\newblock In \emph{Computer Vision--ECCV 2016: 14th European Conference,
  Amsterdam, The Netherlands, October 11-14, 2016, Proceedings, Part II 14},
  pages 69--85. Springer, 2016.

\bibitem[Yuan et~al.(2021)Yuan, Chen, Chen, Codella, Dai, Gao, Hu, Huang, Li,
  Li, et~al.]{yuan2021florence}
Lu~Yuan, Dongdong Chen, Yi-Ling Chen, Noel Codella, Xiyang Dai, Jianfeng Gao,
  Houdong Hu, Xuedong Huang, Boxin Li, Chunyuan Li, et~al.
\newblock Florence: A new foundation model for computer vision.
\newblock \emph{arXiv preprint arXiv:2111.11432}, 2021.

\bibitem[Yuan et~al.(2023)Yuan, Xue, Wang, Liu, Zhao, and Wang]{yuan2023artgpt}
Zhengqing Yuan, Huiwen Xue, Xinyi Wang, Yongming Liu, Zhuanzhe Zhao, and Kun
  Wang.
\newblock Artgpt-4: Artistic vision-language understanding with
  adapter-enhanced minigpt-4.
\newblock \emph{arXiv preprint arXiv:2305.07490}, 2023.

\bibitem[Zhai and Wu(2018)]{zhai2018classification}
Andrew Zhai and Hao-Yu Wu.
\newblock Classification is a strong baseline for deep metric learning.
\newblock \emph{arXiv preprint arXiv:1811.12649}, 2018.

\bibitem[Zhai et~al.(2022)Zhai, Kolesnikov, Houlsby, and
  Beyer]{zhai2022scaling}
Xiaohua Zhai, Alexander Kolesnikov, Neil Houlsby, and Lucas Beyer.
\newblock Scaling vision transformers.
\newblock In \emph{Proceedings of the IEEE/CVF Conference on Computer Vision
  and Pattern Recognition}, pages 12104--12113, 2022.

\bibitem[Zhang et~al.(2023{\natexlab{a}})Zhang, Fei, Yao, Ji, Li, Liu, and
  Chua]{zhang2023transfer}
Ao~Zhang, Hao Fei, Yuan Yao, Wei Ji, Li~Li, Zhiyuan Liu, and Tat-Seng Chua.
\newblock Transfer visual prompt generator across llms.
\newblock \emph{arXiv preprint arXiv: 2305.01278}, 2023{\natexlab{a}}.

\bibitem[Zhang et~al.(2023{\natexlab{b}})Zhang, Ge, Xu, Shan, and
  Shou]{zhang2023taca}
Binjie Zhang, Yixiao Ge, Xuyuan Xu, Ying Shan, and Mike~Zheng Shou.
\newblock Taca: Upgrading your visual foundation model with task-agnostic
  compatible adapter.
\newblock \emph{arXiv preprint arXiv: 2306.12642}, 2023{\natexlab{b}}.

\bibitem[Zhang et~al.(2023{\natexlab{c}})Zhang, Han, Qiao, Kim, Bae, Lee, and
  Hong]{mobile_sam}
Chaoning Zhang, Dongshen Han, Yu~Qiao, Jung~Uk Kim, Sung-Ho Bae, Seungkyu Lee,
  and Choong~Seon Hong.
\newblock Faster segment anything: Towards lightweight sam for mobile
  applications.
\newblock \emph{arXiv preprint arXiv:2306.14289}, 2023{\natexlab{c}}.

\bibitem[Zhang et~al.(2023{\natexlab{d}})Zhang, Han, Qiao, Kim, Bae, Lee, and
  Hong]{zhang2023faster}
Chaoning Zhang, Dongshen Han, Yu~Qiao, Jung~Uk Kim, Sung-Ho Bae, Seungkyu Lee,
  and Choong~Seon Hong.
\newblock Faster segment anything: Towards lightweight sam for mobile
  applications.
\newblock \emph{arXiv preprint arXiv: 2306.14289}, 2023{\natexlab{d}}.

\bibitem[Zhang et~al.(2023{\natexlab{e}})Zhang, Zhang, Kang, Kim, Bae, and
  Kweon]{zhang2023attack}
Chenshuang Zhang, Chaoning Zhang, Taegoo Kang, Donghun Kim, Sung-Ho Bae, and
  In~So Kweon.
\newblock Attack-sam: Towards evaluating adversarial robustness of segment
  anything model.
\newblock \emph{arXiv preprint arXiv:2305.00866}, 2023{\natexlab{e}}.

\bibitem[Zhang et~al.(2023{\natexlab{f}})Zhang, Zhang, Zhang, and
  Kweon]{zhang2023text}
Chenshuang Zhang, Chaoning Zhang, Mengchun Zhang, and In~So Kweon.
\newblock Text-to-image diffusion model in generative ai: A survey.
\newblock \emph{arXiv preprint arXiv:2303.07909}, 2023{\natexlab{f}}.

\bibitem[Zhang et~al.(2023{\natexlab{g}})Zhang, Liu, Cui, Huang, Lin, Yang, and
  Hu]{Chunhui2023SAM}
Chunhui Zhang, Li~Liu, Yawen Cui, Guanjie Huang, Weilin Lin, Yiqian Yang, and
  Yuehong Hu.
\newblock A comprehensive survey on segment anything model for vision and
  beyond.
\newblock \emph{arXiv preprint arXiv:2305.08196}, 2023{\natexlab{g}}.

\bibitem[Zhang et~al.(2022{\natexlab{a}})Zhang, Zhang, Hu, Chen, Li, Dai, Wang,
  Yuan, Hwang, and Gao]{zhang2022glipv2}
Haotian Zhang, Pengchuan Zhang, Xiaowei Hu, Yen-Chun Chen, Liunian Li, Xiyang
  Dai, Lijuan Wang, Lu~Yuan, Jenq-Neng Hwang, and Jianfeng Gao.
\newblock Glipv2: Unifying localization and vision-language understanding.
\newblock \emph{Advances in Neural Information Processing Systems},
  35:\penalty0 36067--36080, 2022{\natexlab{a}}.

\bibitem[Zhang et~al.(2023{\natexlab{h}})Zhang, Du, Shan, Zhou, Du, Tenenbaum,
  Shu, and Gan]{zhang2023building}
Hongxin Zhang, Weihua Du, Jiaming Shan, Qinhong Zhou, Yilun Du, Joshua~B
  Tenenbaum, Tianmin Shu, and Chuang Gan.
\newblock Building cooperative embodied agents modularly with large language
  models.
\newblock \emph{arXiv preprint arXiv:2307.02485}, 2023{\natexlab{h}}.

\bibitem[Zhang et~al.(2023{\natexlab{i}})Zhang, Pertsch, Zhang, and
  Lim]{zhang2023sprint}
Jesse Zhang, Karl Pertsch, Jiahui Zhang, and Joseph~J Lim.
\newblock Sprint: Scalable policy pre-training via language instruction
  relabeling.
\newblock \emph{arXiv preprint arXiv:2306.11886}, 2023{\natexlab{i}}.

\bibitem[Zhang(2023{\natexlab{a}})]{zhang2023graph}
Jiawei Zhang.
\newblock Graph-toolformer: To empower llms with graph reasoning ability via
  prompt augmented by chatgpt.
\newblock \emph{arXiv preprint arXiv:2304.11116}, 2023{\natexlab{a}}.

\bibitem[Zhang(2023{\natexlab{b}})]{zhang2023graphtoolformer}
Jiawei Zhang.
\newblock Graph-toolformer: To empower llms with graph reasoning ability via
  prompt augmented by chatgpt.
\newblock \emph{arXiv preprint arXiv: 2304.11116}, 2023{\natexlab{b}}.

\bibitem[Zhang et~al.(2023{\natexlab{j}})Zhang, Han, Zhou, Hu, Yan, Lu, Li,
  Gao, and Qiao]{zhang2023llama}
Renrui Zhang, Jiaming Han, Aojun Zhou, Xiangfei Hu, Shilin Yan, Pan Lu,
  Hongsheng Li, Peng Gao, and Yu~Qiao.
\newblock Llama-adapter: Efficient fine-tuning of language models with
  zero-init attention.
\newblock \emph{arXiv preprint arXiv:2303.16199}, 2023{\natexlab{j}}.

\bibitem[Zhang et~al.(2023{\natexlab{k}})Zhang, Sun, Chen, Xiao, Shao, Zhang,
  Chen, and Luo]{zhang2023gpt4roi}
Shilong Zhang, Peize Sun, Shoufa Chen, Min Xiao, Wenqi Shao, Wenwei Zhang, Kai
  Chen, and Ping Luo.
\newblock Gpt4roi: Instruction tuning large language model on
  region-of-interest, 2023{\natexlab{k}}.

\bibitem[Zhang et~al.(2022{\natexlab{b}})Zhang, Roller, Goyal, Artetxe, Chen,
  Chen, Dewan, Diab, Li, Lin, et~al.]{zhang2022opt}
Susan Zhang, Stephen Roller, Naman Goyal, Mikel Artetxe, Moya Chen, Shuohui
  Chen, Christopher Dewan, Mona Diab, Xian Li, Xi~Victoria Lin, et~al.
\newblock Opt: Open pre-trained transformer language models.
\newblock \emph{arXiv preprint arXiv:2205.01068}, 2022{\natexlab{b}}.

\bibitem[Zhang et~al.(2021)Zhang, Zhang, Li, Xu, Wang, Zhan, Xu, Ke, Zeng, Su,
  et~al.]{zhang2021sar}
Tianwen Zhang, Xiaoling Zhang, Jianwei Li, Xiaowo Xu, Baoyou Wang, Xu~Zhan,
  Yanqin Xu, Xiao Ke, Tianjiao Zeng, Hao Su, et~al.
\newblock Sar ship detection dataset (ssdd): Official release and comprehensive
  data analysis.
\newblock \emph{Remote Sensing}, 13\penalty0 (18):\penalty0 3690, 2021.

\bibitem[Zhang et~al.(2023{\natexlab{l}})Zhang, Zeng, Zhang, and
  Li]{zhang2023toward}
Xinsong Zhang, Yan Zeng, Jipeng Zhang, and Hang Li.
\newblock Toward building general foundation models for language, vision, and
  vision-language understanding tasks.
\newblock \emph{arXiv preprint arXiv:2301.05065}, 2023{\natexlab{l}}.

\bibitem[Zhang et~al.(2023{\natexlab{m}})Zhang, Zhang, Gu, Zhou, Lipka, Yang,
  and Sun]{zhang2023llavar}
Yanzhe Zhang, Ruiyi Zhang, Jiuxiang Gu, Yufan Zhou, Nedim Lipka, Diyi Yang, and
  Tong Sun.
\newblock Llavar: Enhanced visual instruction tuning for text-rich image
  understanding.
\newblock \emph{arXiv preprint arXiv: 2306.17107}, 2023{\natexlab{m}}.

\bibitem[Zhang et~al.(2023{\natexlab{n}})Zhang, Huang, Ma, Li, Luo, Xie, Qin,
  Luo, Li, Liu, et~al.]{zhang2023recognize}
Youcai Zhang, Xinyu Huang, Jinyu Ma, Zhaoyang Li, Zhaochuan Luo, Yanchun Xie,
  Yuzhuo Qin, Tong Luo, Yaqian Li, Shilong Liu, et~al.
\newblock Recognize anything: A strong image tagging model.
\newblock \emph{arXiv preprint arXiv:2306.03514}, 2023{\natexlab{n}}.

\bibitem[Zhang et~al.(2023{\natexlab{o}})Zhang, Zhang, Li, Zhao, Karypis, and
  Smola]{zhang2023multimodal}
Zhuosheng Zhang, Aston Zhang, Mu~Li, Hai Zhao, George Karypis, and Alex Smola.
\newblock Multimodal chain-of-thought reasoning in language models.
\newblock \emph{arXiv preprint arXiv:2302.00923}, 2023{\natexlab{o}}.

\bibitem[Zhao et~al.(2023{\natexlab{a}})Zhao, Yuan, Jiang, Cai, Yu, Wang, and
  Chen]{zhao2023erra}
Chao Zhao, Shuai Yuan, Chunli Jiang, Junhao Cai, Hongyu Yu, Michael~Yu Wang,
  and Qifeng Chen.
\newblock Erra: An embodied representation and reasoning architecture for
  long-horizon language-conditioned manipulation tasks.
\newblock \emph{IEEE Robotics and Automation Letters}, 2023{\natexlab{a}}.

\bibitem[Zhao et~al.(2023{\natexlab{b}})Zhao, Zhou, Li, Tang, Wang, Hou, Min,
  Zhang, Zhang, Dong, Du, Yang, Chen, Chen, Jiang, Ren, Li, Tang, Liu, Liu,
  Nie, and Wen]{Zhao2023LLM}
Wayne~Xin Zhao, Kun Zhou, Junyi Li, Tianyi Tang, Xiaolei Wang, Yupeng Hou,
  Yingqian Min, Beichen Zhang, Junjie Zhang, Zican Dong, Yifan Du, Chen Yang,
  Yushuo Chen, Zhipeng Chen, Jinhao Jiang, Ruiyang Ren, Yifan Li, Xinyu Tang,
  Zikang Liu, Peiyu Liu, Jian-Yun Nie, and Ji-Rong Wen.
\newblock A survey of large language models.
\newblock \emph{arXiv preprint arXiv:2303.18223}, 2023{\natexlab{b}}.

\bibitem[Zhao et~al.(2023{\natexlab{c}})Zhao, Ding, An, Du, Yu, Li, Tang, and
  Wang]{zhao2023fast}
Xu~Zhao, Wenchao Ding, Yongqi An, Yinglong Du, Tao Yu, Min Li, Ming Tang, and
  Jinqiao Wang.
\newblock Fast segment anything.
\newblock \emph{arXiv preprint arXiv: 2306.12156}, 2023{\natexlab{c}}.

\bibitem[Zhao et~al.(2023{\natexlab{d}})Zhao, Li, Weber, Hafez, and
  Wermter]{zhao2023chat}
Xufeng Zhao, Mengdi Li, Cornelius Weber, Muhammad~Burhan Hafez, and Stefan
  Wermter.
\newblock Chat with the environment: Interactive multimodal perception using
  large language models.
\newblock \emph{arXiv preprint arXiv: 2303.08268}, 2023{\natexlab{d}}.

\bibitem[Zhao et~al.(2023{\natexlab{e}})Zhao, Lin, Zhou, Huang, Feng, and
  Kang]{zhao2023bubogpt}
Yang Zhao, Zhijie Lin, Daquan Zhou, Zilong Huang, Jiashi Feng, and Bingyi Kang.
\newblock Bubogpt: Enabling visual grounding in multi-modal llms.
\newblock \emph{arXiv preprint arXiv:2307.08581}, 2023{\natexlab{e}}.

\bibitem[Zheng et~al.(2023{\natexlab{a}})Zheng, Chiang, Sheng, Zhuang, Wu,
  Zhuang, Lin, Li, Li, Xing, et~al.]{zheng2023judging}
Lianmin Zheng, Wei-Lin Chiang, Ying Sheng, Siyuan Zhuang, Zhanghao Wu, Yonghao
  Zhuang, Zi~Lin, Zhuohan Li, Dacheng Li, Eric Xing, et~al.
\newblock Judging llm-as-a-judge with mt-bench and chatbot arena.
\newblock \emph{arXiv preprint arXiv:2306.05685}, 2023{\natexlab{a}}.

\bibitem[Zheng et~al.(2023{\natexlab{b}})Zheng, Su, You, Wang, Qian, Xu, and
  Albanie]{zheng2023can}
Mingkai Zheng, Xiu Su, Shan You, Fei Wang, Chen Qian, Chang Xu, and Samuel
  Albanie.
\newblock Can gpt-4 perform neural architecture search?
\newblock \emph{arXiv preprint arXiv:2304.10970}, 2023{\natexlab{b}}.

\bibitem[Zheng et~al.(2023{\natexlab{c}})Zheng, Su, You, Wang, Qian, Xu, and
  Albanie]{zheng2023gpt4}
Mingkai Zheng, Xiu Su, Shan You, Fei Wang, Chen Qian, Chang Xu, and Samuel
  Albanie.
\newblock Can gpt-4 perform neural architecture search?
\newblock \emph{arXiv preprint arXiv: 2304.10970}, 2023{\natexlab{c}}.

\bibitem[Zhong et~al.(2023)Zhong, Wei, Yang, Wu, Liu, Wei, Li, Yao, Ma, Li,
  et~al.]{zhong2023chatabl}
Tianyang Zhong, Yaonai Wei, Li~Yang, Zihao Wu, Zhengliang Liu, Xiaozheng Wei,
  Wenjun Li, Junjie Yao, Chong Ma, Xiang Li, et~al.
\newblock Chatabl: Abductive learning via natural language interaction with
  chatgpt.
\newblock \emph{arXiv preprint arXiv:2304.11107}, 2023.

\bibitem[Zhong et~al.(2022)Zhong, Yang, Zhang, Li, Codella, Li, Zhou, Dai,
  Yuan, Li, et~al.]{zhong2022regionclip}
Yiwu Zhong, Jianwei Yang, Pengchuan Zhang, Chunyuan Li, Noel Codella,
  Liunian~Harold Li, Luowei Zhou, Xiyang Dai, Lu~Yuan, Yin Li, et~al.
\newblock Regionclip: Region-based language-image pretraining.
\newblock In \emph{Proceedings of the IEEE/CVF Conference on Computer Vision
  and Pattern Recognition}, pages 16793--16803, 2022.

\bibitem[Zhou et~al.(2017)Zhou, Zhao, Puig, Fidler, Barriuso, and
  Torralba]{zhou2017scene}
Bolei Zhou, Hang Zhao, Xavier Puig, Sanja Fidler, Adela Barriuso, and Antonio
  Torralba.
\newblock Scene parsing through ade20k dataset.
\newblock In \emph{Proceedings of the IEEE conference on computer vision and
  pattern recognition}, pages 633--641, 2017.

\bibitem[Zhou et~al.(2019)Zhou, Zhao, Puig, Xiao, Fidler, Barriuso, and
  Torralba]{zhou2019semantic}
Bolei Zhou, Hang Zhao, Xavier Puig, Tete Xiao, Sanja Fidler, Adela Barriuso,
  and Antonio Torralba.
\newblock Semantic understanding of scenes through the ade20k dataset.
\newblock \emph{International Journal of Computer Vision}, 127:\penalty0
  302--321, 2019.

\bibitem[Zhou et~al.(2023)Zhou, Li, Li, Yu, Liu, Wang, Zhang, Ji, Yan, He,
  et~al.]{zhou2023comprehensive}
Ce~Zhou, Qian Li, Chen Li, Jun Yu, Yixin Liu, Guangjing Wang, Kai Zhang, Cheng
  Ji, Qiben Yan, Lifang He, et~al.
\newblock A comprehensive survey on pretrained foundation models: A history
  from bert to chatgpt.
\newblock \emph{arXiv preprint arXiv:2302.09419}, 2023.

\bibitem[Zhou et~al.(2022)Zhou, Loy, and Dai]{zhou2022maskclip}
Chong Zhou, Chen~Change Loy, and Bo~Dai.
\newblock Extract free dense labels from clip.
\newblock In \emph{European Conference on Computer Vision (ECCV)}, 2022.

\bibitem[Zhu et~al.(2023{\natexlab{a}})Zhu, Chen, Haydarov, Shen, Zhang, and
  Elhoseiny]{zhu2023chatgpt}
Deyao Zhu, Jun Chen, Kilichbek Haydarov, Xiaoqian Shen, Wenxuan Zhang, and
  Mohamed Elhoseiny.
\newblock Chatgpt asks, blip-2 answers: Automatic questioning towards enriched
  visual descriptions.
\newblock \emph{arXiv preprint arXiv:2303.06594}, 2023{\natexlab{a}}.

\bibitem[Zhu et~al.(2023{\natexlab{b}})Zhu, Chen, Shen, Li, and
  Elhoseiny]{zhu2023minigpt}
Deyao Zhu, Jun Chen, Xiaoqian Shen, Xiang Li, and Mohamed Elhoseiny.
\newblock Minigpt-4: Enhancing vision-language understanding with advanced
  large language models.
\newblock \emph{arXiv preprint arXiv:2304.10592}, 2023{\natexlab{b}}.

\bibitem[Zhu et~al.(2007)Zhu, Mumford, et~al.]{zhu2007stochastic}
Song-Chun Zhu, David Mumford, et~al.
\newblock A stochastic grammar of images.
\newblock \emph{Foundations and Trends{\textregistered} in Computer Graphics
  and Vision}, 2\penalty0 (4):\penalty0 259--362, 2007.

\bibitem[Zhu et~al.(2023{\natexlab{c}})Zhu, Hessel, Awadalla, Gadre, Dodge,
  Fang, Yu, Schmidt, Wang, and Choi]{zhu2023multimodal}
Wanrong Zhu, Jack Hessel, Anas Awadalla, Samir~Yitzhak Gadre, Jesse Dodge, Alex
  Fang, Youngjae Yu, Ludwig Schmidt, William~Yang Wang, and Yejin Choi.
\newblock Multimodal c4: An open, billion-scale corpus of images interleaved
  with text, 2023{\natexlab{c}}.

\bibitem[Zhu et~al.(2016)Zhu, Groth, Bernstein, and Fei-Fei]{zhu2016visual7w}
Yuke Zhu, Oliver Groth, Michael Bernstein, and Li~Fei-Fei.
\newblock Visual7w: Grounded question answering in images, 2016.

\bibitem[Zong et~al.(2023)Zong, Aodha, and Hospedales]{Zong2023SSML}
Yongshuo Zong, Oisin~Mac Aodha, and Timothy Hospedales.
\newblock Self-supervised multimodal learning: A survey.
\newblock \emph{arXiv preprint arXiv:2304.01008}, 2023.

\bibitem[Zou et~al.(2022)Zou, Dou, Yang, Gan, Li, Li, Dai, Wang, Yuan, Peng,
  Wang, Lee, and Gao]{zou2022xdecoder}
Xueyan Zou, Zi-Yi Dou, Jianwei Yang, Zhe Gan, Linjie Li, Chunyuan Li, Xiyang
  Dai, Jianfeng Wang, Lu~Yuan, Nanyun Peng, Lijuan Wang, Yong~Jae Lee, and
  Jianfeng Gao.
\newblock Generalized decoding for pixel, image and language.
\newblock 2022.

\bibitem[Zou et~al.(2023)Zou, Yang, Zhang, Li, Li, Gao, and
  Lee]{zou2023segment}
Xueyan Zou, Jianwei Yang, Hao Zhang, Feng Li, Linjie Li, Jianfeng Gao, and
  Yong~Jae Lee.
\newblock Segment everything everywhere all at once.
\newblock \emph{arXiv preprint arXiv:2304.06718}, 2023.

\end{thebibliography}
}

\newpage
\end{document}